\documentclass{article}

\PassOptionsToPackage{square,sort,comma,numbers}{natbib}

\usepackage[preprint]{neurips_2023}
\usepackage{epsfig}
\usepackage[utf8]{inputenc} 
\usepackage[T1]{fontenc}    
\usepackage{hyperref}       
\usepackage{url}            
\usepackage{booktabs}       
\usepackage{amsfonts}       
\usepackage{nicefrac}       
\usepackage{microtype}      
\usepackage{xcolor}         

\usepackage{wrapfig}
\usepackage{mathrsfs}
\usepackage{tabularx}
\usepackage{amsmath}
\usepackage{amssymb}
\usepackage{stmaryrd}
\usepackage{graphicx}
\usepackage{float}
\usepackage{bbold}
\graphicspath{ {./Image/} }
\usepackage[english]{babel}
\usepackage{graphicx}
\usepackage{fullpage}
\usepackage{caption}
\usepackage{subcaption}
\usepackage{lipsum}
\usepackage{ comment }
\usepackage{geometry}
\usepackage{algorithm}
\usepackage{algpseudocode}
\usepackage{enumitem}
\usepackage{tikz}
\usepackage{amsthm}
\usepackage{ marvosym }
\usepackage{todonotes}

\newcommand{\lc}[1]{\text{#1}}

\newcommand{\rgg}[1]{{\textcolor{purple}{#1}}}

\newcommand*{\defeq}{\stackrel{\text{def}}{ \,\, = \,}}

\newcommand{\R}{\mathbb{R}}
\newcommand{\argmin}{\text{argmin}}
\newcommand{\pivot}{\mu^{1 \to 2}_\theta}

\newcommand{\geodesic}{\mu^{1 \to 2}(t)}
\newcommand{\bary}{\mu^{1 \to2}}
\newcommand{\genegeodesic}{\mu^{1 \to 2}_{g}(t)}
\newcommand{\gene}{\mu^{1 \to2}_{g, \theta}}
\newcommand{\gbarycenter}{\mu^{1 \to2}_{g}}

\newcommand{\lgbarycenter}{\mu^{1 \to2}_{g, \theta}}

\newcommand{\thetap}{P^{\theta}_\#}
\newcommand{\thetapp}{Q^{\theta}_\#}
\newcommand{\swgg}{\textnormal{SWGG}}
\newcommand{\swggr}{\textnormal{SWGG}_2}
\newcommand{\mswgg}{\textnormal{min-SWGG}}
\newcommand{\pwd}{\textnormal{PWD}}
\newcommand{\maxsw}{\textnormal{max-SW}}
\newcommand{\sw}{\textnormal{SW}}

\newcommand{\tow}{\overset{\mathcal{L}}{\longrightarrow}}
\newcommand{\towd}{\overset{\mathcal{L},2}{\longrightarrow}}

\newcommand{\tok}{\underset{k}{\longrightarrow}}
\newcommand{\Sd}{\mathbb{S}^{d-1}}
\newcommand{\Sn}{\mathcal{S}(n)}

\newcommand{\Pd}{\mathcal{P}_2(\R^d)}
\newcommand{\Pdn}{\mathcal{P}^n_2(\R^d)}
\newcommand{\Pall}{\mathcal{P}(\R^d)}

\newcommand{\D}{\mathcal{D}}

\theoremstyle{definition}
\newtheorem{thm}{Theorem}[section]
\newtheorem{definition}[thm]{Definition}

\newtheorem{proposition}[thm]{Proposition}

\newtheorem{lemma}[thm]{Lemma}

\newtheorem{remark}[thm]{Remark}

\title{Fast Optimal Transport through Sliced Wasserstein Generalized  Geodesics}

\author{Guillaume Mahey\\
    INSA Rouen-Université Bretagne Sud\\
    LITIS - IRISA\\
    \texttt{guillaume.mahey@insa-rouen.fr}\\
    \And
    Laetitia Chapel\\
    Université Bretagne Sud\\
    IRISA\\
    \texttt{laetitia.chapel@univ-ubs.fr}\\
    \And
    Gilles Gasso\\
    INSA Rouen\\
    LITIS\\
    \texttt{gilles.gasso@insa-rouen.fr}\\
    \And
    Clément Bonet\\
    Université Bretagne Sud\\
    LMBA\\
    \texttt{clément.bonet@univ-ubs.fr}\\
    \And
    Nicolas Courty\\
    Université Bretagne Sud\\
    IRISA\\
    \texttt{nicolas.courty@univ-ubs.fr}\\
}

\begin{document}

\maketitle

\begin{abstract}

Wasserstein distance (WD) and the associated optimal transport plan have proven 
useful in many applications where probability measures are at stake. In this paper, we propose a new proxy \lc{for} the squared WD, coined $\mswgg$, \lc{which relies} on the transport map induced by an optimal one-dimensional projection of the two input distributions. We draw connections between $\mswgg$ and Wasserstein generalized geodesics with a pivot measure 
supported on a line. We notably provide a new closed form of 
the Wasserstein distance in the particular case \lc{where} one of the distributions is supported on a line, allowing us to derive a fast computational scheme that is amenable to gradient descent optimization. We show that $\mswgg$ is an upper bound of WD and that it has a complexity similar to \lc{that of} Sliced-Wasserstein, with the additional feature of providing an associated transport plan. We also investigate some theoretical properties such as metricity, weak convergence, computational and topological properties. Empirical evidences support the benefits of $\mswgg$ in various contexts, from gradient flows, shape matching and image colorization, among others.
\end{abstract}

\section{Introduction}
Gaspard Monge, in his seminal work on Optimal Transport (OT)  \cite{monge1781memoire}, studied the following problem: how to move with minimum cost the probability mass of a source measure to a target one, for a given transfer cost function? At the heart of OT is the optimal map that describes the optimal displacement as the Monge problem can be reformulated as an assignment problem. It has been relaxed by \cite{kantorovich1942translocation} \lc{by} finding a plan that describes the amount of mass moving from the source to the target. Beyond this optimal plan, an interest of OT is that it defines a distance between probability measures: the Wasserstein distance (WD). 

Recently, OT has been successfully employed in a wide range of machine learning applications, in which the Wasserstein distance is estimated from the data, such as supervised learning \cite{frogner2015learning}, natural language processing \cite{kusner2015word} or generative modelling \cite{arjovsky2017wasserstein}. Its \lc{capacity} to provide meaningful distances between empirical distributions is at the core of distance-based algorithms such as kernel-based methods \cite{togninalli2019wasserstein} or $k$-nearest neighbors \cite{backurs2020scalable}. The optimal transport plan has also been used successfully in many applications {where a matching between empirical samples is sought} such as color transfer \cite{rabin2014adaptive}, domain adaptation \cite{courty2017Optimal} and positive-unlabeled learning \cite{chapel2020partial}.

Solving the OT problem is computationally intensive; the most common algorithmic tools to solve the discrete OT problem are borrowed from combinatorial optimization and linear programming, leading to a cubic complexity with the number of samples that prevents its use in large scale application\lc{s} \cite{COTFNT}. To reduce the computation burden, regularizing the OT problem with e.g. an entropic term \lc{has led to} solvers with a quadratic complexity \cite{cuturi2013sinkhorn}. Other methods based on the existence of a closed form of OT \lc{have also been} devised to efficiently compute a proxy \lc{for} WD\lc{, as outlined below}. 

\noindent\textbf{Projections-based OT.} 
\lc{The} Sliced-Wasserstein distance (SWD) \cite{Rabin-SWD-2011, bonneel2015sliced} leverages 1D-projections of distributions to \lc{provide} a lower approximation of the Wasserstein distance, relying on the closed form of OT for 1D probability distributions. Computation of SWD leads to a linearithmic time complexity. 
While SWD averages WDs computed over several 1D projections, max-SWD \cite{deshpande2019max} 
keeps only the most informative projection. 
\lc{These frameworks provide} efficient algorithms that can handle millions of samples 
and have similar topological properties as WD \cite{nadjahi2020statistical}. 
Other 
works restrain SWD and max-SWD to projections onto low dimensional subspaces
 \cite{paty2019subspace, lin2021projection} 
 to provide more robust estimation of those OT metrics.
 \lc{Although} effective as \lc{proxies} for WD, those methods do not provide a transport plan in the original space $\R^d$. To overcome this limitation, \cite{muzellec2019subspace} aims \lc{to compute} transport plans in a subspace which are extrapolated \lc{to} the original space.

\noindent\textbf{Pivot measure-based OT.} Other research works 
rely on a pivot, yet intermediate measure. They decompose the OT metric into \lc{Wasserstein} distances between each input measure and the considered pivot measure. 
They exhibit better properties such as statistical sample complexity or \lc{computational efficiency} \cite{forrow2019statistical,wang2013linear}. Even though the OT problems are split, they are still expensive when dealing with large sample size distributions, notably when only two distributions are involved. 

\noindent\textbf{Contributions.} \looseness=-1 We introduce a new proxy \lc{for} the squared WD that exploits the principles of \lc{aforementioned} approximations of OT metric. The original idea is to rely on projections and one-dimensional
assignment 
of the projected distributions to compute the new proxy. The approach is well-grounded as it hinges on the notion of Wasserstein generalized geodesics \cite{ambrosio2005gradient} with pivot measure supported on a line. The main features of the method are as:
i) its computational complexity is on par with SW, ii) it provides an optimal transport plan through the 1D assignment problem, iii) it acts as an upper bound of WD, and iv) is amenable to optimization to find the optimal pivot measure. As an additional contribution, we establish a closed form \lc{of} the WD when an input measure is supported on a line. 

\noindent\textbf{Outline.}  Section \ref{sec:background} presents some background of OT. Section \ref{sec:swgg v0} 
formulates our new WD proxy, provides some of its topological properties and a numerical computation scheme.
Section \ref{sec:swgg v1} builds upon 
the \lc{concept} of Wasserstein generalized geodesics to reformulate our OT metric approximation as the Sliced Wasserstein Generalized Geodesics (SWGG) along its optimal variant coined $\mswgg$.
\lc{This} reformulation allows deriving additional topological properties and an optimization scheme. Finally, Section \ref{sec:experimentation} provides experimental evaluations.

\noindent\textbf{Notations.} 
Let $\langle \cdot,\cdot \rangle$ be the Euclidean inner product on $\R^d$ and let $\Sd =\{ \boldsymbol{u} \in \R^d \text{ s.t. } \|\boldsymbol{u}\|_2=1\}$, the unit sphere. We denote $\Pall$ the set of probability measures on $\R^d$ endowed with the $\sigma-$algebra of Borel set and $\Pd \subset \Pall$ those with finite second-order moment i.e. $\Pd = \{\mu \in \mathcal{P}(\R^d) \text{ s.t. } \int_{\R^d} \|\boldsymbol{x}\|^2_2d\mu(\boldsymbol{x}) < \infty \}$. Let $\Pdn$ be the subspace of $\Pd$ defined by empirical measures with $n$-atoms and uniform masses. For any measurable function $f:\R^d \to \R^d$, we denote $f_\#$ its push forward, namely for $\mu \in \Pd$ and for any measurable set $A \in \R^d$, $f_\#\mu(A)=\mu(f^{-1}(A))$, with $f^{-1}(A)=\{\boldsymbol{x} \in \R^d \text{ s.t. } f(\boldsymbol{x}) \in A\}$.

\section{Background on Optimal Transport}
\label{sec:background}
 
\begin{definition}[Wasserstein distance]
\label{def:wd}
The squared WD \cite{villani2009optimal} between $\mu_1, \mu_2\in \Pd$ 
is defined as:
\begin{equation}
\label{eq:W2-distance}
    W^2_2(\mu_1,\mu_2)\defeq\inf_{\pi \in \Pi(\mu_1,\mu_2)}\int_{\R^d \times \R^d}\|\boldsymbol{x}-\boldsymbol{y}\|_2^2d\pi(\boldsymbol{x},\boldsymbol{y})
\end{equation}
with $\Pi(\mu_1,\mu_2) = \{\pi \in \mathcal{P}_2(\R^d \times \R^d)$ s.t. $\pi(A \times \R^d)=\mu_1(A)$ and $\pi(\R^d \times A)=\mu_2(A)$, $\forall A$ measurable set of $\R^d \}$.
\end{definition}


The  $\arg \min$ of Eq. \eqref{eq:W2-distance} is \lc{referred to as} the optimal transport plan. Denoted $\pi^*$, it  expresses how to move the probability mass from $\mu_1$ to  $\mu_2$ with minimum cost. 
In some cases, 
$\pi^*$ is 
of the form $(Id,T)_\#\mu_1$ for a measurable map $T:\R^d\to\R^d$, \textit{i.e.} there is no mass splitting during the transport. This map is called a Monge map and is denoted $T^{\mu_1 \to \mu_2}$ (or \lc{shortly} $T^{1 \to 2}$). Thus, one has $W^2_2(\mu_1,\mu_2)=\inf_{T \; \text{s.t.}\; T_\#\mu_1=\mu_2}\int_{\R^d}\|\boldsymbol{x}-T(\boldsymbol{x})\|^2_2d\mu_1(\boldsymbol{x})$. This occurs, for instance, when $\mu_1$ has a density w.r.t. the Lebesgue measure \cite{brenier1991polar} or when $\mu_1$ and $\mu_2$ are in $\Pdn$  \cite{santambrogio2015optimal}. 


Endowed with the WD, the space $\Pd$ is a geodesic space. Indeed, since there exists a Monge map $T^{ 1\to 2}$ between $\mu_1$ and $\mu_2$, one can define a geodesic curve $\mu^{1\to 2}:[0,1]\to \Pd$ 
\cite{gangbo1996geometry} as: 
\begin{equation}
    \forall t\in [0,1],\ \geodesic\defeq(tT^{1\to 2}+(1-t)Id)_\#\mu_1
    \label{eq:geodesic}
\end{equation}
which represents the shortest path w.r.t. Wasserstein distance in $\Pd$ between $\mu_1$ and $\mu_2$. The Wasserstein mean between $\mu_1$ and $\mu_2$ corresponds to $t=0.5$ and we simply write 
$\bary$. 



This notion of geodesic allows the study of the curvature of the Wasserstein space \cite{alexandrov1951theorem}. Indeed, 
the Wasserstein space is of positive curvature \cite{otto2001geometry}, \textit{i.e.} it respects the following inequality:
\begin{equation}
    W^2_2(\mu_1,\mu_2) \geq 2W^2_2(\mu_1,\nu)+2W^2_2(\nu,\mu_2)-4W^2_2(\mu^{1\to 2},\nu)
    \label{eq:inequality positive curvature}
\end{equation}
for all pivot measures $\nu \in \Pd$.

\paragraph{Solving and approximating Optimal Transport.}
The Wasserstein distance between empirical measures $\mu_1,\mu_2$ with $n$-atoms can be computed in  $\mathcal{O}(n^3\log n)$, preventing from the use of OT for large scale applications \cite{bonneel2011displacement}.
Several algorithms have been proposed to lower this complexity, for example the Sinkhorn algorithm \cite{cuturi2013sinkhorn} that provides an approximation in near $\mathcal{O}(n^2)$ complexity \cite{altschuler2017near}.

Notably, when $\mu_1=\frac{1}{n}\sum_{i=1}^n \delta_{x_i}$ and $\mu_2=\frac{1}{n}\sum_{i=1}^n \delta_{y_i}$ are 1D distributions, 
computing the WD can be done by matching the sorted empirical samples, leading to an overall complexity of $\mathcal{O}(n\log n)$. More precisely, let 
$\sigma$ and $\tau$ two permutation operators s.t. $x_{\sigma(1)}\leq x_{\sigma(2)} \leq...\leq x_{\sigma(n)}$ and $y_{\tau(1)}\leq y_{\tau(2)} \leq...\leq y_{\tau(n)}$.  Then, the 1D Wasserstein distance is given by:
\begin{equation}
    W^2_2(\mu_1,\mu_2)=\frac{1}{n}\sum_{i=1}^n(x_{\sigma(i)}-y_{\tau(i)})^2.
    \label{eq:closed-form-1D-WD}
\end{equation}



\paragraph{Sliced WD.}

The Sliced-Wasserstein distance (SWD)  \cite{Rabin-SWD-2011} aims to scale up the computation of OT by \lc{leveraging} the closed form expression \eqref{eq:closed-form-1D-WD} of the Wasserstein distance for 1D distributions. It is defined as the expectation of 1D-WD computed along projection directions $\theta \in \Sd$ over the unit sphere:
\begin{equation}
    \sw_2^2(\mu_1, \mu_2) \defeq \int_{\Sd} W^2_2(\thetap\mu_1,\thetap\mu_2)d\omega(\theta),
    \label{eq:swd}
\end{equation}
where $\thetap \mu_1$ and $\thetap \mu_2$ are projections onto the direction $\theta \in \Sd$  with  $P^{\theta} : \R^d\to\R$, $\boldsymbol{x} \mapsto \langle \boldsymbol{x},\theta\rangle$ and where $\omega$ is the uniform distribution over $\Sd$.

Since the integral in Eq. \eqref{eq:swd} is intractable, one resorts, in practice, to Monte-Carlo estimation to approximate the SWD. 

Its computation only involves projections and permutations. For $L$ directions, the 
computational complexity is $\mathcal{O}(dLn + Ln\log n)$ and the memory complexity is $\mathcal{O}(Ld+Ln)$. However, in high dimension, several 
projections are necessary to approximate accurately the SWD and many projections lead to 1D-WD close to 0. This issue is well known in the SW community \cite{zhang2023projection}, where different ways of performing effective sampling have been proposed \cite{nguyen2021distributional, nadjahi2021fast, nguyen2022hierarchical} such as distributional or hierarchical slicing. In particular, this motivates the definition of max-Sliced-Wasserstein \cite{deshpande2019max} which keeps only the most informative slice:
\begin{equation}
    \maxsw_2^2(\mu_1, \mu_2) \defeq\max_{\theta \in \Sd}W^2_2(\thetap\mu_1,\thetap\mu_2).
\end{equation} 
While being a non convex problem, it can be optimized efficiently using a gradient ascent scheme.

The SW-like distances are attractive since they are fast to compute and enjoy theoretical  properties: they are proper metrics and metricize the weak convergence. However, they  do not provide an OT plan.

\paragraph{Projected WD.} 

\lc{Another} quantity of interest based on the 1D-WD is the projected Wasserstein distance (PWD) \cite{rowland2019orthogonal}. It leverages the permutations of the projected distributions in 1D in order to derive couplings between the original distributions. 

Let $\mu_1=\frac{1}{n}\sum_{i=1}^n \delta_{\boldsymbol{x}_i}$ and $\mu_2=\frac{1}{n}\sum_{i=1}^n \delta_{\boldsymbol{y}_i}$ in $\Pdn$. The PWD is defined as: 
\begin{equation}
\label{eq:pwd}
    \pwd_2^2(\mu_1,\mu_2)\defeq\int_{\Sd} \frac{1}{n}\sum_{i=1}^n\|\boldsymbol{x}_{\sigma_\theta(i)}- \boldsymbol{y}_{\tau_\theta(i)}\|^2_2d\omega(\theta),
\end{equation}
where $\sigma_\theta,\tau_\theta$ are the permutations obtained by sorting $\thetap \mu_1$ and $\thetap\mu_2$.

As some permutations are not optimal, we straightforwardly have $W^2_2 \leq \pwd_2^2$. Note that some permutations can appear 
highly irrelevant in the original space, leading to an overestimation of $W^2_2$ (typically when the distributions are multi-modal or with support lying in a low \lc{dimensional} manifold, see Supp. \ref{app:PWD} for a discussion).

In this paper, we restrict ourselves to empirical distributions with the same number of samples. They are defined as $\mu_1 =\frac{1}{n}\sum_{i=1}^n \delta_{\boldsymbol{x}_i}$ and $\mu_2 =\frac{1}{n}\sum_{i=1}^n \delta_{\boldsymbol{y}_i}$ in $\Pdn$. Note that the results presented therein can be extended to any discrete measures by mainly using quantile functions instead of permutations and transport plans instead of transport maps (see Supp. \ref{app:quantile swgg}). 

\section{Definition and properties of min-SWGG}
\label{sec:swgg v0}

The fact that the $\pwd$ overestimates $W^2_2$ motivates the introduction of our new loss function coined $\mswgg$ which keeps only the most informative \lc{permutation}. Afterwards, we derive a property of distance and grant an estimation of $\mswgg$ via random search of the directions.

\begin{definition}[SWGG and min-SWGG]
\label{def:swgg permutation}
Let $\mu_1,\mu_2\in\Pdn$ and $\theta\in \Sd$. Denote by $\sigma_\theta$ and $\tau_\theta$ the permutations obtained by sorting the 1D projections $\thetap\mu_1$ and $\thetap\mu_2$. We define respectively $\swgg$ and $\mswgg$ as:
\begin{align}
        \swgg_2^2(\mu_1,\mu_2,\theta) &\defeq\frac{1}{n} \sum_{i=1}^n \|\boldsymbol{x}_{\sigma_\theta(i)}-\boldsymbol{y}_{\tau_\theta(i)}\|^2_2, \\
        \mswgg_2^2(\mu_1,\mu_2) &\defeq \min_{\theta \in \Sd}\swgg_2^2(\mu_1,\mu_2,\theta).
        \label{eq:swgg_1}
\end{align}

\end{definition}
One shall remark that the function $\swgg$ corresponds to the building block of PWD in eq. \eqref{eq:pwd}.

One main feature of $\mswgg$ is that it comes with a transport map. 
Let $\theta^* \in \argmin~ \swgg_2^2(\mu_1,\mu_2,\theta)$ be the optimal projection direction. The associated transport map is:
\begin{equation}
T(\boldsymbol{x}_i) = \boldsymbol{y}_{\tau_{\theta^*}^{-1}(\sigma_{\theta^*}(i))}, \quad \forall 1 \leq i \leq n.
\end{equation}

In Supp. \ref{app:example transport plan} we give several examples of such transport plan. These examples show that the overall structure of the optimal transport plan is respected by the transport plan obtained via $\mswgg$.

We now give some theoretical properties of the quantities $\mswgg$ and $\swgg$. Their proofs are given 
in Supp. \ref{app:distance swgg cp}.

\begin{proposition}[Distance and Upper bound]
\label{prop:dist upper cp}
Let $\theta \in \Sd$. $\swggr(\cdot,\cdot,\theta)$ defines a distance on $\Pdn$. Moreover, $\mswgg$ is an upper bound of $W^2_2$, and  $W^2_2\leq \mswgg_2^2 \leq \pwd_2^2$, with equality between $W_2^2$ and $\mswgg^2_2$ when $d>2n$. 
\end{proposition}

\begin{remark}
Similarly to max-SW, $\mswgg$ retains only one optimal direction $\theta^* \in \Sd$. However, the two distances strongly differ: i) $\mswgg$ is an upper bound and max-SW a lower bound of $W^2_2$, ii) the optimal $\theta^*$ \lc{may differ} (see Supp. \ref{app:diff mswgg and maxsw} for an illustration), and iii) max-SW does not provide a transport plan between $\mu_1$ and $\mu_2$.
\end{remark}

Solving Eq. \eqref{eq:swgg_1} can be achieved using a random search, by sampling $L$ directions $\theta \in \Sd$ and keeping only the one leading to the lowest value of $\swgg$. 

This gives an overall computational complexity of $\mathcal{O}(Ldn+Ln\log n)$ and a memory complexity of $\mathcal{O}(dn)$. In low dimension, the random search estimation is effective: covering all possible
permutations through $\Sd$ can be done with a low number of directions. In high dimension, many more directions $\theta$ are needed to have a relevant approximation, typically $\mathcal{O}(L^{d-1})$. This motivates the 
design
of gradient descent techniques for finding $\theta^*$.

\section{SWGG as minimizing along the Wasserstein generalized geodesics}
\label{sec:swgg v1}

\begin{wrapfigure}{R}{0.5\textwidth}
    \vskip -.3in
    \centering
    \includegraphics[width=0.24\textwidth]{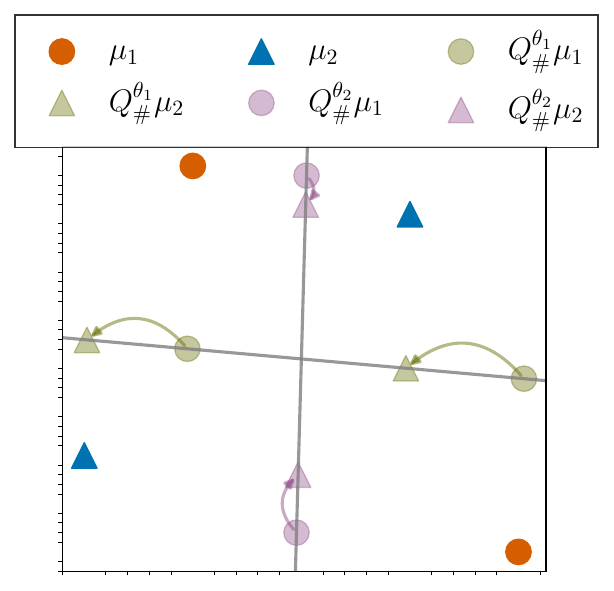}
    \hfill
    \includegraphics[width=0.24\textwidth]{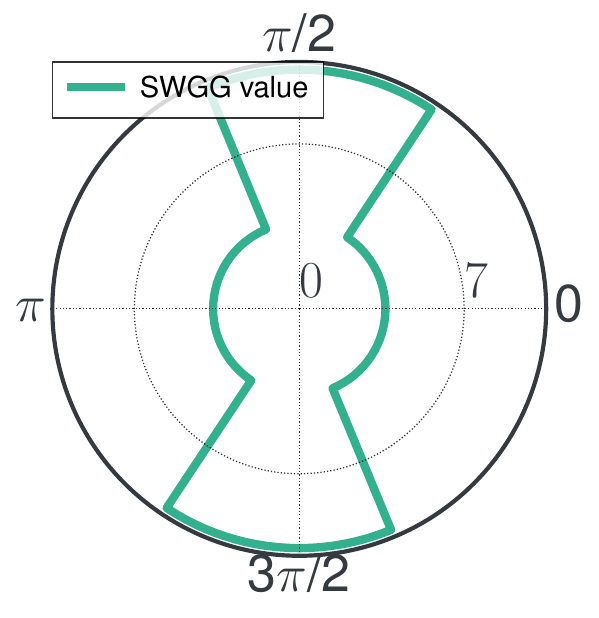}

    \caption{(Left) Empirical distributions with examples of 2 sampled lines (Right) that lead 
    to 2 possible values of $\swgg$ when $\theta \in [0, 2\pi]$.}
    \label{fig:non continuity of swgg}

    \vskip -.2in
\end{wrapfigure}

Solving problem in Eq. \eqref{eq:swgg_1} 
amounts
to optimize 
over a set of admissible permutations. \lc{This} problem is 
hard 
since $\swgg$ is  non convex w.r.t. $\theta$ and \lc{piecewise} constant,
thus not differentiable over $\Sd$. Indeed, as long as the permutations remain the same for different directions $\theta$,
the value of $\swgg$ \lc{remains} constant. \lc{When} the permutations change, the objective SWGG "jumps" as illustrated in Fig.~\ref{fig:non continuity of swgg}. 
 
In this section, we tackle this problem by providing an alternative formulation of  $\mswgg$ 
that allows smoothing 
the different kinks of $\swgg$, hence, making $\mswgg$ 
amenable to optimization. 
This formulation relies on Wasserstein generalized geodesics we introduce hereinafter.

\looseness=-1 We show that this alternative formulation 
brings in
computational advantages  
and allows establishing some additional topological properties and deriving an efficient optimization scheme. We also provide a new closed form expression of the Wasserstein distance $W^2_2(\mu_1,\mu_2)$ when either $\mu_1$ or $\mu_2$ is supported on a line.

\subsection{SWGG based on Wasserstein Generalized Geodesics}
\label{sec:definition and computation gg}
 
Wasserstein generalized geodesics (see Supp. \ref{app:generalized geodesic} for more details) were first introduced in \cite{ambrosio2005gradient} in order to ensure the convergence of Euler scheme for Wasserstein Gradient Flows. This concept has been used notably in \cite{forrow2019statistical, muzellec2019subspace} to speed up some computations and to derive some 
theoretical properties. Generalized geodesic is also highly related with the idea of linearization of the Wasserstein distance via an $L^2$ space \cite{wang2013linear,moosmuller2020linear}, see Supp. \ref{app:related works gg} for more details on the related works.

Generalized geodesics lay down on a pivot measure $\nu \in \Pdn$ to transport the distribution $\mu_1$ toward $\mu_2$. Indeed, one can leverage the optimal transport maps $T^{\nu \to \mu_1}$ and $T^{\nu \to \mu_2}$ to construct a curve $t\mapsto \genegeodesic$ linking $\mu_1$ to $\mu_2$ as 
\begin{equation}
    \genegeodesic \defeq \left((1-t)T^{\nu \to \mu_1} + tT^{\nu \to \mu_2}\right)_\# \nu, \quad \quad \forall t \in[0,1].
    \label{eq:gengeo}
\end{equation}
The related generalized Wasserstein mean  corresponds to $t=0.5$ and is denoted $\gbarycenter$.

Intuitively, the optimal transport maps between $\nu$ and $\mu_i, i=1,2$ give rise to a sub-optimal transport map between $\mu_1$ and $\mu_2$:
\begin{equation}
    T_\nu^{1\to 2} \defeq T^{\nu \to \mu_2} \circ T^{\mu_1 \to \nu} \quad \text{with} \quad (T_\nu^{1\to 2})_\#\mu_1=\mu_2.
    \label{eq:composition transport plan generalized geodesic}
\end{equation}
One can be interested in the cost induced 
by the transportation of $\mu_1$ to $\mu_2$ via the transport map $T_\nu^{1\to 2}$, known as the $\nu$-based Wasserstein distance \cite{nenna2023transport} and defined as
\begin{equation}
    W^2_\nu(\mu_1,\mu_2)\defeq\int_{\R^d} \|\boldsymbol{x}-T_\nu^{1\to 2}(\boldsymbol{x})\|^2_2d\mu_1(\boldsymbol{x}) =2W^2_2(\mu_1,\nu)+2W^2_2(\nu,\mu_2)-4W^2_2(\gbarycenter,\nu).
    \label{eq:v-based Wasserstein def}
\end{equation}

Notably, the second part of Eq. \eqref{eq:v-based Wasserstein def} 
straddles the square Wasserstein distance with Eq. \eqref{eq:inequality positive curvature}.
Remarkably, the computation of $W^2_\nu$ can be efficient if the pivot measure $\nu$ is chosen \lc{appropriately}. As established in Lemma \ref{lemma:closed form}, it is the case when $\nu$ is supported on a line. Based on these facts, we propose hereafter an alternative formulation of $\swgg$.

\begin{definition}[Pivot measure]
\label{def:pivot measure}
Let $\mu_1$ and $\mu_2 \in \Pdn$. We \lc{restrict} the pivot measure $\nu$ to be the Wasserstein mean of the measures $\thetapp \mu_1$ and $\thetapp \mu_2$:
    $$
    \pivot\defeq\arg \min_{ \mu \in \Pdn}W^2_2(\thetapp \mu_1,\mu)+W^2_2(\mu,\thetapp \mu_2),
    $$ 
    where $\theta \in \Sd$ and $Q^\theta : \R^d \to \R^d$, $\boldsymbol{x} \mapsto \theta \langle \boldsymbol{x},\theta \rangle$ is the projection onto the subspace generated by $\theta$. Moreover $\pivot$ is always defined as the middle of a geodesic as in Eq \eqref{eq:geodesic}.
\end{definition}

One shall notice that $\thetapp \mu_1$ and $\thetapp \mu_2$ are supported on the line defined by the direction $\theta$, so is the pivot measure $\nu = \pivot$. We are now ready to \lc{reformulate} the metric $\swgg$.

\begin{proposition}[SWGG based on generalized geodesics] \label{def:swgg generalized geodesic}
    Let $\theta \in \Sd$, $\mu_1,\mu_2 \in \Pdn$ and $\pivot$ be the pivot measure. 
    Let $\lgbarycenter$ be the generalized Wasserstein mean between $\mu_1$ and $\mu_2 \in \Pdn$ with pivot measure $\pivot$. Then,
    \begin{equation} \label{eq:defswgg}
        \swgg_2^2(\mu_1,\mu_2,\theta) = 2W^2_2(\mu_1,\pivot)+2W_2^2(\pivot,\mu_2)-4W_2^2(\lgbarycenter,\pivot).
    \end{equation}
\end{proposition}

The proof is in Supp.\ref{app:equivalence swgg}. From Proposition \ref{def:swgg generalized geodesic}, $\swgg$ is the $\pivot$-based Wasserstein distance between $\mu_1$ and $\mu_2$. This alternative formulation allows establishing additional properties of $\mswgg$.

\subsection{Theoretical properties }
\label{sec:properties gg}

Additionally to the properties derived in Section \ref{sec:swgg v0} ($\swgg$ is a distance and $\mswgg$ is an upper bound of $W^2_2$), we provide below other theoretical guarantees. 

\begin{proposition}[Weak Convergence]
\label{prop:weak convergence}
$\mswgg$ metricizes the weak convergence in $\Pdn$. In other words, let $(\mu_{k})_{k \in \mathbb{N}}$ be a sequence of measures in $\Pdn$ and $\mu \in \Pdn$. We have:
\begin{align}
    \mu_{k} \underset{k}{\towd} \mu \iff  \mswgg_2^2(\mu_{k},\mu)\tok 0 \nonumber,
\end{align}
where $\towd$ stands for the weak convergence of measure i.e. $\int_{\R^d}fd\mu_{k}\to \int_{\R^d}fd\mu$ for all continuous bounded functions $f$.
\end{proposition}

Beyond the weak convergence, $\mswgg$ possesses the translation property, \textit{i.e.} the translations can be factored out as  the Wasserstein distance does (see \cite[remark 2.19]{COTFNT} for a recall).

\begin{proposition}[Translation]
\label{proposition translation}
Let $T^u$ (resp. $T^v$) be the map $\boldsymbol{x} \mapsto \boldsymbol{x}-\boldsymbol{u}$ (resp. $\boldsymbol{x} \mapsto \boldsymbol{x}-\boldsymbol{v}$), with $\boldsymbol{u},\boldsymbol{v}$ vectors of $\R^d$. We have:
\begin{align*}
\mswgg_2^2(T^u_\#\mu_1,T^v_\#\mu_2)=&~\mswgg_2^2(\mu_1,\mu_2)+\|\boldsymbol{u}-\boldsymbol{v}\|_2^2-2\langle \boldsymbol{u}-\boldsymbol{v},\boldsymbol{m_1}-\boldsymbol{m_2}\rangle
\end{align*}
where $\boldsymbol{m_1}=\int_{\R^d} \boldsymbol{x} d\mu_1(\boldsymbol{x})$ and $\boldsymbol{m_2}=\int_{\R^d} \boldsymbol{x} d\mu_2(\boldsymbol{x})$ are the means of $\mu_1$, $\mu_2$.
\end{proposition}
This property is useful in some applications such as shape matching, in which translation invariances are sought.

The proofs of the two Propositions are deferred to Supp. \ref{proof prop weak convergence} and \ref{proof prop translation}.

\begin{remark}[Equality] 
$\mswgg$ and $W^2_2$ are equal in different cases. First, \cite{moosmuller2020linear} showed that it is the case whenever $\mu_1$ is the shift and scaling of $\mu_2$ (see Supp. \ref{app:shift and scaling} for a full discussion). 
In Lemma  \ref{lemma:closed form}, we will state that it is also the case if 
one of the two distributions is supported on a line. 
\end{remark}

\subsection{Efficient computation of SWGG}
\label{sec:efficient computation + optimization scheme}
$\swgg$ defined in Eq.~\eqref{eq:defswgg} involves computing three WDs that are fast to compute, with an overall $\mathcal{O}(dn+n\log n)$ complexity, as detailed below. Building on this result, we provide an optimization scheme that allows optimizing over $\theta$ with $\mathcal{O}(sdn+sn\log sn)$ operations at each iteration, with $s$ a (small) integer. We first start by giving a new closed form expression of the WD whenever one distribution is supported on a line, that proves useful \lc{for deriving} an efficient computation scheme.

\paragraph{New closed form of the WD.}
\label{para:closed formed}
The following lemma states that 
$W^2_2(\mu_1,\mu_2)$ admits a closed form whenever $\mu_2$ is supported on a line. 
 
\lc{This lemma leverages} the computation of the WD between $\mu_2$ and the orthogonal projection of $\mu_1$ onto the linear subspace defined by the line.
Additionally, \lc{it provides an explicit formulation for} the optimal transport map $T^{1 \to 2}$.
\begin{lemma}
\label{lemma:closed form}
    Let $\mu_1,\mu_2$ in $\Pdn$ with $\mu_2$ supported on a line of direction  $\theta \in \Sd$. We have:
    \begin{equation}
        W^2_2(\mu_1,\mu_2)=W^2_2(\mu_1,Q_\#^\theta\mu_1)+W^2_2(Q_\#^\theta\mu_1,\mu_2)
    \end{equation}
with  $Q^\theta$ as in Def. \ref{def:pivot measure}.
Note that $W^2_2(\mu_1,Q^\theta_\# \mu_1)=\frac{1}{n}\sum \|\boldsymbol{x}_i-Q^\theta(\boldsymbol{x}_i)\|^2_2$ and  $W^2_2(Q_\#^\theta\mu_1,\mu_2)=W^2_2(P_\#^\theta\mu_1,P_\#^\theta\mu_2)$ are the WD between 1D distributions.
Additionally, the optimal transport map is given by $T^{1\to 2} =  T^{Q^\theta_\# \mu_1 \to \mu_2} \circ T^{\mu_1 \to Q^\theta_\# \mu_1}=T^{Q^\theta_\# \mu_1 \to \mu_2}\circ Q^\theta$. In particular, the map $T^{1\to 2}$ can be obtained via the permutations of the 1D distributions $P^\theta_\#\mu_1$ and $P^\theta_\#\mu_2$. The proof is \lc{provided} in Supp. \ref{app:proof lemma closed form}. 
\end{lemma}

\paragraph{Efficient computation of $\swgg$.}
\label{para:efficient computation}
Eq. \eqref{eq:defswgg} is defined as the Wasserstein distance between a distribution (either $\mu_1$ or $\mu_2$ or $\lgbarycenter)$ and a distribution supported on a line ($\pivot$). As detailed in Supp. \ref{app:fast computation}, computation of Eq. \eqref{eq:defswgg} \lc{involves} three Wasserstein distances between distributions and their projections: i) $W^2_2(\mu_1,Q^\theta_\#\mu_1)$, ii) $W^2_2(\mu_2,Q^\theta_\#\mu_2)$, iii) $W^2_2(\lgbarycenter,\pivot)$,  and \lc{a} one dimensional Wasserstein distance $W_2^2(P^\theta_\# \mu_1,P^\theta_\# \mu_2)$, resulting in a $\mathcal{O}(dn+n\log n)$ complexity.

\paragraph{Optimization scheme for min-SWGG.} 
\label{sec:optimization scheme}
 
The term $W_2^2(\lgbarycenter,\pivot)$ in Eq. \eqref{eq:defswgg} is \lc{not} continuous w.r.t. $\theta$. Indeed, the generalized mean $\lgbarycenter$ depends \lc{only} on the transport maps $T^{\pivot \to \mu_1}$ and $T^{\pivot \to \mu_2}$, \lc{which remain} constant as long as 
different projection directions $\theta$ lead to the same permutations $\sigma_\theta$ and $\tau_\theta$. Hence, we rely on a smooth surrogate $\widetilde{\lgbarycenter}$ of the generalized mean and we aim \lc{to minimize} the following objective function:
\begin{equation} 
\label{eq:defswggsmooth}
\widetilde{\swgg_2^2}({\mu}_1,{\mu}_2,\theta)\defeq  2W^2_2(\mu_1,\pivot) 
 +2W_2^2(\pivot,\mu_2)  -4{W_2^2}(\widetilde{\lgbarycenter},\pivot).
\end{equation}

To define $\widetilde{\lgbarycenter}$, one option would be to use entropic maps in Eq. \eqref{eq:gengeo} but at the price of a quadratic time complexity. 
We rather build upon the blurred Wasserstein distance \cite{feydy2020geometric} to define $\widetilde{\lgbarycenter}$ as it can be seen as an efficient surrogate of entropic transport plans in 1D. In 
one dimensional setting, $\widetilde{\lgbarycenter}$ can be  approximated \lc{efficiently} by adding an empirical Gaussian noise followed by a sorting pass. In our case, it resorts in making $s$ copies of each sorted \lc{projection} $P^\theta(\boldsymbol{x}_{\sigma(i)})$ and $P^\theta (\boldsymbol{y}_{\tau(i)})$ respectively, to add an empirical Gaussian noise of deviation $\sqrt{\epsilon}/2$ and to compute averages of sorted blurred copies $\boldsymbol{x}^s_{\sigma^s}$, $\boldsymbol{y}^s_{\tau^s}$. We finally have $(\widetilde{\lgbarycenter})_i = \frac{1}{2s} \sum_{k=(i-1)s+1}^{is} \boldsymbol{x}^s_{\sigma^s(k)}+\boldsymbol{y}^s_{\tau^s(k)}$. 
\cite{feydy2020geometric} showed that this blurred WD has the same asymptotic properties as the Sinkhorn divergence.

\begin{figure}[ht]
    \centering
    \vskip -0.1in
    \includegraphics[scale=0.3]{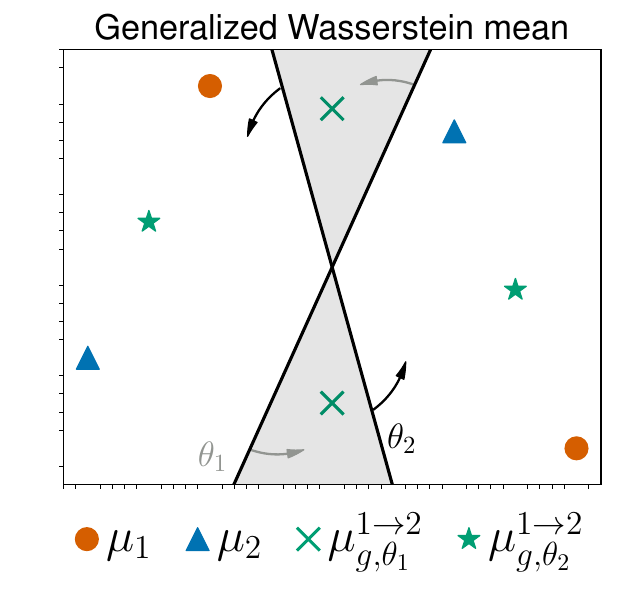}
    \includegraphics[scale=0.3]{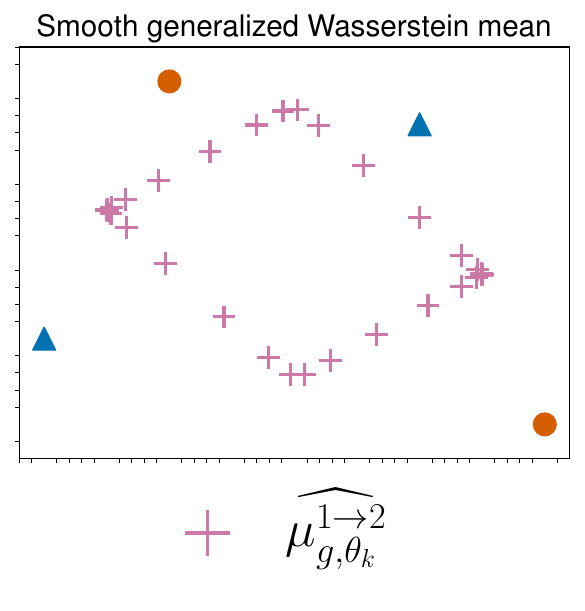}
    \includegraphics[scale=0.3]{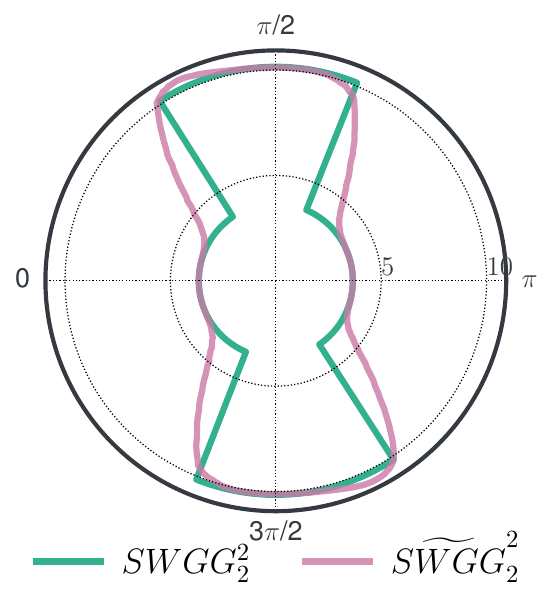}
    \vskip -0.1in
    \caption{Illustration of the smoothing effect in the same setting as in Fig. \ref{fig:non continuity of swgg}. (Left) Two sets of generalized Wasserstein means are possible, depending on the direction of the sampled line w.r.t. $\theta_1$ and $\theta_2$, giving rise to 2 different values for $\swgg$. (Middle) The surrogate provides a smooth transition between the two sets of generalized Wasserstein means as the direction $\theta$ changes, (Right) providing a smooth approximation of $\swgg$ that is amenable to optimization.}
    \label{fig:smooth swgg}
    \vskip -0.2in
\end{figure}

The surrogate $\widetilde{\swgg}({\mu}_1,{\mu}_2,\theta)$ is smoother w.r.t. $\theta$ and can thus be optimized \lc{using} gradient descent, converging towards a local minima. Once the optimal direction $\theta^*$ is found, $\mswgg$ resorts to be the solution provided by $\swgg(\mu_1,\mu_2,\theta^*)$. Fig. \ref{fig:smooth swgg} illustrates the effect of the smoothing on a 
toy example 
and more details are given in Supp. \ref{app:smooth SWGG}. The computation of  $\widetilde{\swgg}({\mu}_1,{\mu}_2,\theta)$ is summarized \lc{in} Alg. \ref{alg:computation swgg}.

\begin{algorithm}[ht]
\caption{Computing $\widetilde{\swgg_2^2}${$({\mu}_1,{\mu}_2,\theta)$}}
\label{alg:computation swgg}
\begin{algorithmic}
\Require $\mu_1=\frac{1}{n}\sum \delta_{\boldsymbol{x}_i}$, $\mu_2=\frac{1}{n}\sum \delta_{\boldsymbol{y}_i}$,  $\theta \in \Sd$, $s \in \mathbb{N}_+$  and $\epsilon \in \R_+$

\State ${\sigma}, {\tau} \gets$ ascending ordering of $(P^\theta(\boldsymbol{x}_i))_i$, $(Q^\theta(\boldsymbol{y}_i))_i$

\State $\boldsymbol{x}^s \gets s$ copies of $(\boldsymbol{x}_{\sigma(i)})_i$, $\boldsymbol{y}^s \gets s$ copies of $(\boldsymbol{y}_{\tau(i)})_i$

\State ${\sigma^s}, {\tau^s} \gets$ ascending ordering of $\langle \boldsymbol{x}^s, \theta\rangle + \boldsymbol{\xi}$, $\langle \boldsymbol{y}^s, \theta\rangle +\boldsymbol{\xi}$  for  $  \xi_i \sim
\mathcal{N}(0, \epsilon/2)$, $\forall i \leq sn$

\State 
\State $a \gets \frac{2}{n}\sum_i \left(\bigl \|\boldsymbol{x}_i- Q^\theta(\boldsymbol{x}_i) \|^2_2 + \|\boldsymbol{y}_i- Q^\theta(\boldsymbol{y}_i) \|^2_2\right)$ 
\Comment{\emph{$2W^2_2(\mu_1,\thetapp \mu_1) + 2W^2_2(\mu_2,\thetapp \mu_2)$}}

\State $b \gets \frac{2}{n}  \sum_i\bigl \|P^\theta (\boldsymbol{x}_{\sigma(i)})+P^\theta(\boldsymbol{x}_{\tau(i)}) \bigr \|^2_2$  
\Comment{\emph{$2W^2_2(\thetap \mu_1,\thetap \mu_2)$}}

\State $c \gets \frac{4}{n} \sum\limits_i \bigl \| \frac{1}{2}(Q^\theta(\boldsymbol{x}_{\sigma(i)}) +Q^\theta(\boldsymbol{y}_{\tau(i)})) - \frac{1}{2s} \sum\limits_{k=(i-1)s+1}^{is} (\boldsymbol{x}^s_{\sigma^s(k)}+\boldsymbol{y}^s_{\tau^s(k)}) \bigr \|^2_2$
\Comment{\emph{$4W^2_2(\widetilde{\lgbarycenter},\pivot)$}}\\
\textbf{Output} $a+b-c$
\end{algorithmic}
\end{algorithm}

\section{Experiments}
\label{sec:experimentation}

We highlight that $\mswgg$ is fast to compute, gives an approximation of the WD 
and the associated transport plan. 
We start 
by comparing the random search and the gradient descent schemes for 
finding the optimal direction in subsection \ref{sec:monte-carlo vs optim}. Subsection \ref{sec:gradient flows} illustrates the weak convergence property of $\mswgg$ through a gradient flow application to match distributions. We then implement an efficient algorithm for colorization of gray scale images in \ref{sec:colorization}, thanks to the new closed form expression of the WD. We finally evaluate $\mswgg$ 
\lc{in a shape matching context} in subsection \ref{sec:ICP}. When possible from the context, we compare $\mswgg$ with the main methods for approximating the WD namely SW, max-SW, Sinkhorn \cite{cuturi2013sinkhorn}, factored coupling \cite{forrow2019statistical} and subspace robust WD (SRW) \cite{paty2019subspace}. 
Supp. \ref{app:experimentation} provides additional results on the behavior of $\mswgg$ and  experiments on other tasks such as color transfer or on data sets distance computation. All the code is available at \footnote{\url{https://github.com/MaheyG/SWGG}}

\subsection{Computing min-SWGG} 
\label{sec:monte-carlo vs optim}

\begin{figure}[!ht]
    \centering
    \vskip -0.1in
    \hskip -.4in
    \includegraphics[scale=0.5]{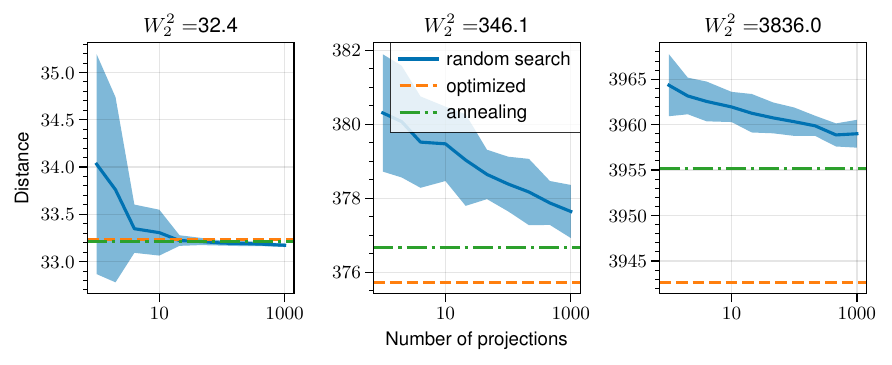}
    \hskip -.1in
    \includegraphics[scale=0.35]{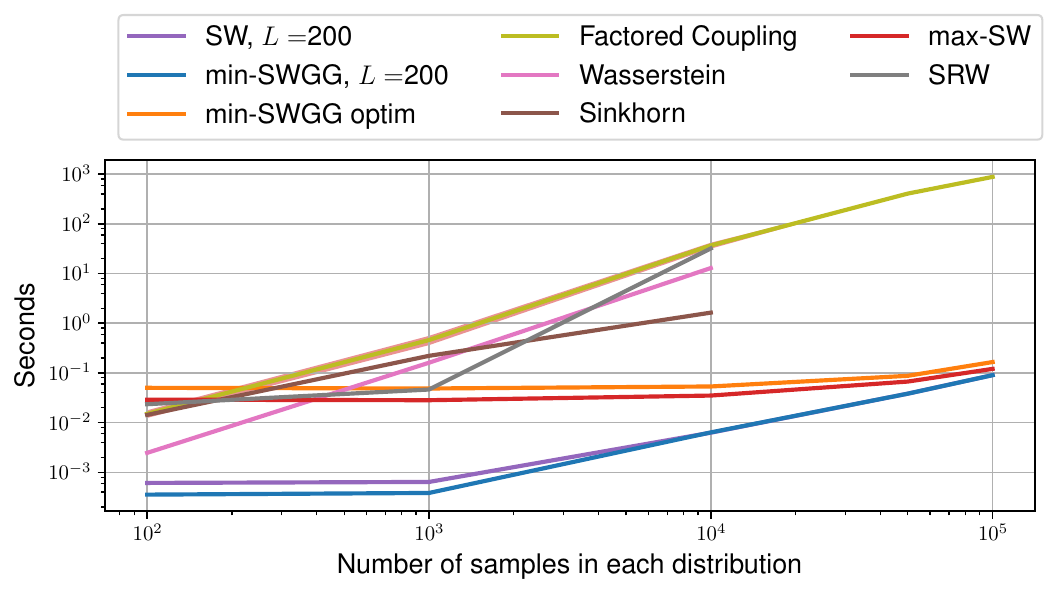}
    \vskip -0.1in
    \caption{(Left) evolution of $\mswgg$ with different numbers of projections and with the dimension $d$ in  $\{2, 20, 200\}$. (Right) Runtimes. }
    \label{fig:mc_vs_optim}
    \vskip -.1in
\end{figure}

Let consider Gaussian distributions in dimension\lc{s} $d \in \{2, 20, 200\}$. We first sample $n=1000$ points from each distribution to define $\mu_1$ and $\mu_2$. We then compute $\mswgg_2^2(\mu_1, \mu_2)$ computed using different schemes, either by random search, by simulated annealing \cite{pincus1970monte} 
or by gradient descent. We report the obtained results in Fig. \ref{fig:mc_vs_optim} (left). For the random search scheme, we repeat each experiment 20 times and we plot the average 
value of $\mswgg$ $\pm$ 2 times the standard deviation. 

For the gradient descent, we select a random initial $\theta$.
We observe that, in low dimension, 
all schemes provide similar values of $\mswgg$. When the dimension increases, optimizing the direction $\theta$ \lc{yields a more accurate approximation of the true Wasserstein distance} (see  plots' title in Fig. \ref{fig:mc_vs_optim}). On Fig. \ref{fig:mc_vs_optim} (right), we compare the empirical runtime evaluation for $\mswgg$ with different competitors for $d = 3$ and using $n$ samples from Gaussian distributions, with $n \in \{10^2, 10^3, 10^4, 5\times 10^4, 10^5\}$. We observe that, as expected, $\mswgg$ with random search is as fast as $\sw$ with a super linear time complexity. With the optimization process, it is faster than SRW for a given number of samples. We also note that SRW is more demanding in memory and hence does not scale as well as $\mswgg$. We give more details on this experimentation and a comparison with competitors in Supp. \ref{app:monte-carlo vs optim}. 

\subsection{Gradient Flows}
\label{sec:gradient flows}
\begin{figure}[ht] 
\vskip -0.1in
\centering
\includegraphics[scale=0.32]{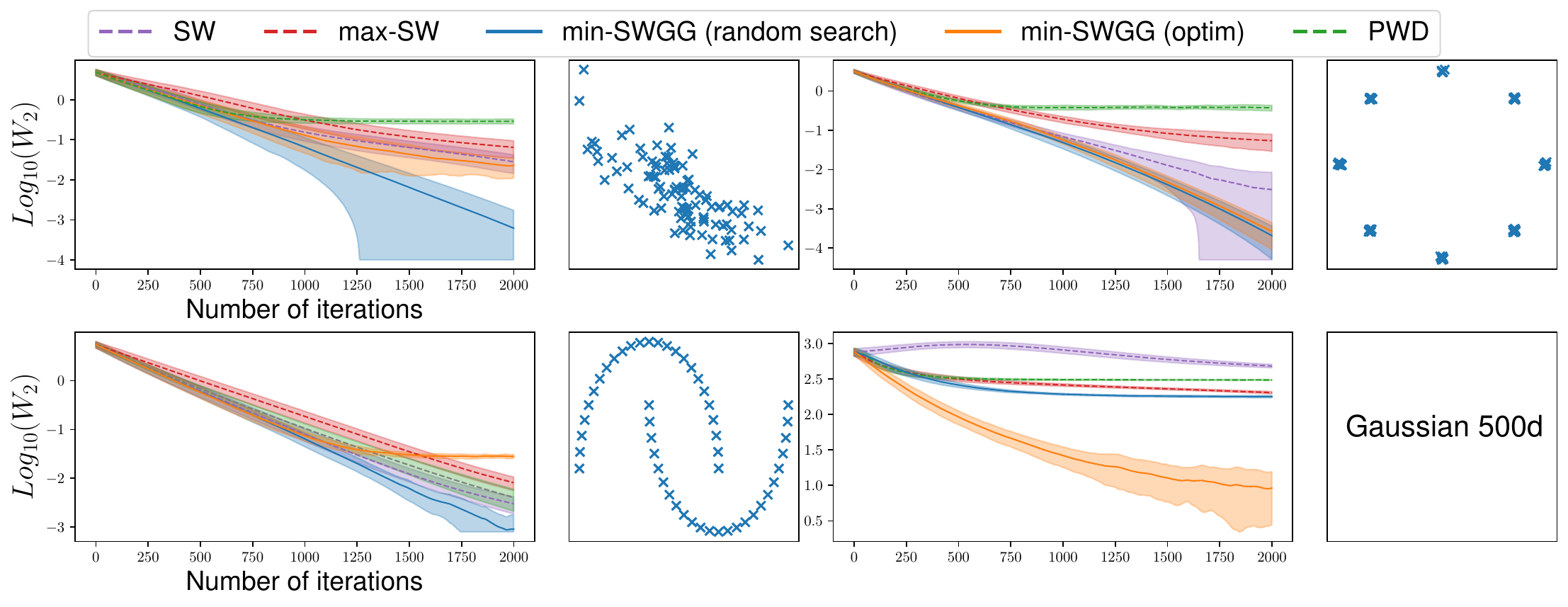}
\vskip -0.1in
\caption{Log of the WD 
between different source and target 
distributions as a function of the number of iterations.}
\label{fig:gradient flow}
\vskip -.22in
\end{figure}

We highlight the weak convergence property of $\mswgg$.
\lc{Initiating} from a random initial distribution, we aim \lc{to move} the particles of a source distribution $\mu_1$ \lc{towards} a target one $\mu_2$ by reducing the objective  $\mswgg_2^2(\mu_1,\mu_2)$ at each step. We compare both variants of $\mswgg$ \lc{against} $\sw$, $\maxsw$ and $\pwd$, relying on the code provided in \cite{kolouri2019generalized} for running the experiment; we report the results on Fig. \ref{fig:gradient flow}. We consider several target distributions, representing diverse scenarios and fix $n=100$. We run each experiment 10 times and report the mean $\pm$ the standard deviation. In every case, one can see that $\mu_1$ moves \lc{towards} $\mu_2$ and that all methods tend to have similar behavior. One can notice though that, for the distributions in $d=500$ dimensional space, $\mswgg$ computed with the optimization scheme leads to the best alignment  of the distributions. 

\subsection{Gray scale image colorization}
\label{sec:colorization}

\begin{figure}[ht]
    \centering
    \vskip -0.1in
    \begin{center}
        \includegraphics[scale=0.2]{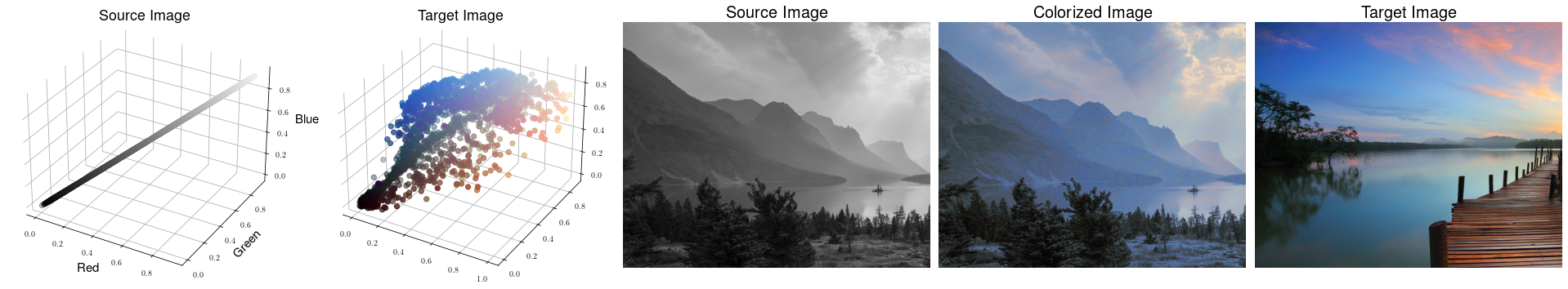}
    \end{center}
    \vskip -0.1in
    \caption{Cloud point source and target (left) colorization of image (right).}
    \vskip -0.15in
    \label{fig:colorization}
\end{figure}

Lemma \ref{lemma:closed form} states that the WD has a closed form when one of the 2 distributions is supported on a line, allowing us to compute the WD and the OT map with a complexity of $\mathcal{O}(dn+n\log n)$.
This particular situation arises for instance with RBG images ($\mu_1,\mu_2 \in \mathcal{P}_2^n(\R^3)$), where black and white images are supported on a line (the line of grays). One can 
address the problem of image colorization through color transfer \cite{ferradans2013regularized}, where a black and white image is the source and a colorful image the target. Our fast procedure allows 
considering large images without sub-sampling 
with a reasonable computation time. 
Fig. \ref{fig:colorization} gives an example of colorization of an image of size 1280$\times$1024 that was computed in less than 0.2 second, while being totally untractable for the $\mathcal{O}(n^3\log n)$ solver of WD.

This procedure can be lifted to pan-sharpening \cite{vivone2021benchmarking} where one aims \lc{to construct} a super-resolution multi-chromatic satellite image with the help of a super-resolution mono-chromatic image (source) and a low-resolution multi-chromatic image (target). Obtained results are given in the Supp. \ref{app:colorization and pan}.

\subsection{Point clouds registration}
\label{sec:ICP}
Iterative Closest Point (ICP) is an algorithm for aligning point clouds based on their geometries \cite{besl1992method}. Roughly, its most popular version defines a one-to-one correspondence between point clouds, computes a rigid transformation (namely translation, rotation or reflection), moves the source point clouds using the transformation, and iterates the procedure until convergence. The rigid transformation is the solution of the Procrustes problem \textit{i.e.} $\arg \min_{(\Omega,t) \in O(d)\times\R^d}\|\Omega (\boldsymbol{X}-t)-\boldsymbol{Y}\|_2^2$, where $\boldsymbol{X},\boldsymbol{Y}$ are the source and the target cloud points and $O(d)$ the space of orthogonal matrices of dimension $d$. This Procrustes problem can be solved using a SVD \cite{schonemann1966generalized} for instance.   

We perform the ICP algorithm with different variants to compute the one-to-one correspondence: neareast neighbor (NN) correspondence, OT transport map (for small size datasets) and  $\mswgg$ transport map. Note that SW, PWD, SRW, factored coupling and Sinkhorn cannot be run in this context where a one-to-one correspondence is mandatory; subspace detours \cite{muzellec2019subspace} are irrelevant in this context (see Supp. \ref{app:ICP}). 
We evaluate the results of the ICP algorithm 
in terms of: i) the quality of the final alignment, measured by the Sinkhorn divergence between the re-aligned and target point cloud; ii) the speed of the algorithm given by the running time until convergence. We consider 3 datasets of different sizes. The results are shown in Table \ref{table:ICP sinkhorn} and more details about the setup, can be found in Supp. \ref{app:ICP}. In Supp. \ref{app:ICP} we give a deeper analysis of the results, notably with different criteria for the final assignment, namely the Chamfer and the Frobenius distance. One can see that the assignment provided by OT-based methods is better than NN. $\mswgg$ allows working with large datasets, while OT fails to provide a solution for $n=150 000$.

\begin{table}[ht]
\parbox{.33\linewidth}{
\centering
\vskip -0.4in
\includegraphics[width=0.18\textwidth]{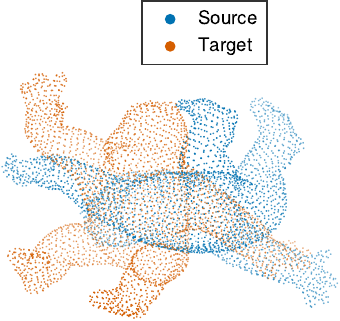}
   \label{table:ICP time}}
\parbox{.65\linewidth}{
\centering
    \begin{tabular}{l c c c c} 
    $n$ & 500 &3000 & 150 000\\
    \hline
    NN&3.54 (\textbf{0.02}) &96.9 (\textbf{0.30})&23.3 (\textbf{59.37})\\
    OT & 0.32 (0.18)&48.4 (58.46)& $\cdot$ \\
    $\mswgg$ & \textbf{0.05} (0.04) &\textbf{37.6} (0.90)&  \textbf{6.7} (105.75)\\
     \hline
    \end{tabular}
    \caption{Sinkhorn Divergence between final transformation on the source and the target. Timings in seconds are into parenthesis. Best values are boldfaced. An example of a point clouds ($n=3000$) is provided on the left.}
    \label{table:ICP sinkhorn}}
\hfill
\vskip -.5in
\end{table}

\section{Conclusion}
In this paper, we hinge on the properties of sliced Wasserstein distance and on the Wasserstein generalized geodesics to define $\mswgg$, a new upper bound of the Wasserstein distance that comes with an associated transport map. Topological properties of $\swgg$ are provided, showing that it defines a metric and that $\mswgg$ metrizes the weak convergence of measure. We also \lc{propose} two algorithms for computing $\mswgg$,  either \lc{through} a random search scheme or a gradient descent procedure after smoothing the generalized geodesics definition of $\mswgg$. We illustrate its behavior in several experimental setups, notably showcasing its interest in applications where a transport map is needed.

The set of permutations \lc{covered} by $\mswgg$ is the one induced by projections and permutations on the line. It is a subset of the original Birkhoff polytope and it would be interesting to \lc{characterize} how these two sets relates. In particular, in the case of  empirical \lc{realizations} of continuous distributions, the behavior of $\mswgg$, when $n$ grows, needs to be investigated.
In addition, the fact that $\mswgg$ and WD coincide when $d>2n$ calls for embedding the distributions in higher dimensional spaces to benefit from \lc{the greater} expressive power of projection onto the line. \lc{Another} important consideration is to \lc{establish} a theoretical upper bound for $\mswgg$.

\section*{Acknowledgments}
The authors gratefully acknowledge
the financial support of the French Agence Nationale de la Recherche (ANR), under grant ANR-20-CHIA-0021-01 (project RAIMO\footnote{\url{https://chaire-raimo.github.io/}}), grant ANR-20-CHIA-0030 (project OTTOPIA) and grant  ANR-18-CE23-0022-01 (project MULTISCALE).
Cl\'ement Bonet is supported by the project DynaLearn from Labex CominLabs and Région Bretagne ARED DLearnMe.

\newpage

\bibliographystyle{plain}
\bibliography{main}

\newpage

\appendix
\section{Proofs and supplementary results related to Section \ref{sec:swgg v0}}

\subsection{Overestimation of WD by PWD}
\label{app:PWD}
As stated in Section \ref{sec:background}, the projected Wasserstein distance $\pwd$ (see Eq. \ref{eq:pwd}) tends to overestimate the Wasserstein distance. 
This is due to the fact that some permutations $\sigma_\theta$ and $\tau_\theta$ (with $\theta \in \Sd$) involved in PWD computation may be  irrelevant. Such situation occurs when the distributions are in high dimension but supported on a low dimensional manifold or when the distributions are multi-modal. 




Let consider the distributions $\mu_1$ and $\mu_2$ lying on a low dimensional manifold.  In high dimension, randomly sampled  vectors $\theta$ tend to be orthogonal.  Moreover, vectors  orthogonal to the low dimensional manifold lead to ``collapsed'' projected distributions $\thetap \mu_1$ and $\thetap \mu_2$  onto  $\theta$. Hence, such projection directions lead to 
permutations that can be randomly chosen. To empirically illustrate  this behavior of PWD, we consider $\mu_1$ and $\mu_2$ as Gaussian distributions in $\mathbb{R}^d$, $d=10$ but supported on the first two coordinates and we sample 200 points per distribution. Table \ref{tab:PWD} summarizes the obtained corresponding distances and shows that PWD overestimates the WD. 


Now, let us consider two multimodal distributions $\mu_1,\mu_2$ with $K$ clusters such that each cluster of $\mu_1$ has a close cluster from $\mu_2$ (cyclical monotonicity assumption). Also we assume the same number of points in each cluster. OT plan will match the corresponding clusters and will lead to a relatively low value for $W^2_2$ (since cluster from $\mu_1$ has a closely related cluster in $\mu_2$). However as $\pwd$ may allow permutations that make correspondences between points from different clusters (since a source cluster and a target cluster can be far in the original space but very close when projected on 1D), the resulting distance will be much more larger, leading to an overestimation of the Wasserstein distance. Table \ref{tab:PWD} provides an illustration for $K=10$ clusters and $d=2$. 

\begin{table}[ht]
    \centering
    \caption{Values of $W^2_2$, $\pwd$ and $\mswgg$ on two toy examples. PWD 
    samples $\theta$ uniformly over $\Sd$; PWD Orthogonal Projections seek orthogonal vectors (see \cite{rowland2019orthogonal} for more details)}
    \begin{tabular}{l c c} 
    Distributions & Multi-modal &Low dimensional manifold\\
    \hline
    $W^2_2$&12&12 \\
    $\pwd_2^2$ Monte-Carlo 
    & 54&29 \\
    $\pwd_2^2$ Orthogonal Projections& 54&37 \\
    $\mswgg_2^2$ & 13&13\\
     \hline
    \end{tabular}
    \label{tab:PWD}
\end{table}


\subsection{Quantile version of SWGG}
\label{app:quantile swgg}
The main body of the paper expresses $\swgg$ for  empirical  distributions $\mu_1$ and $\mu_2$ with  the same number of points and 
uniform probability masses. 
In this section we derived $\swgg$ in a more general setting of discrete distributions.

Let 
remark that $\mswgg$ 
relies on solving a 1D optimal transport (OT) problem. So far, the 1D OT problem was derived for $\mu_1,\mu_2 \in \mathcal{P}_2^n(\R)$ and thus was expressed using the permutation operators $\tau$ and $\sigma$. In the general setting of distributions $\mu_1 \in \mathcal{P}_2^n(\R)$ and $\mu_2 \in \mathcal{P}_2^m(\R)$ with $n \neq m$, the 1D optimal transport is computed based on quantile functions. Hence, the expression of $\swgg$ in the general setting of  $\mu_1 \in \mathcal{P}_2^n(\R)$ and $\mu_2 \in \mathcal{P}_2^m(\R)$ hinges on  quantile functions instead of permutations.

More formally, let $\mu \in \mathcal{P}_2^n(\R)$; its cumulative function is defined as:
\begin{align}
    F_\mu : \R \to [0,1] \; , \; x\mapsto \int_{-\infty}^xd\mu
\end{align}
and its quantile function (or pseudo inverse), is given by:
\begin{align}
    q_\mu : [0,1] \to \R \; , \; r \mapsto \min \{x \in \R \cup \{-\infty\} \; \text{s.t.} \; F_\mu(x)\geq r \} 
\end{align}
An important remark is that the quantile function is a step function 
with $n$ (the number of atoms) discontinuities. Thus, it can be stored efficiently using two vectors of size $n$ (one for the locations of the discontinuities and the other for the values of the discontinuities).

For $\mu_1 \in \mathcal{P}_2^n(\R)$ and $\mu_2\in \mathcal{P}^m_2(\R)$, we recover the Wasserstein distance through quantiles with:
\begin{align}
    W^2_2(\mu_1,\mu_2)=\int_0^1|q_{\mu_1}(r)-q_{\mu_2}(r)|^2dr
\end{align}
Moreover, the optimal transport plan is given by:
\begin{align}
    \pi=(q_{\mu_1},q_{\mu_2})_\# \lambda_{[0,1]}
\end{align}
where $\lambda_{[0,1]}$ is the Lebesgue measure on $[0,1]$. The transport plan can be stored efficiently using two vectors of size $(n+m-1)$ (see \cite{COTFNT} Prop 3.4).

Following \cite[Remark 9.6]{COTFNT}, one can define the quantile function related to the Wasserstein mean by :
\begin{align}
    q_{\mu^{1\to 2}}=\frac{1}{2}q_{\mu_1}+\frac{1}{2}q_{\mu_2}.
\end{align}

Now, let $\mu_1 \in \mathcal{P}^n_2(\R^d)$ and $\mu_2 \in \mathcal{P}^m_2(\R^d)$. Let $\pivot$ be the Wasserstein mean of the projected distributions on $\theta$. Finally let $\pi^{\theta \to 1}$ denote the transport plan from $\pivot$ to $\mu_1$ and $\pi^{\theta \to 2}$ be the transport plan from $\pivot$ to $\mu_2$. Following the construction of \cite[Sec. 9.2]{ambrosio2005gradient}, we shall introduce a multi marginal plan defined as:
\begin{align}
    \pi \in \mathcal{P}_2(\R^d\times \R^d\times \R^d) \text{ s.t. } P^{12}_\#\pi=\pi^{\theta \to 1} \; , \; P^{13}_\#\pi=\pi^{\theta\to 2} \text{ and } \pi \in \Pi(\pivot,\mu_1,\mu_2)
\end{align}
where $P^{12}:(\R^d)^3\to (\R^d)^2$ projects to the first two coordinates and $P^{13}$ projects to the coordinates 1 and 3. In particular, $P^{12}_\#\pi$ is the projection of $\pi$ on its 2 first marginals and $P^{13}_\#\pi$ on the first and 3rd marginal. Similarly to the 2-marginal transport plan we defined  $\Pi(\pivot,\mu_1,\mu_2) = \{\pi \in \mathcal{P}_2(\R^d \times \R^d\times \R^d)$ s.t. $\pi(A \times \R^d\times \R^d)=\pivot(A)$ , $\pi(\R^d \times A,\times \R^d)=\mu_1(A)$ and $\pi(\R^d \times \R^d \times A)=\mu_2(A)$, $\forall A$ measurable set of $\R^d \}$:

The generalized barycenter $\gene$ is then defined as:
\begin{align}
    \gene=\left(\frac{1}{2}P^2+\frac{1}{2}P^3\right)_\#\pi
\end{align}
where $P^i$ is the projection on the $i$-th coordinate.

We finally have all the building blocks to compute $\swgg$ in the general case. Let remark that the complexity goes from $\mathcal{O}(dn+n\log n)$ in the $\Pdn$ case to $\mathcal{O}(d(n+m)+(n+m)\log (n+m))$ in the general case.

\subsection{Proof of Proposition \ref{prop:dist upper cp}} 
\label{app:distance swgg cp}
We aim to prove that $\swgg^2_2(\mu_1,\mu_2,\theta)$ 
is an upper bound of $W_2^2(\mu_1,\mu_2)$ and that $\swgg(\mu_1,\mu_2,\theta)$ is a distance $\forall \theta \in \Sd ,\mu_i \in \Pdn$, $i=1,2$. 

\paragraph{Distance.}
Note that this proof will be derived for the alternative definition of $\swgg$ in supp. \ref{app:proof prop distance}.

Let $\mu_1=\frac{1}{n}\sum \delta_{\boldsymbol{x}_i},\mu_2=\frac{1}{n}\sum \delta_{\boldsymbol{y}_i},\mu_3=\frac{1}{n}\sum \delta_{\boldsymbol{z}_i}$ be in $\Pd$, let $\theta \in \Sd$. We note $\sigma$ (resp. $\tau$ and $\pi$) the permutation such that $\langle \boldsymbol{x}_{\sigma(1)},\theta \rangle \leq ... \leq... \langle \boldsymbol{x}_{\sigma(n)} ,\theta \rangle$ (resp. $\langle \boldsymbol{y}_{\tau(1)},\theta \rangle \leq ... \leq... \langle \boldsymbol{y}_{\tau(n)} ,\theta \rangle$ and $\langle \boldsymbol{z}_{\pi(1)},\theta \rangle \leq ... \leq... \langle \boldsymbol{z}_{\pi(n)} ,\theta \rangle$).

\textit{Non-negativity and finite value.} From the $\ell_2$ norm, it is derived 

\textit{Symmetry.} $\swgg^2_2(\mu_1,\mu_2,\theta)=\frac{1}{n}\sum_i\|\boldsymbol{x}_{\sigma(i)}-\boldsymbol{y}_{\tau(i)}\|_2^2=\frac{1}{n}\sum_i\|\boldsymbol{y}_{\tau(i)}-\boldsymbol{x}_{\sigma(i)}\|_2^2=\swgg^2_2(\mu_2,\mu_1,\theta)$

\textit{Identity property.} From one side, $\mu_1=\mu_2$ implies that $\langle \boldsymbol{x}_i ,\theta \rangle=\langle \boldsymbol{y}_i,\theta \rangle$, $\forall 1 \leq i \leq n$ and that $\sigma = \tau$, which implies  $\swgg^2_2(\mu_1,\mu_2,\theta)=0$. 

From the other side, $\swgg^2_2(\mu_1,\mu_2,\theta)=0 \implies \frac{1}{n}\sum\|\boldsymbol{x}_{\sigma(i)}-\boldsymbol{y}_{\tau(i)}\|^2_2=0 \implies \boldsymbol{x}_{\sigma(i)}=\boldsymbol{y}_{\tau(i)}$, $\forall 1 \leq i \leq n \implies \mu_1=\mu_2$.

\textit{Triangle Inequality.} We have $\swggr(\mu_1,\mu_2,\theta)=\left(\frac{1}{n}\sum_i\|\boldsymbol{x}_{\sigma(i)}-\boldsymbol{y}_{\tau(i)}\|^2_2\right)^{1/2}$ $\leq \left(\sum_i\|\boldsymbol{x}_{\sigma(i)}-\boldsymbol{z}_{\pi(i)}\|^2_2\right.$ $\left.+\sum_i\|\boldsymbol{z}_{\pi(i)}+\boldsymbol{y}_{\tau(i)}\|^2_2 \right)^{1/2}$ $\leq\left(\sum_i\|\boldsymbol{x}_{\sigma(i)}-\boldsymbol{z}_{\pi(i)}\|^2_2\right)^{1/2}+\left(\sum_i\|\boldsymbol{z}_{\pi(i)}+\boldsymbol{y}_{\tau(i)}\|^2_2 \right)^{1/2}=\swggr(\mu_1,\mu_3,\theta)+\swggr(\mu_3,\mu_2,\theta)$

\paragraph{Upper Bound}
The fact that $\mswgg^2_2$ in an upper bound of $W^2_2$ comes from the sub-optimality of the permutations $\sigma_\theta,\tau_\theta$. Indeed, they induce a one-to-one correspondence $\boldsymbol{x}_{\sigma_\theta(i)} \to \boldsymbol{y}_{\tau_\theta(i)}$ $\forall 1 \leq i \leq n$. This correspondence corresponds to a transport map $T^\theta$  such that $T^\theta_\#\mu_1=\mu_2$. Since $W^2_2 = \inf_{ T \;\text{s.t.} \;T_\#\mu_1=\mu_2}\frac{1}{n}\sum\|\boldsymbol{x}-T(\boldsymbol{x})\|^2_2$ we  necessarily have $W^2_2\leq \mswgg_2^2$.

\paragraph{Equality} The equality $W^2_2=\mswgg_2^2$ whenever $d>2n$ comes from the fact that all the permutations are within the range of $\swgg$. In particular minimizing $\swgg$ is equivalent to solve the Monge problem. We refer to Supp. \ref{app:sanity check and behavior} for more details. 

\subsection{Difference between max-SW and min-SWGG}
\label{app:diff mswgg and maxsw}
Herein, we give an example where the selected vectors $\theta$ for $\maxsw$ and $\mswgg$ differ. 

Let $\mu_1,\mu_2 \in \mathcal{P}(\R^2)$ be an empirical sampling of $\mathcal{N}(m_1,\Sigma_1)$ and of $\mathcal{N}(m_2,\Sigma_2)$ with $m_1=
\begin{pmatrix}
    -10 \\ 0
\end{pmatrix}
$, 
$m_2=
\begin{pmatrix}
    10 \\ 0
\end{pmatrix}$, $\Sigma_1=\begin{pmatrix}1 & 0\\ 0&11 \end{pmatrix}$ and $\Sigma_2=\begin{pmatrix}2 & 0\\ 0&2 \end{pmatrix}$.

Since these two distributions are far away on the $x$-coordinate, $\maxsw$ will catch this difference between the means by selecting $\theta\approx\begin{pmatrix}
    1 \\ 0
\end{pmatrix}$. Indeed, the projection on the $x$-coordinate represents the largest 1D WD.

Conversely, $\mswgg$ selects the pivot measure to be supported on $\theta\approx\begin{pmatrix}
    1 \\ 0
\end{pmatrix}$ that separates the two distributions. Indeed, this direction better captures the geometry of the 2 distributions, delivering permutations that are well grounded to minimize the transport cost.

\begin{figure}[ht]
    \centering
    \includegraphics[scale=0.3]{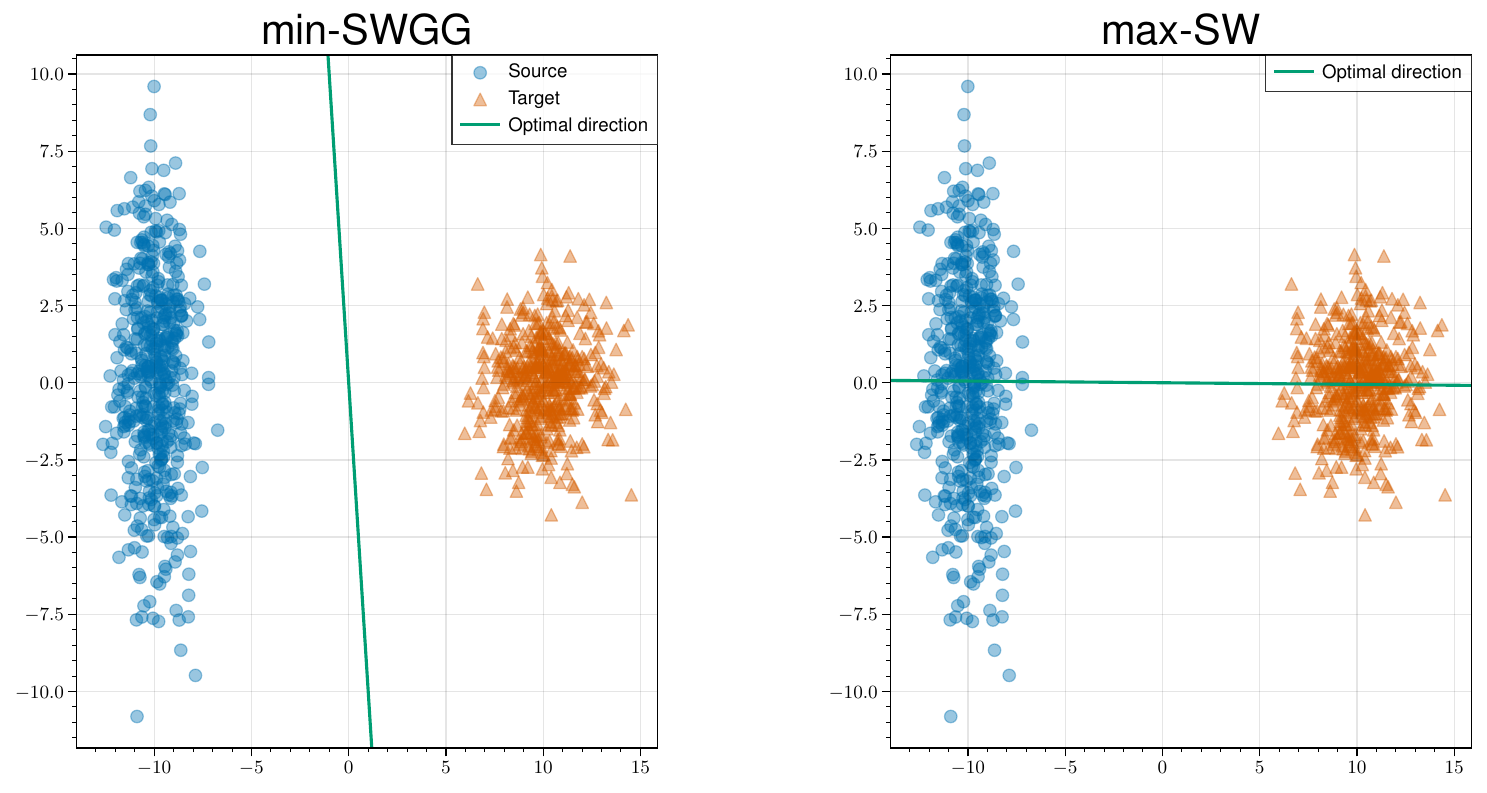}
    \caption{Optimal $\theta$ for $\maxsw$ and $\mswgg$}
    \label{fig:diff maxS}
\end{figure}

 Fig. \ref{fig:diff maxS} illustrates that difference between $\maxsw$ and $\mswgg$.

\subsection{From permutations to transport map}
\label{app:permutation to transport map}
In this section we provide the way of having a transport map from permutations.

Let $\mu_1,\mu_2 \in \Pdn$, let $\theta^* \in \arg \min \swgg$ and let $\sigma_{\theta^*},\tau_{\theta^*}$ the associated permutations. The associated map must be $T(\boldsymbol{x}_{\sigma(i)})=\boldsymbol{y}_{\tau(i)} \; \forall 1 \leq i \leq n$.  In the paper, we formulate the associated transport map as:
\begin{equation}
T(\boldsymbol{x}_i) = \boldsymbol{y}_{\tau_{\theta^*}^{-1}(\sigma_{\theta^*}(i))}, \quad \forall 1 \leq i \leq n.
\end{equation}
Moreover, the matrix representation of $T$ is given by:
\begin{align}
    T_{ij}=\left\{ \begin{array}{cl}
         \frac{1}{n} & \text{ if } \sigma(i)=\tau(j)  \\
         0 & \text{ otherwise } 
    \end{array}\right.
\end{align}

\subsection{Examples of Transport Plan}
\label{app:example transport plan}

\begin{figure}[ht]
    \centering
    \includegraphics[scale=0.13]{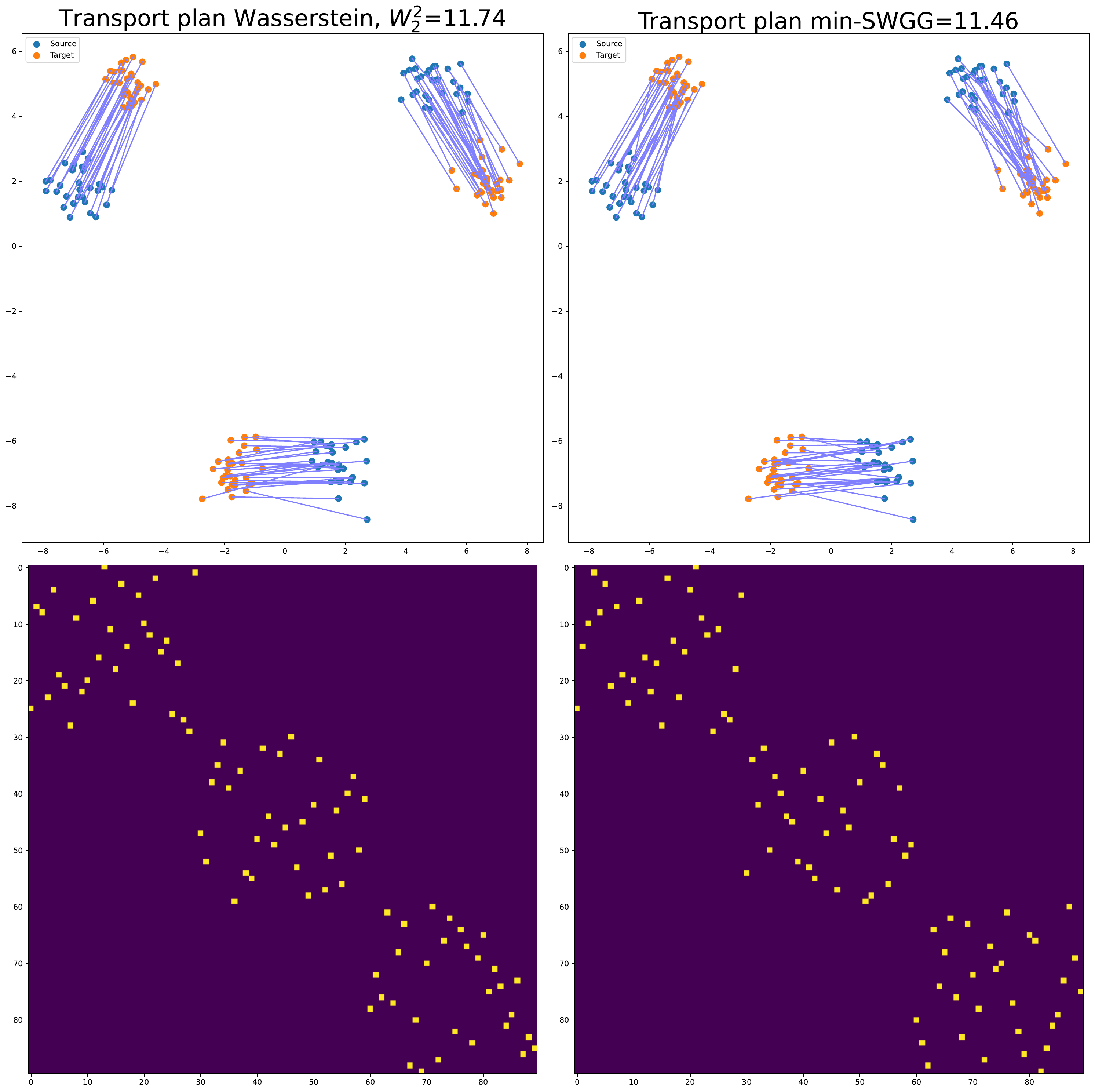}
    \includegraphics[scale=0.13]{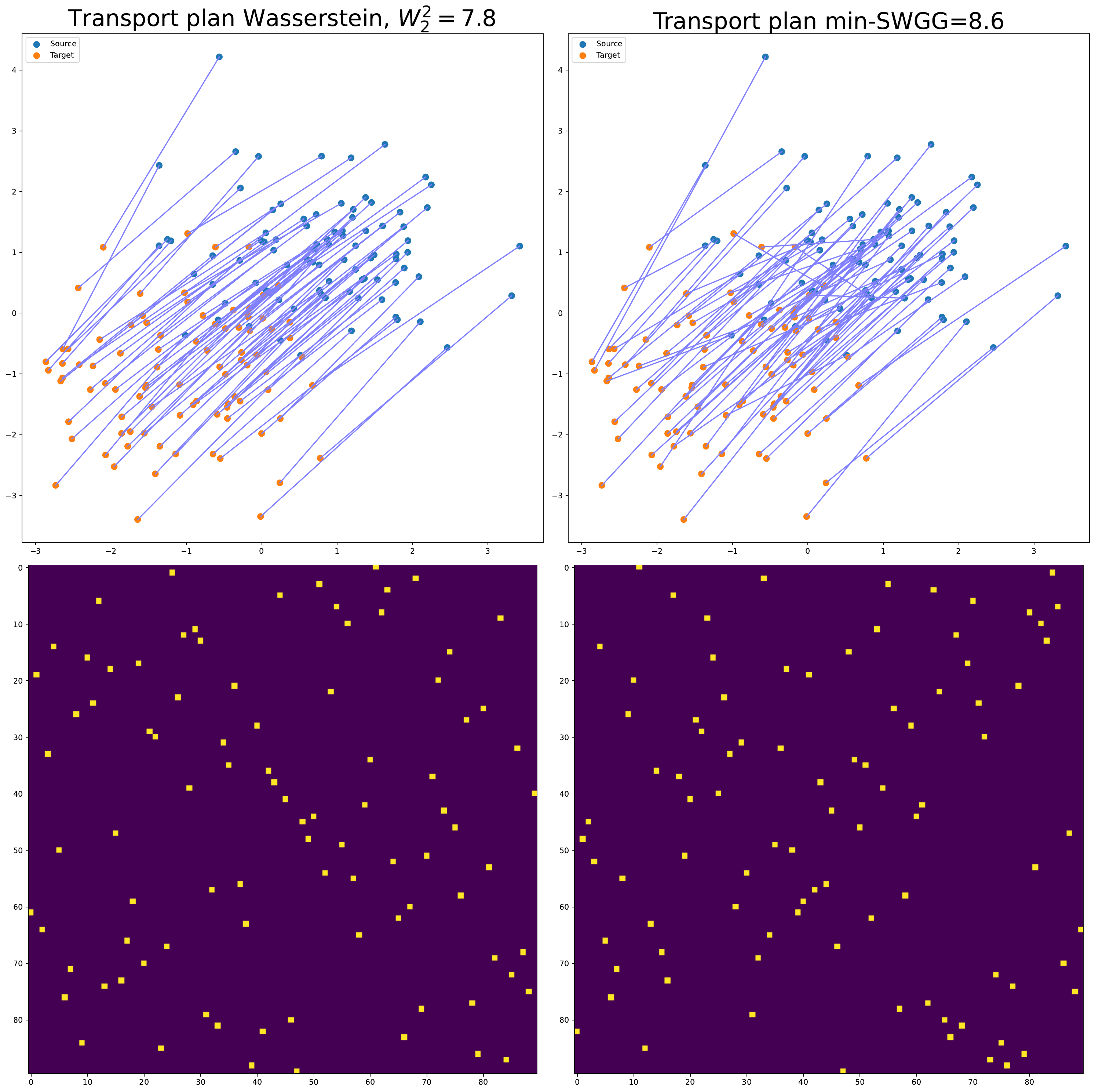}
    \caption{Example of transports plan given by Wasserstein (left and middle-right) and min-SWGG (middle left and right). Transport plan distribution (top) and transport matrix (bottom).The relative distances between source and target are given in the title.}
    \label{fig:example transport plan}
\end{figure}

Fig. \ref{fig:example transport plan} illustrates two instances of the transport plan obtained via $\mswgg$. 
Even though these transport plans are not optimal, 
they were able to capture the 
overall structure of the true optimal transport plans.

The first example shows that the OT plan by $\mswgg$ exhibits a "block" structure, and thus approximates well the true Wasserstein distance. The second example shows that even in a context of superimposed distribution the "general transport direction" in $\mswgg$ is representative of that of the optimal transport map.

\section{Background on Wasserstein Generalized Geodesics}
\label{app:generalized geodesic}

We introduce some concepts related the Wasserstein generalized geodesics in Sec. \ref{sec:definition and computation gg}. In this section, we provide more details about these geodesics in order to provide a wider view on this theory. 

In the following definitions, we do not address the issue of uniqueness of the geodesics. However this is not a problem in our setup since we focus our study on pivot measure with $n$-atoms $\nu \in \Pdn$. In this case, we have uniqueness of the $\nu$-based Wasserstein distance \cite{nenna2023transport}. 

\paragraph{Wasserstein generalized geodesics}
As mentioned in Sec. \ref{sec:definition and computation gg}, Wasserstein generalized geodesics rely on a pivot measure $\nu \in \Pdn$ to transport $\mu_1$ to $\mu_2$. Indeed, one can leverage the optimal transport maps $T^{\nu \to \mu_1}$ and $T^{\nu \to \mu_2}$ to construct a curve linking $\mu_1$ to $\mu_2$. The generalized geodesic with pivot measure $\nu$ is defined as:
\begin{equation}
    \genegeodesic \defeq \left((1-t)T^{\nu \to \mu_1} + tT^{\nu \to \mu_2}\right)_\# \nu \quad \quad \forall t \in[0,1].
\end{equation}
The generalized Wasserstein mean refers to the middle of the geodesic, i.e. when $t=0.5$ and has been denoted $\gbarycenter$. 

Intuitively, the optimal transport maps between $\nu$ and $\mu_i, i=1,2$ give rise to a sub-optimal transport map between $\mu_1$ and $\mu_2$ through:
\begin{align}
    T_\nu^{1\to 2} \defeq T^{\nu \to \mu_2} \circ T^{\mu_1 \to \nu} \quad \text{with} \quad (T_\nu^{1\to 2})_\#\mu_1=\mu_2.
\end{align}

$T_\nu^{1 \to 2}$ links $\mu_1$ to $\mu_2$ via the generalized geodesic:
\begin{align}
    \genegeodesic=((1-t)Id + tT_\nu^{1\to 2})_\# \mu_1.
\end{align}

We recall here the $\nu$-based Wasserstein distance induced by $T_\nu^{1\to 2}$ and introduced in Eq. \eqref{eq:v-based Wasserstein def}.
\begin{definition}
The $\nu$-based Wasserstein distance \cite{craig2016exponential, nenna2023transport} is defined as:
\begin{align}
    W^2_\nu(\mu_1,\mu_2)\defeq& \int_{\R^d} \|\boldsymbol{x}-T_\nu^{1\to 2}(\boldsymbol{x})\|^2_2d\mu_1(\boldsymbol{x})\\
    =&\int_{\R^d} \|T^{\nu \to \mu_1}(\boldsymbol{z})-T^{\nu \to \mu_2}(\boldsymbol{z})\|^2_2d\nu(\boldsymbol{z}). 
\end{align}
\end{definition}

Moreover, this new notion of geodesics comes with an inequality, which is of the opposite side to Eq.~\eqref{eq:inequality positive curvature}:
\begin{align}
   W^2_2(\genegeodesic,\nu) & \leq   (1-t)W^2_2(\mu_1,\nu)+tW_2^2(\nu, \mu_2)- t(1-t)W_2^2(\mu_1, \mu_2).
    \label{eq:inequality negative curvature}
\end{align}

\begin{figure}[ht]
    \centering
    \tikzset{every picture/.style={line width=0.75pt}} 

\begin{tikzpicture}[x=0.75pt,y=0.75pt,yscale=-1,xscale=1]

\draw [line width=1.5]  [dash pattern={on 1.69pt off 2.76pt}]  (32,150.67) .. controls (59.11,130.61) and (106.28,129.81) .. (129.26,148.26) ;
\draw [shift={(132,150.67)}, rotate = 223.84] [fill={rgb, 255:red, 0; green, 0; blue, 0 }  ][line width=0.08]  [draw opacity=0] (13.4,-6.43) -- (0,0) -- (13.4,6.44) -- (8.9,0) -- cycle    ;
\draw [line width=1.5]  [dash pattern={on 1.69pt off 2.76pt}]  (311.33,148) .. controls (336.97,165) and (381.64,167.92) .. (412.3,149.83) ;
\draw [shift={(415.58,147.78)}, rotate = 146.46] [fill={rgb, 255:red, 0; green, 0; blue, 0 }  ][line width=0.08]  [draw opacity=0] (13.4,-6.43) -- (0,0) -- (13.4,6.44) -- (8.9,0) -- cycle    ;
\draw [line width=0.75]    (32,150.67) -- (72.24,87.78) ;
\draw [line width=0.75]    (72.24,87.78) -- (132,150.67) ;
\draw [line width=0.75]    (72.24,87.78) -- (77.21,134.96) ;
\draw [line width=0.75]    (219.58,87.11) -- (223.67,148.91) ;
\draw [line width=0.75]    (180.9,149.6) -- (219.58,87.11) ;
\draw [line width=0.75]    (271.18,149.98) -- (219.58,87.11) ;
\draw [line width=0.75]    (311.9,149.5) -- (351.58,85.11) ;
\draw [line width=0.75]    (351.58,85.11) -- (415.58,147.78) ;
\draw [line width=0.75]    (351.58,85.11) -- (363.02,161.86) ;
\draw [line width=1.5]  [dash pattern={on 1.69pt off 2.76pt}]  (180.9,149.6) -- (267.18,149.96) ;
\draw [shift={(271.18,149.98)}, rotate = 180.24] [fill={rgb, 255:red, 0; green, 0; blue, 0 }  ][line width=0.08]  [draw opacity=0] (13.4,-6.43) -- (0,0) -- (13.4,6.44) -- (8.9,0) -- cycle    ;

\draw (18,150) node [anchor=north west][inner sep=0.75pt]  [font=\small]  {$\mu _{1}$};
\draw (298,150) node [anchor=north west][inner sep=0.75pt]  [font=\small]  {$\mu _{1}$};
\draw (69.07,73.82) node [anchor=north west][inner sep=0.75pt]  [font=\small]  {$\nu $};
\draw (346.93,69.96) node [anchor=north west][inner sep=0.75pt]  [font=\small]  {$\nu $};
\draw (135,150) node [anchor=north west][inner sep=0.75pt]  [font=\small]  {$\mu _{2}$};
\draw (416.67,145) node [anchor=north west][inner sep=0.75pt]  [font=\small]  {$\mu _{2}$};
\draw (168,150) node [anchor=north west][inner sep=0.75pt]  [font=\small]  {$x_{1}$};
\draw (216.27,72) node [anchor=north west][inner sep=0.75pt]  [font=\small]  {$y$};
\draw (269.33,150.02) node [anchor=north west][inner sep=0.75pt]  [font=\small]  {$x_{2}$};
\draw (63,143.02) node [anchor=north west][inner sep=0.75pt]  [font=\scriptsize]  {$\mu ^{1\ \rightarrow 2}(t)$};
\draw (343,168.09) node [anchor=north west][inner sep=0.75pt]  [font=\scriptsize]  {$\mu ^{1\rightarrow 2}_g(t)$};
\draw (185,155) node [anchor=north west][inner sep=0.75pt]  [font=\small]  {$tx_1+(1-t)x_2$};

\end{tikzpicture}
    \caption{Geodesic $(tId+(1-t)T^{1 \to 2})_\# \mu_1$ and generalized geodesic $(tId + (1-t)T^{1 \to 2}_\nu)_\#\mu_1$ in Wasserstein space (Left and Right) in dashed line and parallelogram law in $\R^d$ (middle).}
    \label{fig:geodesics}
\end{figure}
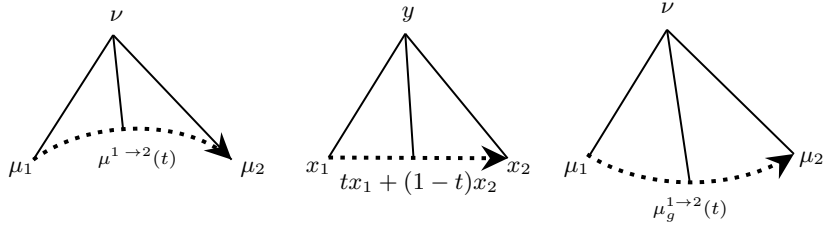

The parallelogram law is not respected but straddles with eq. \eqref{eq:inequality positive curvature} and eq. \eqref{eq:inequality negative curvature}. We refer to Figure \ref{fig:geodesics} for an intuition behind positive curvature \cite{otto2001geometry}, parallelogram law and generalized geodesics.

Setting $t=0.5$ in Eq. \eqref{eq:inequality negative curvature} and reordering the term gives:
\begin{align}
W^2_2(\mu_1,\mu_2) \leq 2W^2_2(\mu_1,\nu)+2W^2_2(\nu,\mu_2)-4W^2_2(\gbarycenter,\nu).
\end{align}

Moreover one can remark that:
\begin{align}
    W^2_\nu(\mu_1,\mu_2)=2W^2_2(\mu_1,\nu)+2W^2_2(\nu,\mu_2)-4W^2_2(\gbarycenter,\nu)
\end{align}

In particular situations $W_\nu^2$ and $W^2_2$ coincide. It is the case for 1D distributions where the Wasserstein space is known to be flat \cite{ambrosio2005gradient}. In that case, the Wasserstein mean and the generalized Wasserstein mean are the same.

\paragraph{Multi-marginal}
Another formulation of the $\nu$-based Wasserstein distance is possible through the perspective of multi-marginal OT \cite{ambrosio2005gradient}. Let $\Pi(\mu_1,\mu_2,\nu)=\{\pi \; \text{s.t.} \; P^{12}_\#\pi=\pi^{1 \to 2} \; , \; P^{13}_\# \pi=\pi^{1 \to \nu} \; \text{and} \; P^{23}_\#\pi=\pi^{2 \to \nu} \}$, where $P^{ij}$ is the projection onto the coordinates $i,j$. Let also $\Pi^*(\mu_i,\nu)$ be the space of optimal transport maps between $\mu_i$ and $\nu$. We have:
\begin{align}
    W_\nu^2(\mu_1, \mu_2)=\inf_{ \pi \in \Pi(\mu_1,\mu_2,\nu) \; \text{s.t.} \;  P^{i3}_\# \pi \in \Pi^*(\mu_i,\nu) \; i=1,2}\int_{\R^d} \|\boldsymbol{x}-\boldsymbol{y}\|^2_2d\pi(\boldsymbol{x},\boldsymbol{y})
    \label{eq:multi ot generalized geodesic}
\end{align}
 Equation \eqref{eq:multi ot generalized geodesic} expresses the fact that we select the optimal plan from $\Pi(\mu_1,\mu_2,\nu)$ which is already optimal for $\Pi(\mu_i,\nu)$. Mathematically, this minimization is not a multi-marginal problem, since the optimal plan is supposed to be already optimal for some coordinate.

The set $\{ \pi \in \Pi(\mu_1,\mu_2,\nu) \; \text{s.t.} \;  P^{i3}_\# \pi \in \Pi^*(\mu_i,\nu) \; i=1,2\}$ is never empty, i.e. there is always existence of $\pi_\nu^{1\to 2}$ (thanks to the gluing lemma \cite{villani2009optimal}, page 23). Moreover, in situations where it is a singleton, there is uniqueness of $\pi^{1\to 2}_\nu$. Uniqueness is an ingredient which overpasses the selection of a final coupling and comes with additional result.
\begin{lemma}[Lemma 6 \cite{nenna2023transport}]
Whenever $\{ \pi \in \Pi(\mu_1,\mu_2,\nu) \; \text{s.t.} \;  P^{i3}_\#\pi \in \Pi^*(\mu_i,\nu) \; i=1,2\}$ is a singleton, $W_\nu^2$ is a proper distance. It is a semi-distance otherwise.
\end{lemma}

Notably, 1D pivot measure was studied in \cite{kim2020optimal} to ensure a dendritic structure of the distributions along the geodesic.

\section{Related Works}
\label{app:related works gg}

In this section we highlight the fact that several upper approximations of $W^2_2$ are in the framework of generalized geodesics. The differences lay in the choice of the pivot measure $\nu$.

\paragraph{Factored Coupling.}
In \cite{forrow2019statistical}, the authors impose a low rank structure on the transport plan by factorizing the couplings through a pivot measure $\nu$ expressed as the $k$-Wasserstein mean between $\mu_1$ and $\mu_2$ ($k \leq n$). It is of particular interest since whenever the pivot distribution is the Wasserstein mean between $\mu_1$ and $\mu_2$, $W_\nu^2$ and $W_2^2$ coincide. 

Factored coupling results in a problem of computing the $k$-Wasserstein mean ($\mu^{1\to 2}$) followed by solving two OT problems between the clustered Wasserstein mean and the two input distributions ($W_2^2(\mu_1,\mu^{1 \to 2})$ and $W^2_2(\mu^{1 \to 2},\mu_2)$). Even though the OT problems are smaller, they are still expensive in practice.

Moreover, in this scenario, the uniqueness of the OT plan $T^{1 \to 2}_\nu$ is not ensured. It 
appears
that \cite{forrow2019statistical} 
chooses the most entropic transport plan, i.e. simply $T^{1\to 2}_\nu=T^{\mu^{1 \to 2} \to \mu_2} \circ T^{\mu_1 \to \mu^{1 \to 2}}$.

\paragraph{Subspace Detours.} 
From a statistical point of view, it is beneficial to consider optimal transport on a lower dimensional manifold \cite{weed2019sharp}. In \cite{muzellec2019subspace}, authors compute an optimal transport plan $T^{\mu^E_1 \to \mu^E_2}$ between projections on a lower linear subspace $E$ of $\mu_1$ and $\mu_2$, i.e. $\mu_i^E=P_E \# \mu_i$, where $P_E$ is the linear projection on $E$. They aimed at leveraging $T^{\mu_1^E \to \mu_2^E}$ to construct a sub-optimal map $T_E^{1\to 2}$ between $\mu_1$ and $\mu_2$.

The problem can be recast as a generalized geodesic problem with $\nu$ being the Wasserstein mean of $\mu_1^E$ and $\mu_2^E$ embedded in $\R^d$. Once again, uniqueness of $T_\nu^{\mu_1 \to \mu_2}$ is not guaranteed, authors provide two ways of selecting the map, namely Monge-Knothe and Monge-Independent lifting. 

Subspace detours result in a problem where one needs to select a linear subspace $E$ (which is a non convex procedure), compute an optimal transport between $\mu_1$ and $\mu_2$ (in $\mathcal{O}(n^3\log n)$ whenever $\dim(E)>1$) and reconstruct $T_E^{\mu_1 \to \mu_2}$.

\paragraph{Linear Optimal Transport (LOT).}
Given a set of distributions $(\mu_i)_{i=1}^m \in \Pd^m$, LOT \cite{wang2013linear} embeds the set of distributions into
the $L^2(\nu)$-space by computing the OT of each distribution to the pivot distribution. Mathematically, it computes $T^{\nu \to \mu_i} \; \forall 1 \leq i \leq m$ and lies on estimating $W_2^2(\mu_i,\mu_j)$ with $W_\nu^2(\mu_i,\mu_j)$ through eq.~\eqref{eq:v-based Wasserstein def}.

In LOT, the pivot measure $\nu$  was chosen to be the average of the input measures \cite{wang2013linear}, the Lebesgue measure on $\R^d$ \cite{merigot2020quantitative} or an isotropic Gaussian distribution \cite{moosmuller2020linear}.

Instead of computing $\binom{m}{2}$ expensive Wasserstein distances, it resorts only  on $m$ Wasserstein distances between $(\mu_i)_i^m$ and $\nu$. While significantly reducing the computational cost when several distributions are at stake, it does not allow speeding up the computation when only two distributions are involved.

\subsection{Linear Optimal Transport with shift and scaling}
\label{app:shift and scaling}

In this section, we recall the result from \cite{moosmuller2020linear}. The theorem states that the $\nu$-based approximation is very close to  WD whenever $\mu_1$, $\mu_2$ are continuous distributions which are very close to be shift and scaling of each other. It can applies to a continuous version of $\swgg$, however it works with discrete measures in the particular case of equality between $W^2_\nu$ and $W_2^2$. 

\begin{thm}[Theorem 4.1 \cite{moosmuller2020linear}]
Let $\Lambda=\{S_a \;\text{(shift)}\;,a\in \R^d\}\cup \{R_c \; \text{(scaling)}\;, c \in \R \}$, $\Lambda_{\mu,R}=\{h\in \Lambda \; \text{s.t.}\; \|h\|_\mu\geq R \}$ and $G_{\mu,R,\epsilon}=\{g \in L^2(\R^d,\mu) \; \text{s.t.} \; \exists h \in \Lambda_{\mu,R} \; \text{s.t.} \; \|g-h\|_\mu \leq \epsilon\}$

Let $\nu$, $\mu \in \Pd$, with $\mu,\nu \ll \lambda$ (the Lebesgue measure). Let $R>0,\epsilon>0$
\begin{itemize}
    \item For $g_1,g_2 \in G_{\mu,R,\epsilon}$ and $\nu=\lambda$ on a convex compact subset of $\R^d$, we have:
    \begin{align}
        W_\nu(g_{1\#}\mu,g_{2\#}\mu)-W_2(g_{1\#}\mu,g_{2\#}\mu)\leq C\epsilon^{\frac{2}{15}}+2\epsilon
    \end{align}
    \item If $\mu$ and $\nu$ satisfy the assumption of Caffarelli's regularity theorem \cite{caffarelli2000monotonicity}, then for $g_1,g_2 \in G_{\mu,R,\epsilon}$, we have:
    \begin{align}
        W_\nu(g_{1\#}\mu,g_{2\#}\mu)-W_2(g_{1\#}\mu,g_{2\#}\mu)\leq \overline{C}\epsilon^{1/2}+C\epsilon
    \end{align}
\end{itemize}
where $C,\overline{C}$ depdends on $\nu,\mu$ and $R$.
\end{thm}

\section{Proofs and other results related to Section \ref{sec:swgg v1}}

\subsection{Proof of Proposition \ref{def:swgg generalized geodesic}: equivalence between the two formulations of SWGG}
\label{app:equivalence swgg}

In this section, we prove that the two definitions of $\swgg$ in Def. \ref{def:swgg permutation} and Prop. \ref{def:swgg generalized geodesic} are equivalent. Let $\theta \in \Sd$ be fixed.

From one side in Def. \ref{def:swgg permutation}, we have:
\begin{align}
    \swgg_2^2(\mu_1,\mu_2,\theta)\defeq \frac{1}{n}\sum_i \|\boldsymbol{x}_{\sigma_\theta(i)}-\boldsymbol{y}_{\tau_\theta(i)}\|^2_2
\end{align}
where $\sigma_\theta$ and $\tau_\theta$ are the permutations obtained by sorting $P^\theta_\#\mu_1$ and $P^\theta_\#\mu_2$.

From the other side we note $D(\mu_1,\mu_2,\theta)$ the quantity:
\begin{align}
     D(\mu_1,\mu_2,\theta) \defeq 2W^2_2(\mu_1,\pivot)+2W_2^2(\pivot,\mu_2)-4W_2^2(\lgbarycenter,\pivot).
\end{align}
We want to prove that $\swgg^2_2(\mu_1,\mu_2,\theta)=D(\mu_1,\mu_2,\theta), \quad \forall \mu_1, \mu_2 \in \Pdn$ and $\theta \in \Sd$.

Eq. \eqref{eq:v-based Wasserstein def} in the main paper states that $D(\mu_1,\mu_2,\theta)$ is equivalent to $\int_{\R^d}\|\boldsymbol{x}-T^{1 \to 2}_{\pivot}(\boldsymbol{x})\|^2_2d\mu_1(\boldsymbol{x})$. 

Finally, Lemma \ref{lemma:closed form} states that the transport map $T^{1\to 2}_{\pivot}$ is fully determined by the permutations on the line: the projections part is a one-to-one correspondence between $\boldsymbol{x}$ and $\theta \langle \boldsymbol{x} ,\theta \rangle$ (resp. between $\boldsymbol{y}$ and $\theta \langle \boldsymbol{y} ,\theta \rangle$). More formally $T^{1 \to 2}_{\pivot}(\boldsymbol{x}_{\sigma_\theta(i)})=\boldsymbol{y}_{\tau_\theta(i)} \quad \forall 1 \leq i \leq n$. And thus we recover:
\begin{align}
    \int_{\R^d}\|\boldsymbol{x}-T^{1 \to 2}_{\pivot}(\boldsymbol{x})\|^2_2d\mu_1(\boldsymbol{x})=\frac{1}{n}\sum_i \|\boldsymbol{x}_{\sigma_\theta(i)}-\boldsymbol{y}_{\tau_\theta(i)}\|^2_2
\end{align}
which concludes the proof.

\subsection{Proof of Weak Convergence (Proposition \ref{prop:weak convergence})}
\label{proof prop weak convergence}
We want to prove that, for a sequence of measures $(\mu_k)_{k \in \mathbb{N}} \in \Pdn$, we have:
\begin{align}
    \mu_k \towd \mu \in \Pdn) \iff \mswgg_2^2(\mu_k,\mu)\tok 0
\end{align}

The notation $\mu_{k} \towd \mu$ stands for the weak convergence in $\Pdn$ i.e. $\int_{\R^d}f(\boldsymbol{x})d\mu_{(k)}(\boldsymbol{x})\to \int_{\R^d}f(\boldsymbol{x})d\mu(\boldsymbol{x})$ for all continuous bounded functions $f$ and for the Euclidean distance $f(\boldsymbol{x})= \|\boldsymbol{x}_0-\boldsymbol{x}\|_2^2$ for all $x_0 \in \R^d$.

From one side, if $\mswgg_2^2(\mu_k,\mu) \to 0 \implies W^2_2(\mu_k,\mu) \to 0 \implies \mu_k \towd \mu$. The first implication is due to the fact that $\mswgg_2^2$ is an upper-bounds of $W^2_2$, the Wasserstein distance, and that WD metrizes the weak convergence.

From another side, assume $\mu_k \towd \mu$; we have for any $\theta$:
\begin{enumerate}
    \item Let $\mu_\theta^{\mu_k \to \mu } \in \Pdn$ stands for the Wasserstein mean of the projections $\thetapp \mu_k$ and $\thetapp \mu$ and let 
    $\mu_\theta^{\mu \to \mu } = \thetapp \mu$. We have $\mu_\theta^{\mu_k \to \mu }$ converges towards (in law) to $\mu_\theta^{\mu \to \mu }$, which implies that:
    \begin{align}
        W^2_2(\mu_k,\mu_\theta^{\mu_k \to \mu }) \tok W^2_2(\mu,\mu_\theta^{\mu \to \mu}).
    \end{align}

    \item Since $\mu \in \Pdn$, we have $T^{\mu_\theta^{\mu_k \to \mu } \to \mu_k} \tok T^{\mu_\theta^{\mu_k \to \mu } \to \mu}$ (see \cite{cuesta1997optimal}, theorem 3.2). It implies that $\mu_{g, \theta}^{\mu_k \to \mu} \tow \mu$ and particularly:
    \begin{align}
        W^2_2(\mu_{g, \theta}^{\mu_k \to \mu}, \mu_\theta^{\mu_k \to \mu }) \tok W^2_2(\mu,\mu_\theta^{\mu \to \mu})
    \end{align}
\end{enumerate}

By combining the previous elements, we get:
\begin{align}
2W^2_2(\mu_k,\mu_\theta^{\mu_k \to \mu })+2W_2^2(\mu_\theta^{\mu_k \to \mu },\mu_k)-4W^2_2(\mu_{g, \theta}^{\mu_k \to \mu}, \mu_\theta^{\mu_k \to \mu })\tok & 2W^2_2(\mu,\mu_\theta^{\mu \to \mu})  \nonumber \\
& +2W^2_2(\mu_\theta^{\mu \to \mu},\mu) \nonumber \\
& -4W^2_2(\mu,\mu_\theta^{\mu \to \mu})=0
\end{align}
The previous relation shows that $\mu_k \towd \mu$ implies $\swgg^2_2(\mu_k,\mu,\theta) \tok 0$ for any $\theta$. Hence, we can conclude that:
\begin{align}
    \mu_k \towd \mu \implies \mswgg_2^2(\mu_k,\mu) \to 0
\end{align}
This concludes the proof. 

Note that when $\mu_1$ and $\mu_2$ are continuous, \cite{merigot2020quantitative} proved that when the distributions are smooth enough (i.e. respecting the Cafarelli theorem \cite{caffarelli2000monotonicity}), there is a bi-Holder equivalence between the $\nu$-based Wasserstein distance and $W^2_2$. Hence, it still holds for $\swgg$ for any $\theta\in S^{d-1}$:
\begin{align}
    W^2_2(\mu_1,\mu_2) \leq \swgg_2^2(\mu_1,\mu_2, \theta) \leq B \times W^2_2(\mu_1,\mu_2)^{2/15} \quad \quad \forall \mu_i \in \Pd
\end{align}
where $B$ depends on $\mu_i, i \in \{1, 2\}, \theta$ and the dimension $d$. This bound is sufficient to prove that $\swgg$ metrizes the weak convergence in this context. We refer to \cite{merigot2020quantitative} for more details.

\subsection{Proof of Translation property (Proposition \ref{proposition translation})}
\label{proof prop translation}
We  prove that $\mswgg^2_2$ has the same behavior w.r.t. the translation as $W^2_2$. This property is well known for Wasserstein and useful in applications such as shape matching.

Let $\mu_1,\mu_2 \in \Pdn$, and let $T^u$ (resp. $T^v$) be the map $\boldsymbol{x} \mapsto \boldsymbol{x}-\boldsymbol{u}$ (resp. $\boldsymbol{x}\mapsto \boldsymbol{x}-\boldsymbol{v}$), with $\boldsymbol{u},\boldsymbol{v}$ vectors of $\R^d$. 

To ease the notations, let define $\tilde{\mu}_1 = T^u_\#\mu_1$ and $\tilde{\mu}_2 = T^v_\#\mu_2$.

Let remind that in the case of Wasserstein distance we have \cite{COTFNT}(Remark 2.19):
\begin{align}
    W^2_2(\tilde{\mu}_1, \tilde{\mu}_2) \defeq W^2_2(T^u_\#\mu_1,T^v_\#\mu_2)=W^2_2(\mu_1,\mu_2)-2\langle \boldsymbol{u}-\boldsymbol{v},\boldsymbol{m}_1-\boldsymbol{m}_2\rangle +\|\boldsymbol{u}-\boldsymbol{v}\|^2_2
\end{align}
with $\boldsymbol{m}_1=\int_{\R^d}\boldsymbol{x}d\mu_1(\boldsymbol{x}) $ and $\boldsymbol{m}_2=\int_{\R^d}\boldsymbol{x}d\mu_2(\boldsymbol{x})$.

We aim to compute $\mswgg^2_2(\tilde{\mu}_1, \tilde{\mu}_2) \defeq \mswgg^2_2(T^u_\#\mu_1,T^v_\#\mu_2)$. Let express first
\begin{align}
\label{eq:swgg of shifted measures}
    \swgg^2_2(\tilde{\mu}_1, \tilde{\mu}_2) & = 2W^2_2(\tilde{\mu}_1,\tilde{\mu}^{1 \to 2}_\theta) +2W^2_2(\tilde{\mu}_2,\tilde{\mu}^{1 \to 2}_\theta)  -4W^2_2(\tilde{\mu}^{1 \to 2}_{g,\theta},\tilde{\mu}^{1 \to 2}_\theta)
\end{align}
where $\tilde{\mu}^{1 \to 2}_{\theta}$ is the Wasserstein mean of the projections along $\theta$ of the shifted measures $\tilde{\mu}_1 = T^u_\#\mu_1$ and $\tilde{\mu}_2 = T^v_\#\mu_2$ as in Proposition \ref{eq:geodesic}. The generalized Wasserstein mean $\tilde{\mu}^{1 \to 2}_{g,\theta}$ is  defined accordingly (see also Proposition \ref{eq:gengeo}).

We have:
\begin{align}
    W^2_2(\tilde{\mu}_1, \tilde{\mu}^{1 \to 2}_{\theta})=W^2_2(\mu_1,\pivot)-2\langle \boldsymbol{u},\boldsymbol{m}_1-\boldsymbol{m}_3\rangle +\|\boldsymbol{u}\|^2_2
\end{align}
where $\boldsymbol{m}_3=\int_{\R^d}\boldsymbol{x}d\tilde{\mu}^{1 \to 2}_{\theta}(\boldsymbol{x})$.

Similarly $W^2_2(\tilde{\mu}_2, \tilde{\mu}^{1 \to 2}_{\theta})=W^2_2(\mu_2, \pivot)-2\langle \boldsymbol{v},\boldsymbol{m}_2-\boldsymbol{m}_3\rangle +\|\boldsymbol{v}\|^2_2 $.

Let express now the third term in eq. \eqref{eq:swgg of shifted measures}. For that we require to define the generalized Wasserstein mean $\tilde{\mu}^{1 \to 2}_{g,\theta}$ with pivot measure $\tilde{\mu}^{1 \to 2}_{\theta}$. By the virtue of eq. \eqref{eq:gengeo} in the main paper, we have:
\begin{align}
    \tilde{\mu}^{1 \to 2}_{g,\theta} &= \left(\frac{1}{2}T^{\tilde{\mu}^{1 \to 2}_{\theta} \to \tilde\mu_1} + \frac{1}{2}T^{\tilde{\mu}^{1 \to 2}_{\theta} \to \tilde\mu_2}\right)_\#\tilde{\mu}^{1 \to 2}_{\theta}\\
    &=\left(\frac{1}{2}T^{{\mu}^{1 \to 2}_{\theta} \to \mu_1} +\frac{1}{2}T^{{\mu}^{1 \to 2}_{\theta} \to \mu_2}-T^\frac{u+v}{2} \right)_\#\tilde{\mu}^{1 \to 2}_{\theta}\\
    &=T^{\frac{u+v}{2}}_\# \left( \left(\frac{1}{2}T^{{\mu}^{1 \to 2}_{\theta} \to \mu_1} +\frac{1}{2}T^{{\mu}^{1 \to 2}_{\theta} \to \mu_2}\right)_\# \pivot \right)
\end{align}

Hence, the third term in \eqref{eq:swgg of shifted measures} is:

\begin{align}
    W^2_2(\tilde{\mu}^{1 \to 2}_{g,\theta}, \tilde{\mu}^{1 \to 2}_{\theta})=W^2_2(\lgbarycenter,\pivot)-2 \Bigl < \frac{\boldsymbol{u}+\boldsymbol{v}}{2},\frac{\boldsymbol{m}_1+\boldsymbol{m}_2}{2}-\boldsymbol{m}_3 \Bigr> + \Bigl \|\frac{\boldsymbol{u}+\boldsymbol{v}}{2} \Bigr \|^2_2
\end{align}
since the mean of a Wasserstein mean is the mean of $m_1$, $m_2$.

Putting all together, we have: 
\begin{align}
    \mswgg^2_2(T^u_\#\mu_1,T^v_\#\mu_2)&=\mswgg^2_2(\mu_1,\mu_2)&-&4 \langle \boldsymbol{u},\boldsymbol{m}_1-\boldsymbol{m}_3\rangle -4 \langle \boldsymbol{v},\boldsymbol{m}_2-\boldsymbol{m}_3\rangle \\
    &&&+ 8 \Bigl< \frac{\boldsymbol{u}+\boldsymbol{v}}{2},\frac{\boldsymbol{m}_1+\boldsymbol{m}_2}{2}-\boldsymbol{m}_3 \Bigr> \nonumber \\
    &&&+2\|\boldsymbol{u}\|^2_2+2\|\boldsymbol{v}\|^2_2-4\Bigl \|\frac{\boldsymbol{u}+\boldsymbol{v}}{2}\Bigr \|^2_2\nonumber\\
    &=\mswgg^2_2(\mu_1,\mu_2)&+&4\langle \boldsymbol{u}+\boldsymbol{v},\boldsymbol{m}_3\rangle\\
    &&&-4\langle \boldsymbol{u}+\boldsymbol{v} ,\boldsymbol{m}_3\rangle-4\langle \boldsymbol{u},\boldsymbol{m}_1\rangle -4 \langle \boldsymbol{v},\boldsymbol{m}_2\rangle \nonumber \\
    &&&+4\langle \boldsymbol{u}+\boldsymbol{v},\boldsymbol{m}_1+\boldsymbol{m}_2\rangle +\|\boldsymbol{u}-\boldsymbol{v}\|^2_2 \tag{Parallelogram law}\nonumber\\
    &=\mswgg^2_2(\mu_1,\mu_2)&-&2\langle \boldsymbol{u},\boldsymbol{m}_1\rangle-2 \langle \boldsymbol{v},\boldsymbol{m}_2\rangle +2\langle \boldsymbol{u},\boldsymbol{m}_2\rangle +2\langle \boldsymbol{v},\boldsymbol{m}_1\rangle\\
    &&&+\|\boldsymbol{u}-\boldsymbol{v}\|^2_2\nonumber\\
    &=\mswgg^2_2(\mu_1,\mu_2)&-&2\langle \boldsymbol{u}-\boldsymbol{v},\boldsymbol{m}_1 - \boldsymbol{m}_2\rangle+\|\boldsymbol{u}-\boldsymbol{v}\|^2_2
\end{align}

\subsection{Proof of the new closed form of the Wasserstein distance (Lemma \ref{lemma:closed form})}
\label{app:proof lemma closed form}

We recall and prove the lemma that makes explicit a new closed form for WD.
Let $\mu_1,\mu_2$ be in $\Pdn$ with $\mu_2$ a distribution supported on a line whose direction is $\theta \in \Sd$. We have:
    \begin{align}
        W^2_2(\mu_1,\mu_2)=W^2_2(\mu_1,Q_\#^\theta\mu_1)+W^2_2(Q_\#^\theta\mu_1,\mu_2).
    \end{align}
Moreover, the optimal map is given by $T^{1\to 2} =  T^{Q^\theta_\# \mu_1 \to \mu_2} \circ T^{\mu_1 \to Q^\theta_\# \mu_1}=T^{Q^\theta_\# \mu_1 \to \mu_2}\circ Q^\theta$.

Let $\mu_1,\mu_2$ be in $\Pdn$ with $\mu_2$ a distribution supported on a line of direction $\theta$. We have:
\begin{align}
        W^2_2(\mu_1,\mu_2)=W^2_2(\mu_1,Q_\#^\theta\mu_1)+W^2_2(Q_\#^\theta\mu_1,\mu_2)
\end{align}
Moreover, the optimal map is given by:
\begin{align}
        T^{1 \to 2}=  T^{Q^\theta_\# \mu_1 \to 2} \circ T^{1 \to Q^\theta_\# \mu_1}=T^{Q^\theta_\# \mu_1 \to 2}\circ Q^\theta 
\end{align}
Here $Q^\theta$ is given in Def. \ref{def:pivot measure} of the paper. 

The proof of the Lemma was first inspired by \cite{buttazzo2022wasserstein}(Proposition 2.3), where authors show that $W^2_C(\mu_1,\mu_2)=W^2_{C^1}(\mu_1,\mu)+W^2_{C^2}(\mu,\mu_2)$ , with $C^1,C^2$ and $C$ some cost matrices with the constraints $C_{ij}=\min_sC^1_{is}+C^2_{sj}$.

Let $\mu_1=\frac{1}{n}\sum \delta_{\boldsymbol{x}_i}$ and $\mu_2=\frac{1}{n} \delta_{\overline{\boldsymbol{y}}_i}$ be in $\Pdn$ with $\mu_2$ a distribution supported on a line with direction $\theta$. Let $Q_\#^\theta \mu_1=\overline{\mu}_1=\frac{1}{n}\sum \delta_{\overline{\boldsymbol{x}}_i} \in \Pdn$. We emphasize here the fact that the atoms of $\overline{\mu}_1$ and $\mu_2$ are supported on a line are denoted by the overline symbol.

From one side, we have:
\begin{align}
    W^2_2(\mu_1,\mu_2)&=\inf_{T^1 \; \text{s.t.} \; T^1_\# \mu_1= \mu_2} \int_{\R^d}\|\boldsymbol{x}-T^1(\boldsymbol{x})\|^2_2d\mu_1(\boldsymbol{x})\\
    \label{app:pythagors}
    &=\inf_{T^1 \; \text{s.t.} \; T^1_\# \mu_1= \mu_2} \int_{\R^d} (\|\boldsymbol{x}-Q^\theta(\boldsymbol{x})\|^2_2+\|Q^\theta(\boldsymbol{x})-T^1(\boldsymbol{x})\|^2_2 \rgg{)}d\mu_1(\boldsymbol{x})\\
    &=\int_{\R^d}\|\boldsymbol{x}- Q^\theta(\boldsymbol{x})\|^2_2d\mu_1(\boldsymbol{x})+\inf_{T^1 \; \text{s.t.} \; T^1_\# \mu_1= \mu_2}\int_{\R^d}\|Q^\theta(\boldsymbol{x})-T^1(\boldsymbol{x})\|^2_2d\mu_1(\boldsymbol{x})\\
    &\geq\inf_{T^2 \; \text{s.t.} \; T^2_\# \mu_1= \overline{\mu}_1}\int_{\R^d}\|\boldsymbol{x}- T^2(\boldsymbol{x})\|^2_2d\mu_1(\boldsymbol{x})+\inf_{T^3 \; \text{s.t.} \; T^3_\# \overline{\mu}_1= \mu_2}\int_{\R^d}\|\overline{\boldsymbol{x}}-T^3(\overline{\boldsymbol{x}})\|^2_2d\overline{\mu}_1(\overline{\boldsymbol{x}})\\
    \label{app:eq 45}
    &\geq W^2_2(\mu_1,\overline{\mu}_1)+W^2_2(\overline{\mu}_1,\mu_2)
\end{align}

Equation \eqref{app:pythagors} is obtained thanks to the Pythagorean theorem since $\langle \boldsymbol{x}_i,Q^\theta(\boldsymbol{x}_i),\overline{\boldsymbol{y}_i}\rangle$ is a right triangle $\forall 1 \leq i\leq n$. The equation \eqref{app:eq 45} is obtained by taking the $\inf$ of the previous first term of the previous equation.

From the other side:
\begin{align}
\label{app:pythagors2}
    W^2_2(\mu_1,\overline{\mu}_1)+W^2_2(\overline{\mu}_1,\mu_2)=& \int_{\R^d} \|\overline{\boldsymbol{x}}-T^3(\overline{\boldsymbol{x}})\|^2_2d\overline{\mu}_1(\overline{\boldsymbol{x}})+\int_{\R^d}\|\overline{\boldsymbol{x}}-T^4(\overline{\boldsymbol{x}})\|^2_2d\overline{\mu}_1(\overline{\boldsymbol{x}})\\
    =&\int_{\R^d}\|T^3(\overline{\boldsymbol{x}})-T^4(\overline{\boldsymbol{x}})\|^2_2 d\overline{\mu}_1(\overline{\boldsymbol{x}})\\
    =&W^2_{\overline{\mu}_1}(\mu_1,\mu_2) \geq W^2_2(\mu_1,\mu_2)
\end{align}

Where $T^3$ and $T^4$ are the optimal plan of $W^2_2(\mu_1,\overline{\mu}_1)$ and$+W^2_2(\overline{\mu}_1,\mu_2)$. Similarly, \eqref{app:pythagors2} is obtained via the Pythagorean theorem.
This concludes the proof. 

\begin{figure}[ht]
    \centering
    \includegraphics[scale=0.4]{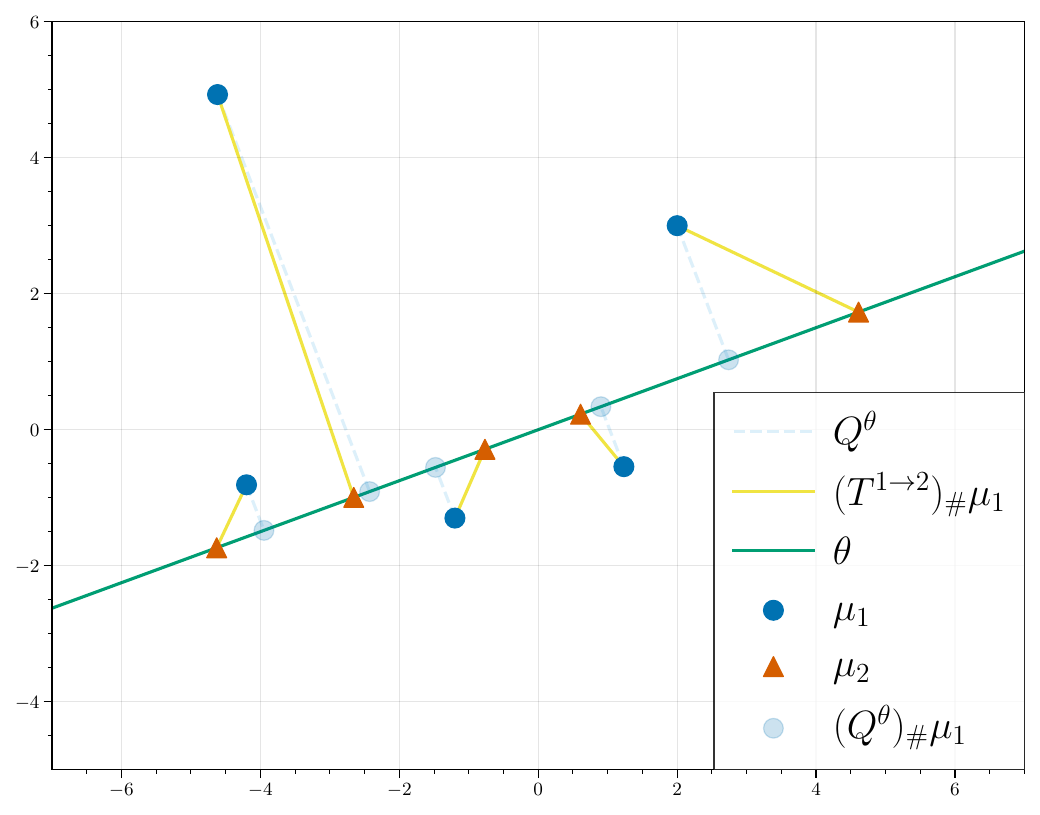}
    \caption{Closed form for Wasserstein with Pythagorus theorem}
    \label{fig:closed form Wasserstein}
\end{figure}

We plot an illustration of the lemma in Figure \ref{fig:closed form Wasserstein}.

\subsection{Details on the efficient computation of SWGG}
\label{app:fast computation}

We decompose the second formulation of $\swgg$. Let first remind that $Q^\theta : \R^d \to \R^d$, $\boldsymbol{x}\mapsto \theta \langle \boldsymbol{x},\theta \rangle$ and $P^\theta: \R^d \to \R$, $\boldsymbol{x} \mapsto \langle \boldsymbol{x} , \theta \rangle$ are the projections on the subspace generated by $\theta$.

We have:
\begin{align}
    \swgg_2^2(\mu_1,\mu_2,\theta)=2W^2_2(\mu_1,\pivot)+2W^2_2(\pivot,\mu_2)-4W^2_2(\lgbarycenter,\pivot). 
\end{align}
First, by lemma \ref{lemma:closed form},
\begin{align}
    2W^2_2(\mu_1,\pivot)=2W^2_2(\mu_1,\thetapp\mu_1)+2W^2_2(\thetap \mu_1,\thetap\pivot) 
\end{align}
as $\pivot$'s support is on a line.
Similarly,
\begin{align}
    2W^2_2(\mu_2,\pivot)=2W^2_2(\mu_2,\thetapp\mu_2)+2W^2_2(\thetap\mu_2,\thetap\pivot).
\end{align}
and
\begin{align}
    -4W^2_2(\lgbarycenter,\pivot)=-4W^2_2(\lgbarycenter,\thetapp\lgbarycenter)-4W^2_2(\thetap\lgbarycenter,\thetap\pivot).
\end{align}

We notice that $2W^2_2(\thetap \mu_1,\thetap \pivot)+2W^2_2(\thetap \pivot,\thetap \mu_2)=W^2_2(\thetap \mu_1,\thetap \mu_2)$ (as $\thetap \pivot$ is the Wasserstein mean between $\thetap \mu_1$ and $\thetap \mu_2$). We also notice that $-4W^2_2(\thetap\lgbarycenter,\thetap\pivot)=0$ (it comes from the fact that the generalized Wasserstein mean is induced by the permutations on the line), we can put all together to have:
\begin{align}
    \swgg^2_2(\mu_1,\mu_2,\theta)=2W^2_2(\mu_1,\thetapp\mu_1)+2W^2_2(\mu_2,\thetapp \mu_2)-4W^2_2(\lgbarycenter,\thetapp \lgbarycenter)+W^2_2(\thetap \mu_1,\thetap \mu_2)
\end{align}
One can show that $\swgg$ is divided into 3 Wasserstein distances between a distribution and its projections on a line and 1D Wasserstein problem. This results in a very fast computation of $\swgg$.


\subsection{Smoothing of SWGG}
\label{app:smooth SWGG}
In this section, we give details on the smoothing procedure of $\mswgg$, an additional landscape of $\swgg$ and its smooth counterpart $\widetilde{\swgg}$ and an empirical heuristic for setting hyperparameters $s$ and $\epsilon$.

\paragraph{Smoothing Procedure.}
A natural surrogate would be to add an entropic regularization within the definition of $T^{\pivot \to \mu_i}$, $i \in\{1,2\}$ and to solve an additional optimal transport problem. Nevertheless, it would lead to an algorithm with an $\mathcal{O}(n^2)$ complexity. Instead, we build upon the blurred Wasserstein distance \cite{feydy2020geometric} between two distributions $\nu_1$ and $\nu_2$:
\begin{align*} 
 B^2_{\epsilon}({\nu}_1, \nu_2) \defeq W_2^2(k_{\epsilon/4} \ast {\nu}_1, k_{\epsilon/4} \ast \nu_2)
\end{align*}
where $\ast$ denotes the smoothing (convolution) operator and $k_{\epsilon/4}$ is the Gaussian kernel of deviation $\sqrt{\epsilon}/2$. In our case, it resorts in making $s$ copies of each sorted projections $P^\theta(\boldsymbol{x}_i)$ and $P^\theta (\boldsymbol{y}_i)$ respectively, to add a Gaussian noise of deviation $\sqrt{\epsilon}/2$ and to compute averages of sorted blurred copies $\boldsymbol{x}^s_{\sigma^s}$, $\boldsymbol{y}^s_{\tau^s}$:
\begin{equation}
\label{eq:tildepivot}
(\widetilde{\pivot})_i = \frac{1}{2s} \sum_{k=(i-1)s+1}^{is} \boldsymbol{x}^s_{\sigma^s(k)}+\boldsymbol{y}^s_{\tau^s(k)}.
\end{equation}

Further, we provide additional examples of the landscape of $\widetilde{\mswgg}({\mu}_1,{\mu}_2)$ and discuss how to choose empirically relevant $s$ and $\epsilon$ values. 

\cite{feydy2020geometric} has shown that the blurred WD has the same asymptotic properties as the Sinkhorn divergence, with parameter $\epsilon$ the strength of the blurring: it interpolates between WD (when $\epsilon \to 0$) and a degenerate constant value (when $\epsilon \to \infty$). 

To find a minimum of Eq. \eqref{eq:defswggsmooth} in the paper (i.e. $\widetilde{\swgg_2^2}({\mu}_1,{\mu}_2,\theta)$), we iterate over:
\begin{align*}
  \theta_{t+1} =& \theta_{t}+\eta\nabla_{\theta} \widetilde{\swgg^2_2}(\mu_1,\mu_2,\theta)\\
  \theta_{t+1} =& \theta_{t+1}/\|{\theta}_{t+1}\|_2
\end{align*}
where $\eta \in \R_+$ is the learning rate. This procedure converges towards a local minima with a complexity of $\mathcal{O}(snd+sn\log(sn))$ for each iteration.
Once the optimal direction $\theta^\star$ is found, the final solution resorts to be the solution provided by $\swgg^2_2(\mu_1,\mu_2,\theta^\star)$, where the induced optimal transport map is an unblurred matrix.

\paragraph{Heuristic for setting the hyperparameters of $\widetilde{\swgg}$} We here provide an heuristic for setting parameters $s$ (number of copies of each points) and $\epsilon$ (strength of the blurring). We then give an example of the behavior of $\widetilde{\swgg}$ w.r.t. these hyper parameters.

Let $\mu_1 =\frac{1}{n}\sum \delta_{\boldsymbol{x}_i}$ and $\mu_2 =\frac{1}{n}\sum \delta_{\boldsymbol{y}_i}$.

\textbullet \; $s\in \mathbb{N}_+$ represents the number of copies of each sample. We observe empirically that the quantity $sn$ should be large to provide a smooth landscape. It means that the $s$ values can be small when $n$ increases, allowing to keep a competitive algorithm (as the complexity depends on $ns$)

\textbullet \; $\epsilon \in \R_+$ represents the variance of the blurred copies of each sample. Empirically, $\epsilon$ should depend on the variance of the distributions projected on the line. Indeed, an $\epsilon$ very close to zero will not smooth enough the discontinuities whereas a large $\epsilon$ will give a constant landscape.

\begin{figure}[ht] 
    \centering
    \includegraphics[width=0.2\textwidth]{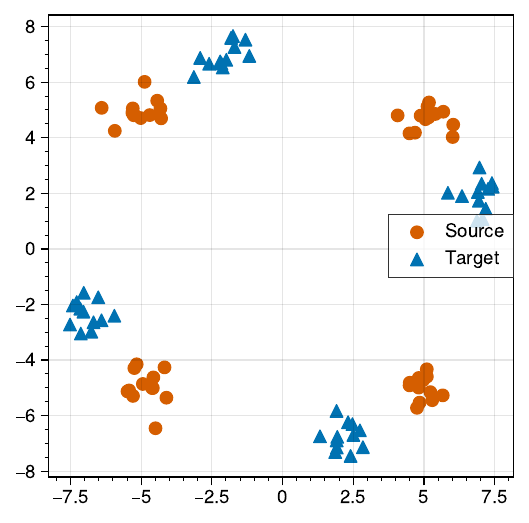}
    \includegraphics[width=0.5\textwidth]{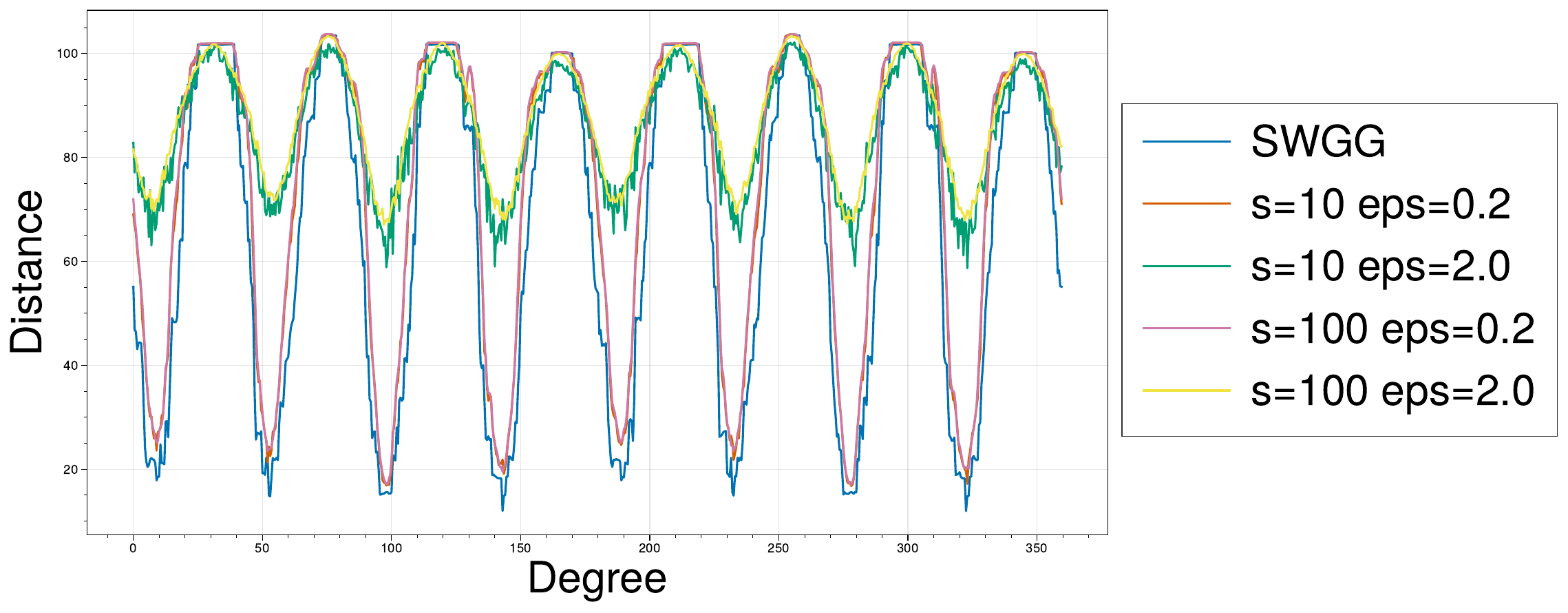}
    \vskip 0.2in
    \includegraphics[width=0.2\textwidth]{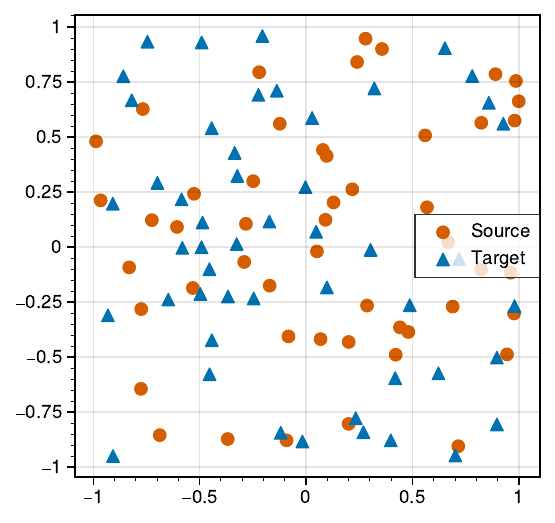}
    \includegraphics[width=0.5\textwidth]{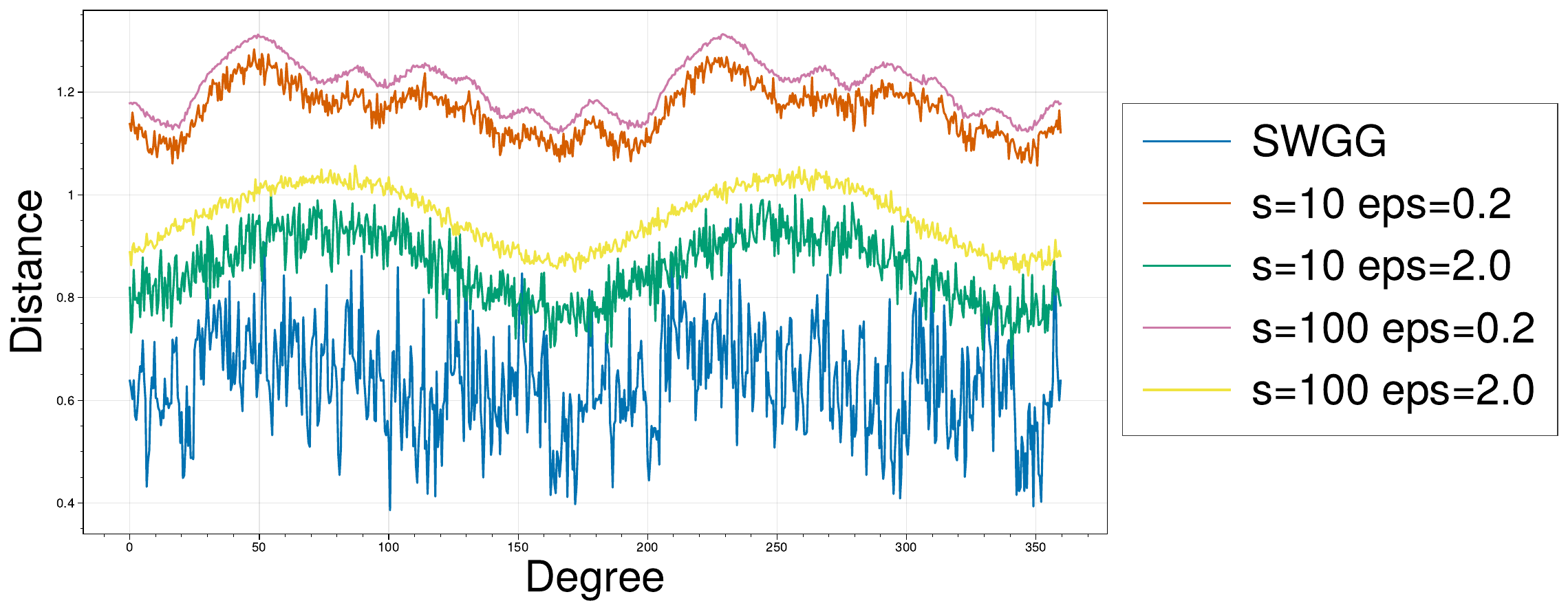}
\caption{Non-convex landscapes for $\swgg$ and  $\widetilde{\swgg}$ with different hyper parameters.}
\label{fig:landscape swgg}
\end{figure}

As discussed in Section \ref{sec:optimization scheme}, finding an optimal $\theta \in \Sd$ is a non convex problem and provides a discontinuous loss function. We give some examples of the landscape of $\widetilde{\swgg}$ w.r.t. different values of the hyperparameters in Fig. \ref{fig:landscape swgg}. The landscapes were computed with a set of projections $\theta$ regularly sampled with angles $\in[0, 2\pi]$.  

We observe that the larger $s$, the smoother  $\widetilde{\swgg}$. Additionally, raising $\epsilon$ tends to flatten $\widetilde{\swgg}$ w.r.t. $\theta$ (erasing local minima). Indeed similarly to Sinkhorn, a large $\epsilon$ blurred the transport plan and thus homogenize all the value of $\swgg$ w.r.t. $\theta$.

Moreover, we empirically observe that the number of samples for $\mu_1$ and $\mu_2$ enforces the continuity of $\swgg$. We then conjecture that the discontinuities of $\swgg$ are due to artifact of the sampling and thus the smoothing operation erases this unwanted behavior. A full investigation of this assumption is left for future work.

\subsection{Inconsequential of the pivot measure}
Importantly, only the direction $\theta$ is of importance for the value of $\swgg$. Indeed, whenever $\nu \in \Pdn$ is supported on a line of direction $\theta$, the position  of the atoms is irrelevant for $W_\nu$ and the associated transport plan whenever the atoms are distinct. Despite the fact that the pivot measure is inconsequential for the value of $\swgg$ (at $\theta$ fixed), we choose it to be $\pivot$. This choice is supported by the fact that $\pivot$ can be efficiently computed (as a 1D Wasserstein mean) and that some computation can be alleviated:
\begin{align}
    2W^2_2(\thetapp \mu_1,\pivot)+2W^2_2(\pivot, \thetapp \mu_2)=W^2_2(\thetapp \mu_1, \thetapp \mu_2)
\end{align}

It is an important comment to derive the property of distance for $\swgg$; it also allows minimizing $\swgg$ over $\theta \in \Sd$ without consideration for $\nu$, since any choice of $\nu$ supported on the subspace generated by $\theta$ give the same result for $\mswgg$. This property of irrelevance comes from the nature of the subspace where $\nu$ is supported, which is uni-dimensional. More formally we give the following proposition and its associated proof.

\begin{proposition}
\label{prop:irrelevance of pivot measure}
Let $\mu_1$, $\mu_2 \in \Pdn$. Let $\theta \in \Sd$. Let $\nu_1, \nu_2 \in \Pdn$ be two pivot measures supported on a line with direction $\theta$, with disctincs atoms for each measure. We then have:
\begin{align}
    W^2_{\nu_1}(\mu_1,\mu_2)=W^2_{\nu_2}(\mu_1,\mu_2)
\end{align}
\end{proposition}
We give a proof of this proposition.

Thanks to lemma \ref{lemma:closed form}, we known that the transport map $T_\nu^{1 \to 2}$ is fully induced by the transport plan $T_\nu^{Q^\theta_\# \mu_1 \to Q^\theta_\# \mu_2}$. Let remind that $T_\nu^{Q^\theta_\#\mu_1 \to Q^\theta_\# \mu_2}$ is given by $T^{\nu \to Q_\#^\theta\mu_2}\circ T^{Q_\#^\theta\mu_1\to \nu}$ (see equation \eqref{eq:composition transport plan generalized geodesic}). Moreover the two optimal transport plans are obtained via the ordering permutations, i.e. let $\sigma,\tau,\pi \in \Sn$ s.t:
\begin{align}
    \overline{\boldsymbol{x}}_{\sigma(1)} \leq ... \leq \overline{\boldsymbol{x}}_{\sigma(n)} \nonumber\\
    \overline{\boldsymbol{y}}_{\tau(1)} \leq ... \leq \overline{\boldsymbol{y}}_{\tau(n)} \nonumber\\
    \overline{\boldsymbol{z}}_{\pi(1)} \leq ... \leq \overline{\boldsymbol{z}}_{\pi(n)} \nonumber
\end{align}
With $\overline{\boldsymbol{x}}_i$ being the atoms of $Q^\theta_\#\mu_1$, $\overline{\boldsymbol{y}}_i$ the atoms of $Q^\theta_\#\mu_2$ and $\overline{\boldsymbol{z}}_i$ being the atoms of $Q^\theta_\#\nu$.

One have $T^{\mu_1 \to \nu}(\boldsymbol{x}_{\sigma(i)})=\boldsymbol{z}_{\pi(i)}$ (resp. $T^{\nu \to \mu_2}(\boldsymbol{z}_{\pi(i)})=\boldsymbol{x}_{\tau(i)})$ $\forall 1 \leq i \leq n$. Composing these two identities gives:
\begin{align}
    T_\nu^{1\to 2}(\boldsymbol{x}_{\sigma(i)})=\boldsymbol{y}_{\tau(i)} \quad \quad \forall 1 \leq i \leq n
\end{align}
The last equation shows that $T_\nu^{1 \to 2}$ is in fact independent of $\pi$ and thus of $\nu$.

\subsection{Proof that min-SWGG is a distance (generalized geodesic formulation)}
\label{app:proof prop distance}
This proof has already been established in \ref{app:distance swgg cp}. However we rephrase the proof in the context of generalized geodesics.

We aim to prove that $\swggr = \sqrt{2W^2_2(\mu_1,\pivot)+2W_2^2(\pivot,\mu_2)-4W_2^2(\lgbarycenter,\pivot)}$ defines a metric.

\textit{Finite and non-negativity.} Each term of $\swgg^2_2$ is finite thus the sum of the three terms is finite. Moreover, being an upper bound of WD makes it non-negative.

\textit{Symmetry.} We have 
\begin{eqnarray*}
\swgg^2_2(\mu_1,\mu_2,\theta)&=& 2W_2^2(\mu_1,\pivot)+2W^2_2(\mu_2,\pivot)-4W^2_2(\lgbarycenter,\pivot) \\
 &= & 2W_2^2(\mu_2,\pivot)+2W^2_2(\mu_1,\pivot)-4W^2_2(\lgbarycenter,\pivot) \\
  &= & \swgg^2_2(\mu_2,\mu_1,\theta).
\end{eqnarray*}

\textit{Identity property.} \\
From one side, when $\mu_1=\mu_2 \implies T^{\mu_1 \to \pivot}=T^{\mu_2 \to \pivot}=Id$, giving $\gene=\mu_1=\mu_2$. Thus:
\begin{align}
   \swgg^2_2(\mu_1,\mu_2,\theta)=2W^2_2(\mu_1,\pivot)+2W^2_2(\mu_1,\pivot)-4W^2_2(\mu_1,\pivot)=0 
\end{align}

From another side, $\swgg^2_2(\mu_1,\mu_2,\theta)=0 \implies W^2_2(\mu_1,\mu_2)=0 \implies \mu_1= \mu_2$ (by being an upper bound of WD).

\textit{Triangle Inequality.} 
We have:
\begin{align}
\swgg^2_2(\mu_1,\mu_2,\theta)&=&&2W^2_2(\mu_1,\pivot)+2W^2_2(\pivot,\mu_2)-4W^2_2(\lgbarycenter,\pivot)\\
&=&& 2\int_{\R^d}\| T_\theta^1(\boldsymbol{x})-\boldsymbol{x}\|_2^2d\pivot(\boldsymbol{x})+2\int_{\R^d}\|T_\theta^2(\boldsymbol{x})-\boldsymbol{x}\|^2_2 d\pivot(\boldsymbol{x})\\
&&&-4\int_{\R^d}\|T_\theta^g(\boldsymbol{x})-\boldsymbol{x}\|_2^2d\pivot(\boldsymbol{x})\nonumber\\
&=&&\int_{\R^d} \left(2\| T_\theta^1(\boldsymbol{x})-\boldsymbol{x}\|_2^2+2\|T_\theta^2(\boldsymbol{x})-\boldsymbol{x}\|^2_2-4\|T_\theta^g(\boldsymbol{x})-\boldsymbol{x}\|_2^2\right) d\pivot(\boldsymbol{x})\\
&=&&\int_{\R^d} \|T_\theta^1(\boldsymbol{x})-T_\theta^2(\boldsymbol{x})\|_2^2d\pivot(\boldsymbol{x})
\end{align}

where, with an abuse of notation for clarity sake, $T_\theta^i$ is the optimal map between $\pivot$ and $\mu_i$ and $T_\theta^g$ is the optimal map between $\pivot$ and $\lgbarycenter$. The last line comes from the parallelogram rule of $\R^d$. Thanks to Proposition \ref{prop:irrelevance of pivot measure} we see that $\swgg$ is simply the $L^2(\R^d,\nu)$  square norm, i.e.:
\begin{align}
    \swgg^2_2(\mu_1,\mu_2,\theta)=\|T_\theta^1-T_\theta^2\|_{\nu}^2 \defeq \int_{\R^d}\|T_\theta^1-T_\theta^2\|^2_2d\nu
\end{align}
with $\nu$ being any arbitrary pivot measure of $\Pdn$. And thus $\swggr$ is the $L^2(\R^d,\nu)$ norm. This observation is enough to conclude that $\swggr$ is a proper distance for $\theta$ fixed.

\section{Experiment details and additional results}
\label{app:experimentation}

WD, SW, Sinkhorn, Factored coupling are computed using the  \texttt{Python OT} Toolbox \cite{JMLR:POT} and our code is available at \url{https://github.com/MaheyG/SWGG}. The Sinkhorn divergence for the point cloud matching experiment was computed thanks to the \texttt{Geomloss} package \cite{feydy2019interpolating}.

\subsection{Behavior of min-SWGG with the dimension and the number of points}
\label{app:sanity check and behavior}

In this section, we draw two experiments to study the behavior of $\mswgg$ w.r.t. the dimension and to the number of points. 

\begin{figure}[ht]
    \centering
    \includegraphics[scale=0.3]{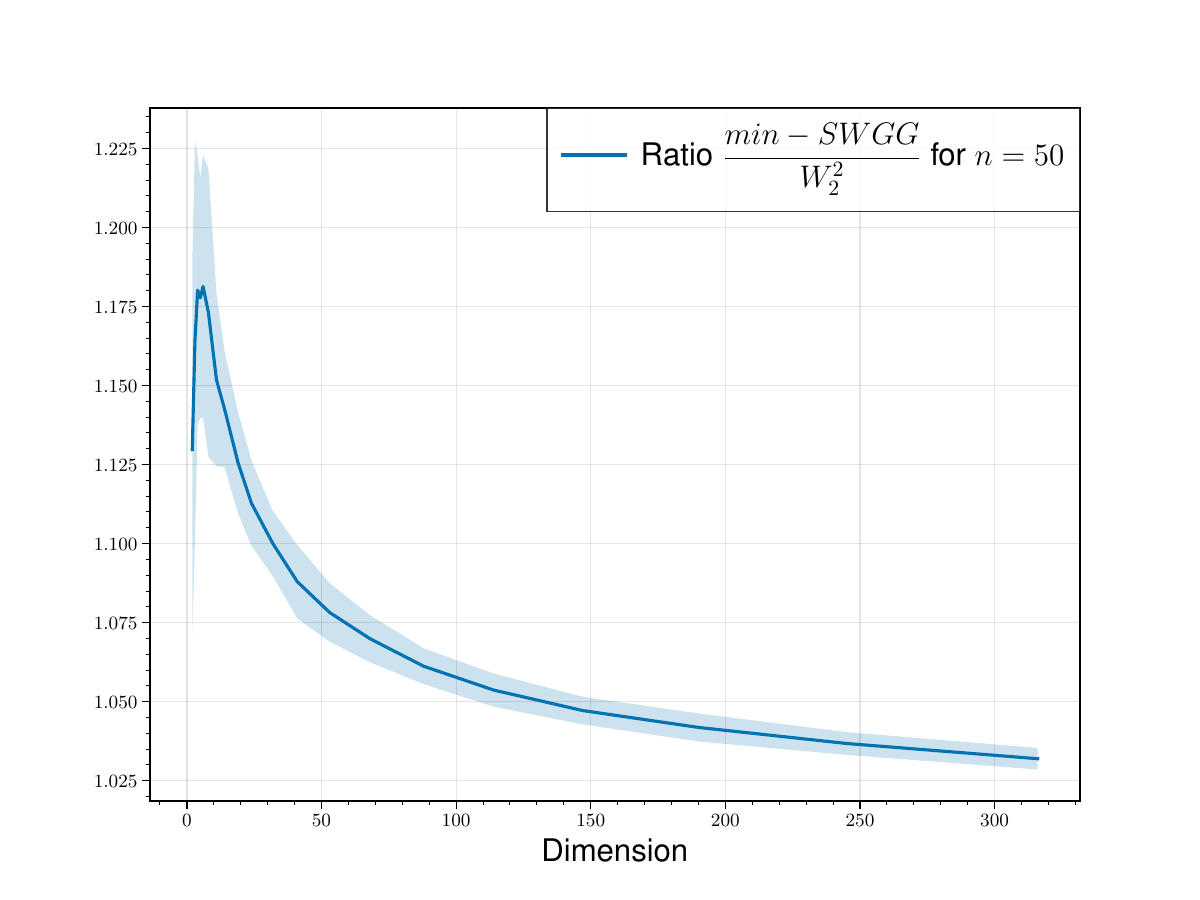}
    \includegraphics[scale=0.3]{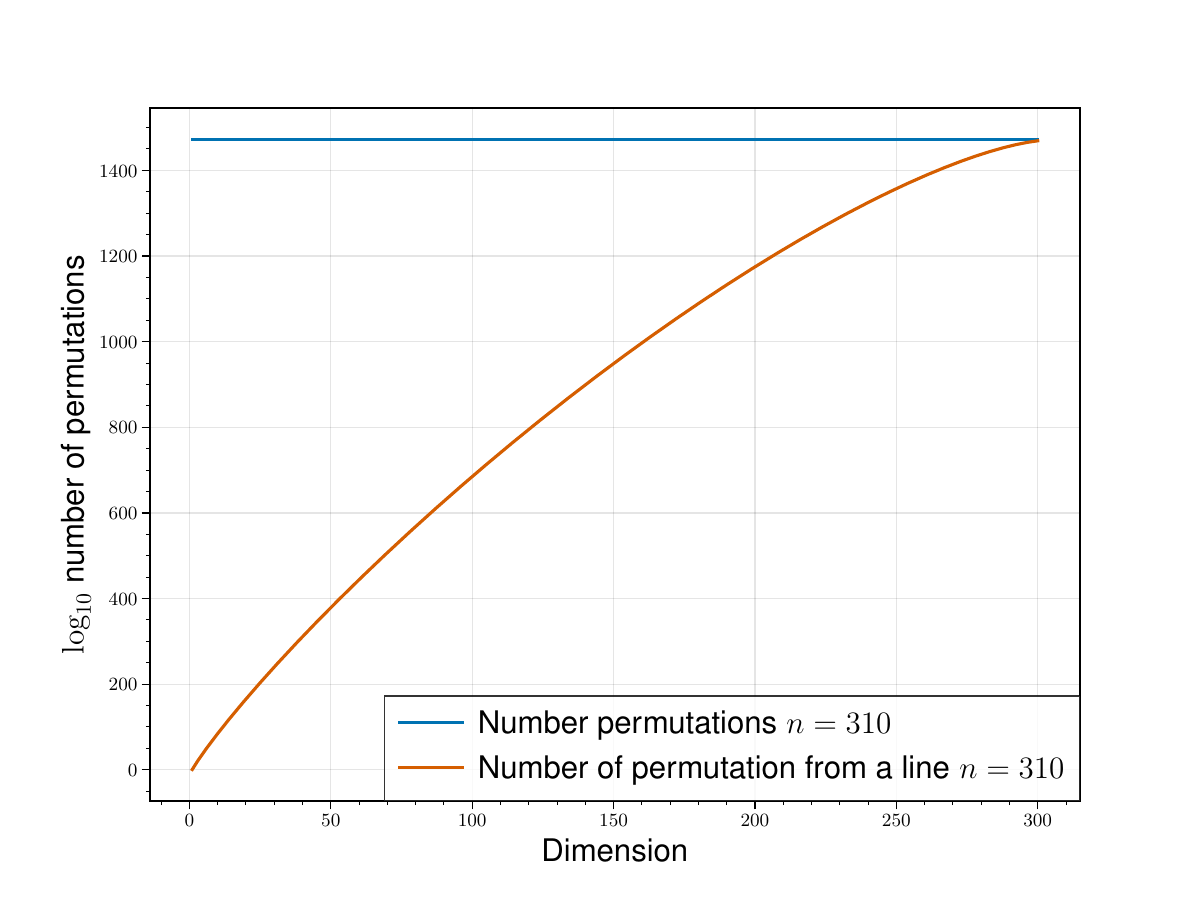}
    \caption{Evolution of  $W^2_2$ and $\mswgg_2^2$ with the dimension $d$ for isotropic Gaussian distributions (left) Number of permutations induced by a direction $\theta \in \Sd$ with $n=310$ and a varying dimension (right)}
    \label{fig:swgg with d}
\end{figure}

\paragraph{Evolution with $d$}
In \cite{cover1967number}[Theorem of Section 2], authors aim at enumerate the number of permutations obtained via the projection of  point clouds on a line. It appears that the number of permutations increases with the dimension. They even show that whenever $d\geq 2n$ ($2n$ being the total number of points of the problem), all the possible permutations ($n!$) are in the scope of a line. Fig. \ref{fig:swgg with d} depicts the number of obtainable permutations as a function of the dimension $d$, for $n$ fixed. This theorem can be applied to $\mswgg$ to conclude that whenever $d \geq 2n$, we have $\mswgg^2_2=W^2_2$.

It turns out empirically that the greater the dimension, the better the approximation of $W^2_2$ with $\mswgg$ (see Fig. \ref{fig:swgg with d}) for a fixed $n$. More formally, the set of all possible transport maps is called the Birkhoff polytope and it is known that the minimum of the Monge problem is attained at the extremal points (which are exactly the set of permutations matrices, a set of $n!$ matrices in our context) \cite{birkhoff1946tres}. The set of the transport maps in the scope of $\swgg$ is a subset of the extremal points of the Birkhoff polytope (there are permutations matrices but not all possibilities are represented). Theoretically, the set of transport maps in the scope of $\swgg$ is larger as $d$ grows, giving a subset that is more and more tight with the extremal points of the Birkhoff polytope. This explains that $\mswgg$ can benefit from higher dimension.

We plot in Fig. \ref{fig:swgg with d} the evolution, over 50 repetitions, of the ratio $\frac{\mswgg(\mu_1,\mu_2)}{W^2_2(\mu_1,\mu_2)}$ with $d$, $n=50$ and $\mu_1 \sim \mathcal{N}(1_{\R^d},Id)$, $\mu_2\sim \mathcal{N}(-1_{\R^d},Id)$.

\begin{figure}[ht]
    \centering
    \includegraphics[scale=0.4]{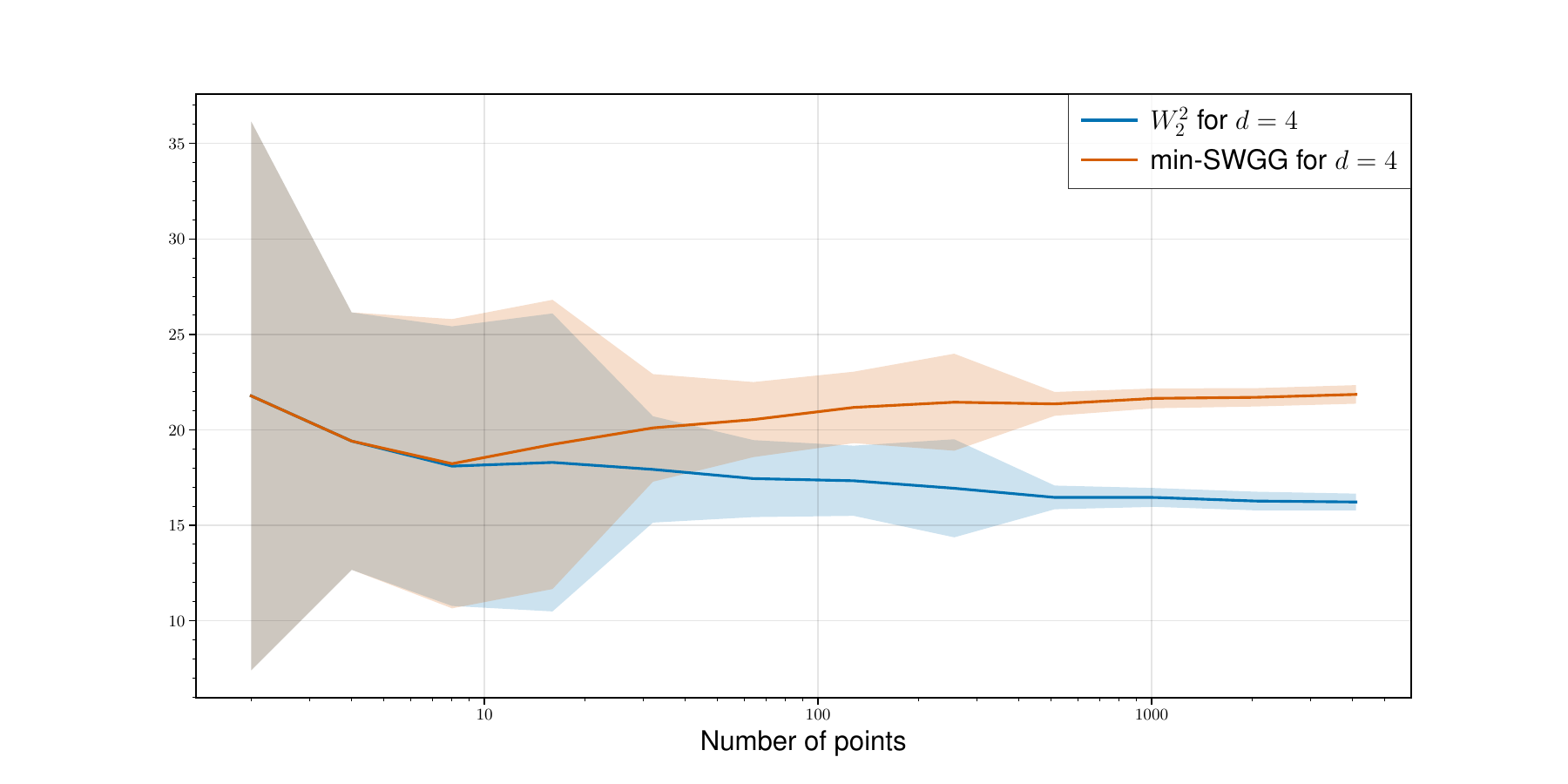}
    \caption{Evolution of $W^2_2$ and $\mswgg$ w.r.t. the number of points}
    \label{fig:swgg with n}
\end{figure}

\paragraph{Evolution with $n$}
Fig. \ref{fig:swgg with n} represents the evolution of $W^2_2(\mu_1,\mu_2)$ and $\mswgg_2^2(\mu_1,\mu_2)$ for two distributions $\mu_1 \sim \mathcal{N}(1_{\R^d},Id)$ and $\mu_2\sim \mathcal{N}(-1_{\R^d},Id)$, with $d=4$ and a varying number of points. The results are averages over 10 repetitions.


We observe that, when $n$ is large enough, $\mswgg$ tends to stabilize around some constant value. 

We conjecture that there may exist an upper bound for $\mswgg$:
\begin{equation}
    \mswgg^2_2(\mu_1,\mu_2)\leq \psi(d,n,d') W^2_2(\mu_1,\mu_2)
\end{equation}
Where $d'$ is the max of the dimensions of the distributions $\mu_1,\mu_2$ \cite{weed2019sharp}, and $\psi$ an unknown function.

\subsection{Computing min-SWGG}
\label{app:monte-carlo vs optim}

We now provide here more details about the experimental setup of the experiments of Section \ref{sec:monte-carlo vs optim}. 

\paragraph{Choosing the optimal $\theta$} We compare three variants for choosing the optimal direction $\theta$: random search, simulated annealing and optimization (defined in Section \ref{sec:efficient computation + optimization scheme}). We choose to compare with simulated annealing since it is widely used in discrete problem (such as the travelling salesman) and known to perform well in high dimension \cite{van1987simulated} \cite{chibante2010simulated} \cite{kirkpatrick1983optimization}. 
We notice in Fig. \ref{fig:mc_vs_optim} of the paper that the smooth version of $\mswgg$ is always (comparable or) better than the simulated annealing. 
In this experiment, we randomly sample 2 Gaussian distributions with different means and covariances matrices, whose parameters are chosen randomly. For optimizing $\mswgg$, we use the Adam optimizer of Pytorch, with a fixed learning rate of $5e^{-2}$ during $100$ iterations, considering $s=10$ and $\epsilon=1$.

\begin{figure}[ht]
    \centering
    \includegraphics[scale=0.6]{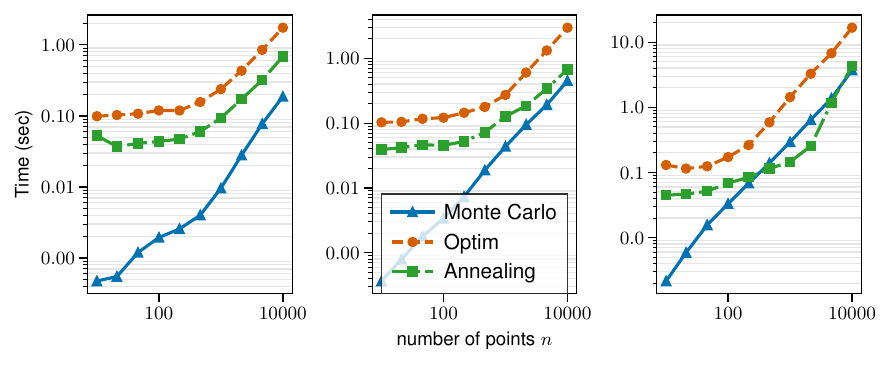}
    \caption{Considering two Gaussian distributions in dimensions $d$ equals to: 2 (left), 20 (middle), 200 (right),we compute $\mswgg$ with random search, simulated annealing schemes and optimization procedure and report the timings for varying number of points and fixed number of projections.}
    \label{fig:mc_vs_optim_time}
\end{figure}

Fig. \ref{fig:mc_vs_optim_time} provides the timings for computing the random search approximation, simulated annealing and the optimization scheme. In all cases, we recover the linear complexity of $\mswgg$ (blue curves) in a log space. For the computation timings we compute $\mswgg$ with random search with $L=500$, simulated annealing (green curves) with 500 iterations with a temperature scheme $(1-\frac{k+1}{500})_{k=1}^{500}$ and the optimization scheme (considering $s=10$ with a fixed number of iterations for the optimization scheme equals to 100).

\begin{figure}[ht]
    \centering
    \includegraphics[scale=0.45]{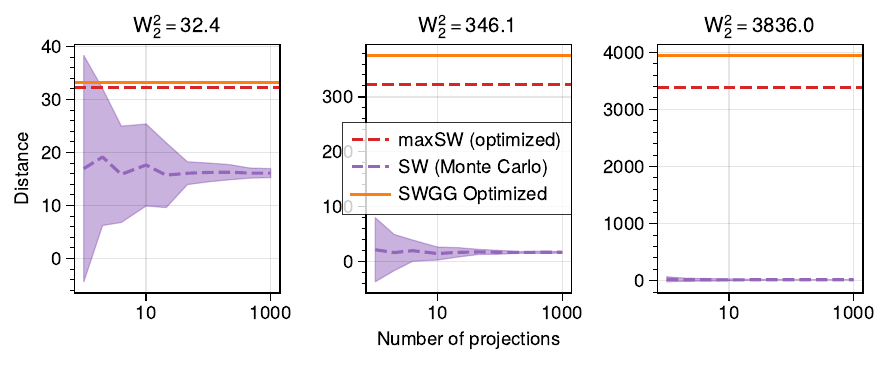}
    \includegraphics[scale=0.45]{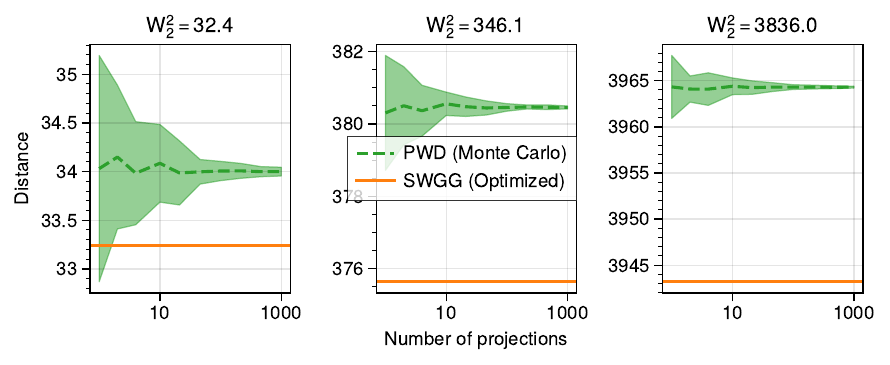}
    \caption{Comparison of $\mswgg$ optim with $\pwd$ (left) and with $\maxsw$ and $\sw$ (right). $\pwd$ and $\sw$ are computed with a growing number of projection}
    \label{fig:optim vs monte carlo competitor}
\end{figure}

Additionally, we reproduce the same setup as in \ref{sec:monte-carlo vs optim} for the $\sw$, $\maxsw$ and $\pwd$ distance. For sake of readability we compared with $\mswgg$ optim and report the results in Fig. \ref{fig:optim vs monte carlo competitor}.

\paragraph{Runtime Evaluation}
In the paper, on Fig. \ref{fig:mc_vs_optim} (Right), we compare the empirical runtime evaluation on GPU for different methods. We consider Gaussian distributions in dimension $d=3$ and we sample $n$ points per distribution with $n\in\{10^2,10^3,10^4,5\cdot 10^4, 10^5\}$. For $\sw$ Monte-Carlo and $\mswgg$ random search, we use $L=200$ projections. For both $\maxsw$ and $\mswgg$ with optimization, we use 100 iterations with a learning rate of 1, and we fix $s=50$ for $\mswgg$. We use the official implementation of the Subspace Robust Wasserstein (SRW) with the Frank-Wolfe algorithm \cite{paty2019subspace}.

\subsection{Gradient Flows}
We rely on the code provided with \cite{kolouri2019generalized} for running the experiment of Section \ref{sec:gradient flows}.

We fix $n=100$, the source distribution is taken to be Gaussian and we consider four different target measures that represent several cases: i) a 2 dimensional Gaussian, ii) a 500 dimensional Gaussian (high dimensional case), iii) 8 Gaussians (multi-modal distribution) and iv) a two-moons distribution (non-linear case).

We fix a global learning rate of $5e^{-3}$ with an Adam optimizer. For $\sw$, $\pwd$ and $\swgg$ (random search), we sample $L=100$ directions. For the optimization methods $\maxsw$, we set a learning rate of $1e^{-3}$ with a number of 100 iterations for i), iii), and iv) and 200 iterations for ii). For $\mswgg$ (optimization), we took a learning rate of i)$1e^{-1}$, ii)$1e^{-3}$, iii)$5e^{-2}$, and iv) $1e^{-3}$. The hyper parameters for the optimization of $\mswgg$ are $s=10$ and $\epsilon=0.5$, except for the 500-dimensional Gaussian for which we pick $\epsilon=10$ .

Each experiment is run 10 times and shaded areas in Fig. \ref{fig:gradient flow} (see the main paper) represent the mean $\pm$ the standard deviation.

\subsection{Gray scale image colorization}
\label{app:colorization and pan}
We now provide additional results on a pan-sharpening application to complete results provided in Section \ref{sec:colorization}.

In pan-sharpening \cite{vivone2021benchmarking}, one aims at constructing a super-resolution multi-chromatic satellite image with the help of a super-resolution mono-chromatic image (source) and low-resolution multi-chromatic image (target). 

To realize this task, we choose to used a color transfer procedure, where the idea is to transfer the color palette from the target to the source image. This transfer is carry out by the optimal transport plan of the Wasserstein distance. More details on color transfer can be found in Supp. \ref{app:color transfer}.

\begin{figure}[ht]
    \centering
    \includegraphics[scale=0.2]{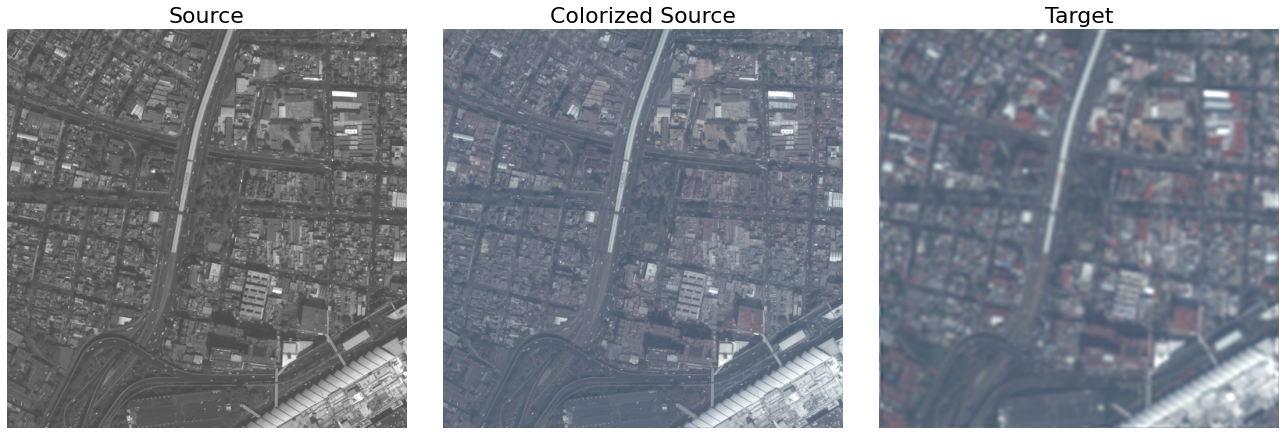}
    \includegraphics[scale=0.2]{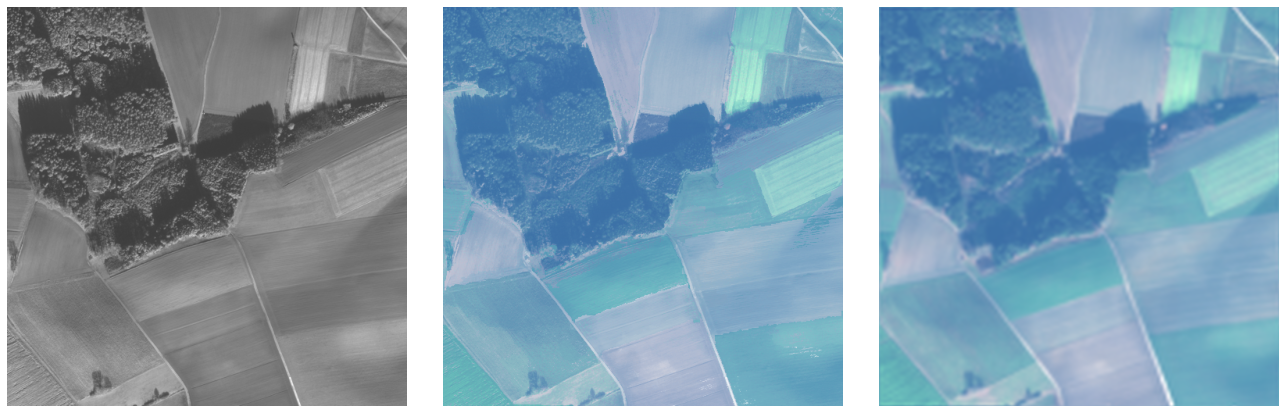}
    \includegraphics[scale=0.2]{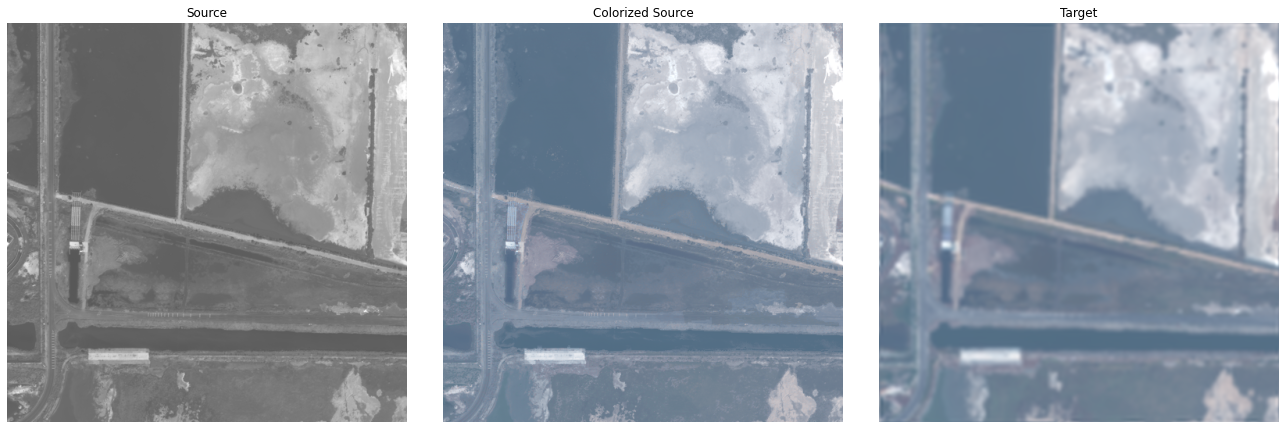}
    \includegraphics[scale=0.2]{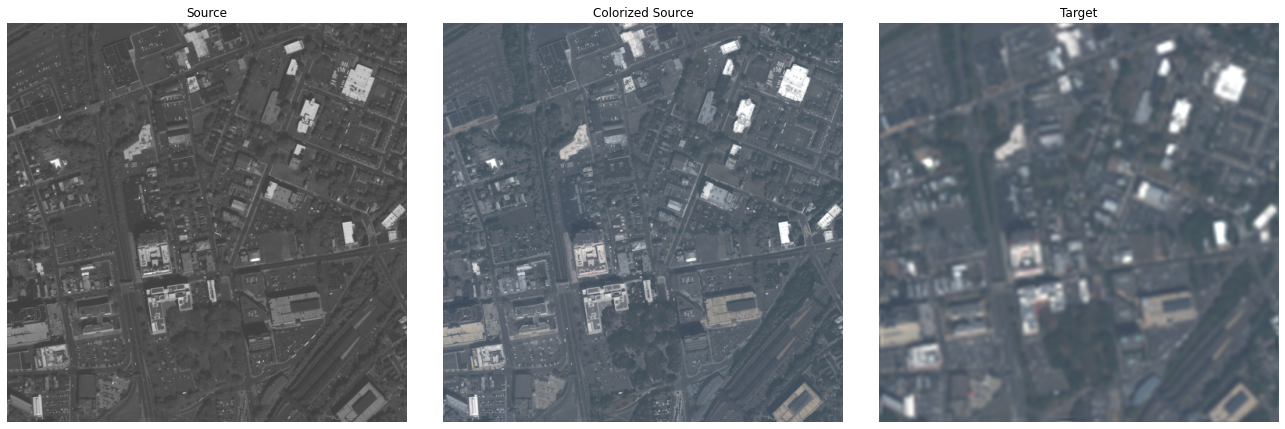}
    \caption{Source high resolution black and white image (left) Target low resolution colorful
image (right) Obtained high resolution colorful image (mid).}
    \label{fig:app pan}
\end{figure}

Additionally, we improve the relevance of the colorization by adding a proximity prior. For that, we used super pixels computed  via the Watershed algorithm \cite{neubert2014compact} thanks to the the \texttt{scikit-image} package \cite{van2014scikit}. Obtained high resolution colorized images of size 512$\times$512 ($n=262~144$) are reported on Fig. \ref{fig:app pan}.

All pan-sharpening experiments were run on the PairMax data set \cite{vivone2021benchmarking}. The hyperparameters (markers and compactness) for the watershed super-pixels are: 500, 200, 200, 200 markers (an upper bound for the number of super pixel) for each image (by order of apparition) and compactness $1e-8$ (high values result in more regularly-shaped watershed basins) for all the images. 

\subsection{Point clouds registration}
\label{app:ICP}
We here provide additional details and results about the experiments in Section \ref{sec:ICP}.

Authors of \cite{bonneel2019spot} highlighted the relevance of OT in the point clouds registration context, plugged into an Iterative Closest Point (ICP) algorithm. They leveraged the 1D partial OT without consideration for the direction of the line. Our experiment shows the importance of $\theta$: the smaller $\swgg$ is, the better the registration. 

In this experiment, having a one to one correspondence is mandatory: as such, we compare $\mswgg$ with a nearest neighbor assignment and the one provided by OT. Note that we do not compare $\mswgg$ with subspace detour \cite{muzellec2019subspace}, since: i) with empirical distributions, the reconstruction of the plan is degenerated (as it doesn't involve any computation),
ii) the research of subspace can be intensive as no prior is provided.

To create the source distributions, we used random transformation $(\Omega,t) \in O(d)\times \R^d$ of the target domain. $\Omega$ was picked randomly from $O(d)$, the set of rotations and reflections, and $t$ has random direction with $\|t\|_2=5$. We also add a Gaussian noise $\mathcal{N}(0,\epsilon Id)$, with $\epsilon=0.1$.

\begin{figure}[ht]
    \centering
    \includegraphics[scale=0.25]{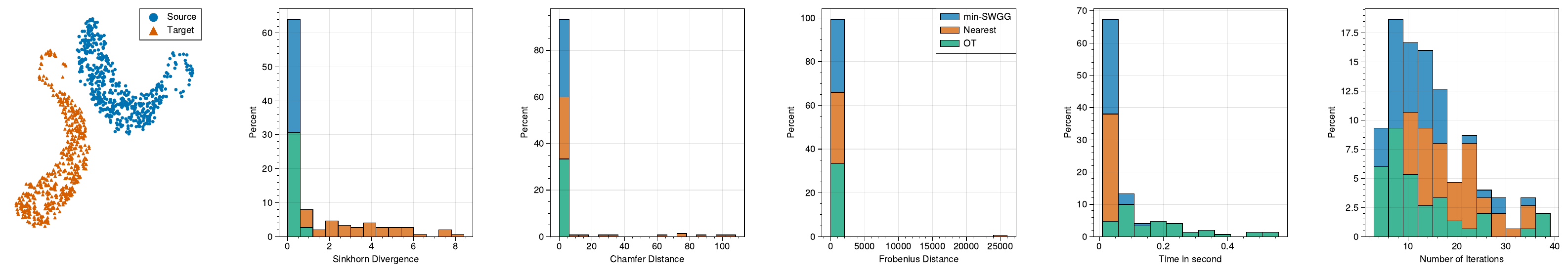}
    \includegraphics[scale=0.25]{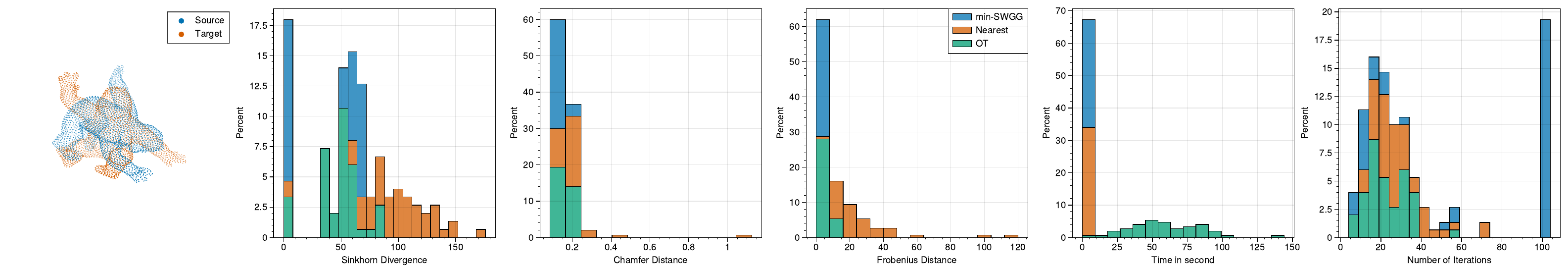}
    \includegraphics[scale=0.25]{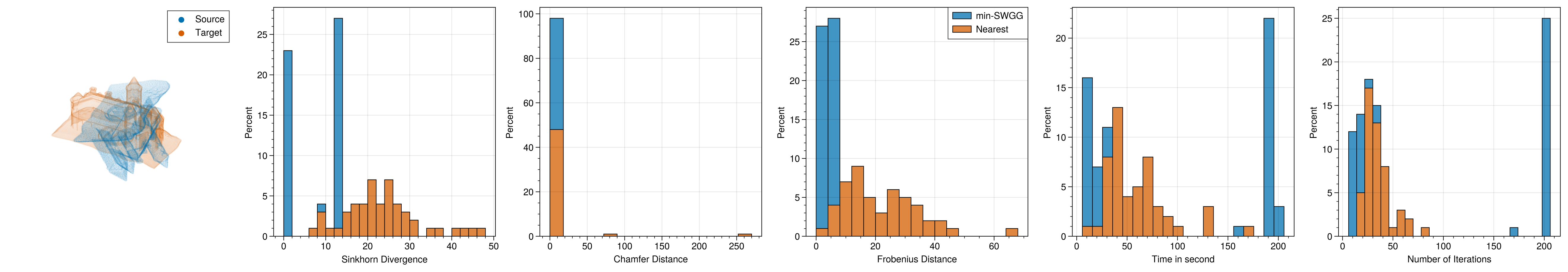}
    \caption{The three datasets (left) and the distributions of the Sinkhorn divergences, the Chamfer distance, the Frobenius distance, the timing and the number of iterations over 50 seeds (right).}
    \label{fig:ICP histo}
\end{figure}

The ICP algorithm was run with 3 datasets with the following features: i) 500 points in 2D, ii) 3000 points in 3D, and iii) 150~000 points in 3D. $\mswgg$ was computed through the random search estimation with $L=100$. A stopping criterion was the maximum number of iterations of the algorithm, which varies with the dataset \textit{i.e.}: i) 50, ii) 100, and iii) 200 respectively. The other stopping criterion is $\|\Omega -Id\|+\|t\|_2\leq \varepsilon$ with $\varepsilon$ chosen respectively for the datasets as follows: i) $1e^{-4}$, ii) $1e^{-2}$, and iii) $1e^{-2}$, where $(\Omega,t) \in O(d)\times \R^d$ is the current transformation and $\|\cdot\|$ is the Frobenius norm. All these settings were run with 50 different seeds. Results are reported in Fig. \ref{fig:ICP histo}.  

From Fig. \ref{fig:ICP histo}, one can see that for:
\begin{itemize}
    \item \textbf{$n=500$:}  The registration obtained via OT is very powerful (it is fast to compute and converges toward a good solution). $\mswgg$ is slightly faster with better convergence result and an equivalent number of iterations. Finally the nearest neighbor does not converge to a  solution closed to the target.
    \item \textbf{$n=3000$:} registration by OT  can converge poorly, moreover the timings are much higher than the competitors. $\mswgg$ shows efficient convergence results with an attractive computation time (order of fews seconds). We observe that the number of iterations can be very large and we conjecture that it is due to the fact that $\mswgg$-based ICP can exit local minima. The nearest neighbor is fast but, most of the time, does not converge to global minima (i.e. the exact matching of the shapes).
    \item \textbf{$n=150000$:} In this setting, OT is totally untractable (the cost matrix needs 180 GB in memory). $\mswgg$ shows good convergence and is most of the time very fast whenever the number of iterations does not attain the stopping criterion. The nearest neighbor assignment is faster but only converges to local minima.
\end{itemize}

Note that, despite the fact that $\mswgg$ is slightly slower than the nearest neighbor, the overall algorithm can be faster due to the better quality of each iteration ($\mswgg$ can attain a minimum with less iterations).

In Table \ref{table:chamfer frobenius} we give additional results on the final distribution. We control the final convergence via the square Chamfer distance and the square Frobenius distance. The Square Chamfer distance is a sqaure distance between point cloud defined as:
\begin{align}
    d^2(X,Y) = \sum_{x \in X} \operatorname*{min}_{y \in Y} ||x-y||^2_2 + \sum_{y \in Y} \operatorname*{min}_{x \in X} ||x-y||^2_2
\end{align}
and the Square Frobenius norm is a square distance between the transformation, defined as:
\begin{align}
\text{Fr}((\Omega_{\text{real}},t_{\text{real}}),(\Omega_{\text{estimated}},t_{\text{estimated}}))=\|\Omega_{\text{real}}-\Omega_{\text{estimated}}\|^2_2+\|t_{\text{real}}-t_{\text{estimated}}\|^2_2
\end{align}
In both cases we can see that the final results for $\mswgg$ are much better than the results for NN and relatively closed to the results from OT.

\begin{table}[ht]
    \centering
    \begin{tabular}{l c c c} 
    $n$ & 500 &3000&1500 00 \\
    \hline
    NN&11.65  &0.20&6.90 \\
    OT & \textbf{0.03} &0.16&$\cdot$ \\
    $\mswgg$ & 0.08 &\textbf{0.13}& $\boldsymbol{8\times10^{-4}}$\\
     \hline
    \end{tabular}
    \vskip 0.1in
    \begin{tabular}{l c c c} 
    $n$ & 500 &3000&150 000 \\
    \hline
    NN&526 &30.04&21.7 \\
    OT & 3.8 &6.5&$\cdot $\\
    $\mswgg$ & \textbf{2} &\textbf{4.5}&\textbf{6.01}\\
     \hline
    \end{tabular}
    \vskip 0.1 in
    \caption{Square Chamfer distance (top) and Square Frobenius distance (bottom) between final transformation on the source and the target. Best values are boldfaced.}
    \label{table:chamfer frobenius}
\end{table}

An other important aspect of ICP is that the algorithm tends to fall into local minima: the current solution is not good and further iterations do not allow a better convergence of the algorithm. We observed empirically that $\mswgg$ can avoid getting stuck on local minima when a reasonable number of directions $\theta$ is sampled ($L \sim 100$). We conjecture that 
the random search approximation is not always the ideal solution and hence may 
escape local minima. 
This may lead to a better convergence solution for $\mswgg$-based ICP.

\subsection{Color Transfer}
\label{app:color transfer}
In this section, we provide an additional experiment in a color transfer context.

We aim at adapting the color of an input image to match the color of a target one \cite{ferradans2013regularized, muzellec2019subspace}. This problem can be recast as an optimal transport problem where we aim at transporting the color of the source image $\boldsymbol{X}$ into the target $\boldsymbol{Y}$. For that, usual methods lie down on the existence of a map $T : \boldsymbol{X} \to \boldsymbol{Y}$. We challenge $\mswgg$ to this problem to highlight relevance of the obtained transport map.

Images are encoded as vector in $\mathbb{R}^{nm\times 3}$, where $n$ and $m$ are the size of the image and $3$ corresponds to the number of channels (here  RBG channels). We first compute a map $T_0 : \boldsymbol{X}_0 \to \boldsymbol{Y}_0$ between a subsample of $\boldsymbol{X}$ and $\boldsymbol{Y}$ of size 20000 and secondly extend this mapping to the complete distributions $T : \boldsymbol{X} \to \boldsymbol{Y}$ using a nearest neighbor interpolation. The subsampling step is mandatory due to the size of the images but can deteriorate the quality of the transfer. 

\begin{figure}[ht]
\centering
\includegraphics[scale=0.2]{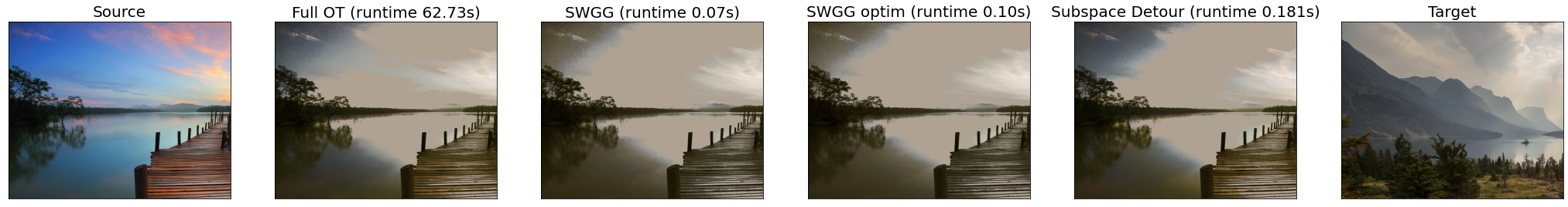}
\includegraphics[scale=0.2]{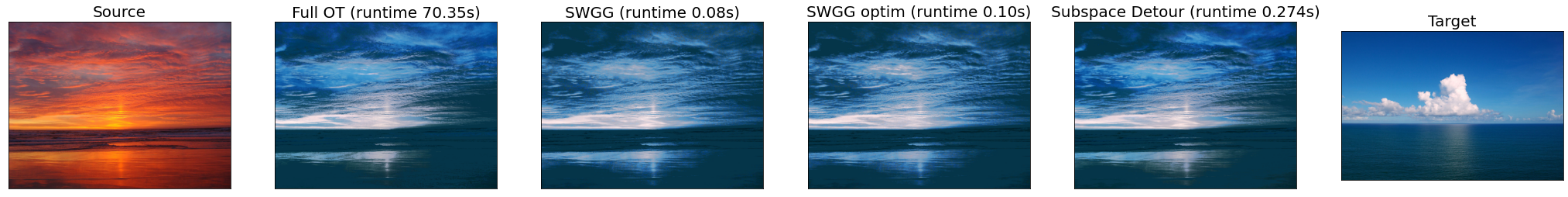}
\caption{Color transfer of images with $W^2_2$, $\mswgg$ and subspace detours, with runtimes.}
\label{fig:additional color transfer}
\end{figure}

We compare the results obtained with maps obtained from Wasserstein distance, $\mswgg$ with random search (100 projections), subspace detour \cite{muzellec2019subspace} and $\mswgg$ (optimized). Obtained images and the associated timings are provided in fig.~\ref{fig:additional color transfer}.

Figure \ref{fig:additional color transfer} shows that $\mswgg$ and $W_2^2$ provide visually equivalent solutions. Since, the quality of the color transfer is dependent on the size of the subsampling: using $\mswgg$ permits larger subsamples than $W^2_2$ and thus improves the quality of the map $T$. Moreover one can note that $\mswgg$ (optimized) is the fastest to compute.

We now give more details about how to perform color transfer between two distributions. The first step is to encode $n\times m$ images as $\mathbb{R}^{nm\times 3}$ vectors, with 3 channels in a RGB image. Note that $m$ and $n$ can differ for the source and target image.
The second step consists of defining subsamples $X_0,Y_0$ of $X,Y$, in our case we took $\boldsymbol{X}_0,\boldsymbol{Y}_0 \in \mathbb{R}^{20000\times d}$. We subsample the same number of points for the source and target image. In order to have a better subsampling of $\boldsymbol{X}$ and $\boldsymbol{Y}$, it is common to perform a $k$-means  \cite{kanungo2002efficient} to derive $\boldsymbol{X}_0$ and $\boldsymbol{Y}_0$ ($\boldsymbol{X}_0,\boldsymbol{Y}_0$ are then taken as centroids of the k-means algorithm).
The third step is to compute $T_0 : \boldsymbol{X}_0 \to \boldsymbol{Y}_0$. We set $T$ as the optimal Monge map given by the Wasserstein distance and $T$ as the optimal map given by $\mswgg$.
Finally, the fourth step deals with extending $T_0 : \boldsymbol{X}_0 \to \boldsymbol{Y}_0$ to $T:\boldsymbol{X}\to \boldsymbol{Y}$. $\forall \boldsymbol{x} \in \boldsymbol{X}$. We compute the closest element $\boldsymbol{x}_0 \in \boldsymbol{X}_0$ and we pose:
\begin{equation}
T(\boldsymbol{x})=T(\boldsymbol{x}_0).
\end{equation} 
More details on the overall procedure can be found in \cite{ferradans2013regularized}.

To perform the experiment, we took $L=100$ projections for $\mswgg$ (random search). For $\mswgg$ (optimized), we fixed the following set of parameters for the gradient descent: learning rate $5e^{-2}$, number of iterations 20, number of copies $s=10$ and $\epsilon=1$.
Regarding the  subspace detour results, we used the code of \cite{muzellec2019subspace} provided at \url{https://github.com/BorisMuzellec/SubspaceOT}.

\begin{figure}[ht]
\centering
\includegraphics[scale=0.2]{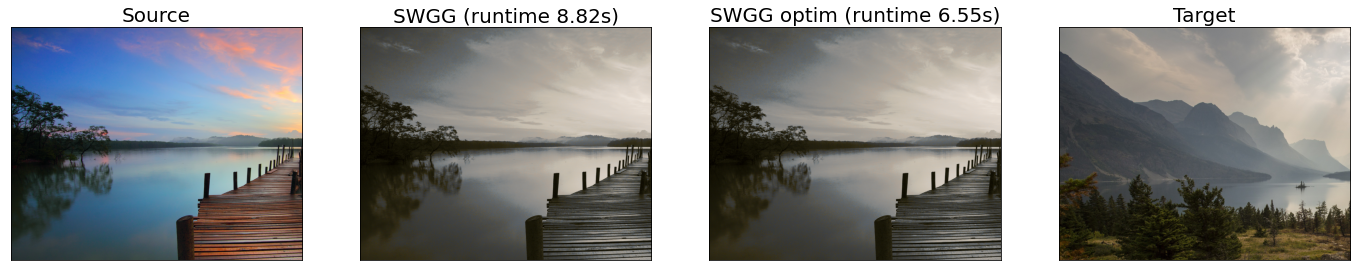}
\caption{Color transfer of images with $\mswgg$ without sub-sampling.}
\label{fig:additional color transfer2}
\end{figure}

Additionally, we perform color transfer without sub-sampling with the help of $\mswgg$ (we the same hyperparameters). This procedure is totally untractable for either $W^2_2$ and subspace detours (due to memory issues). As we mentioned before, the subsampling phase can decrease the quality of the transfer and thus $\mswgg$ can deliver better result than before. Result are give in Fig. \ref{fig:additional color transfer2}

\subsection{Data set distance}
We finally evaluate $\mswgg$ in an other context: computing distances between datasets.
\label{app:OTDD}
Let $\D_1=\{(\boldsymbol{x}^1_i,\boldsymbol{y}^1_i)\}_{i=1}^n$ and $\D_2=\{(\boldsymbol{x}^2_i,\boldsymbol{y}^2_i)\}_{i=1}^n$ be source and target data sets such that $\boldsymbol{x}^1_i,\boldsymbol{x}^2_i\in \R^d$ are samples and $\boldsymbol{y}_i^1,\boldsymbol{y}_i^2$ are labels $\forall 1 \leq i\leq n$. In \cite{alvarez2020geometric}, the authors compare those data sets using the Wasserstein distance with the entries of the cost matrix defined as:
\begin{align}
    C_{ij} = 
    \left(\|\boldsymbol{x}^1_i-\boldsymbol{x}^2_j\|^2_2+W^2_2(\alpha_{\boldsymbol{y}_i^1},\alpha_{\boldsymbol{y}_j^2})\right)^{1/2}
    \label{eq:otdd cost}
\end{align}
and the corresponding distance as:
\begin{align}
    OTDD(\D_1,\D_2) = \min_{P \in U}\langle C,P\rangle
    \label{eq:otdd problem}
\end{align}

where $\alpha_{\boldsymbol{y}}$ is the distribution of all samples with label $\boldsymbol{y}$, namely $\{\boldsymbol{x} \in \R^d | (\boldsymbol{x},\boldsymbol{y}) \in \D \}$ for $\D$ being either $\D_1$ or $\D_2$ and $U$ is the Birkhoff polytope which encodes the marginal constraints. Notice that cost in Eq. \eqref{eq:otdd cost} encompasses the ground distance and a label-to-label distance. This distance is  appealing in transfer learning application since it is model-agnostic. However, it can be cumbersome to compute in practice since it lays down on solving multiple OT problems (to compute the cost matrix and the OTDD). To circumvent that, \cite{alvarez2020geometric} proposed several methods to compute the cost matrix in Eq. \eqref{eq:otdd cost}. They used the Sinkhorn algorithm (in $\mathcal{O}(n^2)$) or they assumed $\alpha_{\boldsymbol{y}} \sim \mathcal{N}(m_{\boldsymbol{y}},\Sigma_{\boldsymbol{y}})$ in order to get the WD through the Bures metric (that provides a closed form of OT for Gaussian distributions in $\mathcal{O}(d^3)$), which is still prohibitive for high dimension. We  challenge $\mswgg$ in this context. 

In this experiment, we compare the following datasets: MNIST \cite{lecun1998mnist}, EMNIST \cite{cohen2017emnist}, FashionMNIST \cite{xiao2017fashion}, KMNIST \cite{clanuwat2018deep} and USPS \cite{uspsdataset}. We rely on the code of OTDD provided at \url{https://github.com/microsoft/otdd}. In order to make it compliant with the $\mswgg$ hypothesis, we require the empirical distributions $\alpha_{\boldsymbol{y}}$ to have the same number of atoms. 

Fig. \ref{fig:OTDD1} provides results for a batch size of $n=40000$ samples using the Sinkhorn divergence (with a regularisation parameter of $1e^{-1}$) and for $\mswgg$ (optimized) on batch of size 40000. We report results for a learning rate of $1e^{-5}$, 20 iterations and $s$ and $\epsilon$ to be 1 and 0. 

\begin{figure}[ht]
\begin{center}
\scalebox{.65}{\begin{tikzpicture}
    [
        box/.style={rectangle,draw=white,thick, minimum size=1cm},
    ]
\foreach \x in {0,1,...,5}{
    \foreach \y in {0,1,...,5}
        \node[box] at (\x,\y){};
}
\definecolor{color}{RGB}{144,93,111}
\node[box,fill=color] at (1,0){\textcolor{white}{1}}; 
\definecolor{color}{RGB}{165,113,90}
\node[box,fill=color] at (2,0){\textcolor{white}{1}}; 
\definecolor{color}{RGB}{102,51,153}
\node[box,fill=color] at (3,0){\textcolor{white}{0.8}}; 
\definecolor{color}{RGB}{171,119,84}
\node[box,fill=color] at (4,0){\textcolor{white}{1.1}}; 
\node[box,fill=white] at (5,0){};

\definecolor{color}{RGB}{226,173,29}
\node[box,fill=color] at (1,1){\textcolor{white}{1.2}}; 
\definecolor{color}{RGB}{225,172,30}
\node[box,fill=color] at (2,1){\textcolor{white}{1.2}}; 
\definecolor{color}{RGB}{255,202,0}
\node[box,fill=color] at (3,1){\textcolor{white}{1.3}}; 
\node[box,fill=white] at (4,1){};
\definecolor{color}{RGB}{173,122,81}
\node[box,fill=color] at (5,1){\textcolor{white}{1}}; 

\definecolor{color}{RGB}{236,184,19}
\node[box,fill=color] at (1,2){\textcolor{white}{1.3}}; 
\definecolor{color}{RGB}{217,165,38}
\node[box,fill=color] at (2,2){\textcolor{white}{1.2}};  
\node[box,fill=white] at (3,2){}; 
\definecolor{color}{RGB}{254,201,1}
\node[box,fill=color] at (4,2){\textcolor{white}{1.3}}; 
\definecolor{color}{RGB}{107,52,152}
\node[box,fill=color] at (5,2){\textcolor{white}{0.8}}; 

\definecolor{color}{RGB}{140,88,115}
\node[box,fill=color] at (1,3){\textcolor{white}{1}};  
\node[box,fill=white] at (2,3){}; 
\definecolor{color}{RGB}{217,165,38}
\node[box,fill=color] at (3,3){\textcolor{white}{1.2}}; 
\definecolor{color}{RGB}{224,171,31}
\node[box,fill=color] at (4,3){\textcolor{white}{1.2}};  
\definecolor{color}{RGB}{164,113,90}
\node[box,fill=color] at (5,3){\textcolor{white}{1}}; 

\node[box,fill=white] at (1,4){}; 
\definecolor{color}{RGB}{140,88,115}
\node[box,fill=color] at (2,4){\textcolor{white}{1}}; 
\definecolor{color}{RGB}{237,184,18}
\node[box,fill=color] at (3,4){\textcolor{white}{1.3}}; 
\definecolor{color}{RGB}{226,173,29}
\node[box,fill=color] at (4,4){\textcolor{white}{1.2}}; 
\definecolor{color}{RGB}{147,96,108}
\node[box,fill=color] at (5,4){\textcolor{white}{1}};

\node at (0,4) {{\tiny MNIST}};
\node at (0,3) {{\tiny EMNIST}};
\node at (0,2) {{\tiny Fashion}};
\node at (0,1) {{\tiny KMNIST}};
\node at (0,0) {{\tiny USPS}};
\node at (1,5) {{\tiny MNIST}};
\node at (2,5) {{\tiny EMNIST}};
\node at (3,5) {{\tiny Fashion}};
\node at (4,5) {{\tiny KMNIST}};
\node at (5,5) {{\tiny USPS}};
\end{tikzpicture}}
\scalebox{.65}{\begin{tikzpicture}
    [
        box/.style={rectangle,draw=white,thick, minimum size=1cm},
    ]
\foreach \x in {0,1,...,5}{
    \foreach \y in {0,1,...,5}
        \node[box] at (\x,\y){};
}
\definecolor{color}{RGB}{142,90,113}
\node[box,fill=color] at (1,0){\textcolor{white}{0.9}}; 
\definecolor{color}{RGB}{155,103,100}
\node[box,fill=color] at (2,0){\textcolor{white}{0.9}}; 
\definecolor{color}{RGB}{102,51,153}
\node[box,fill=color] at (3,0){\textcolor{white}{0.7}}; 
\definecolor{color}{RGB}{159,108,96}
\node[box,fill=color] at (4,0){\textcolor{white}{0.9}}; 
\node[box,fill=white] at (5,0){};

\definecolor{color}{RGB}{203,151,52}
\node[box,fill=color] at (1,1){\textcolor{white}{1.1}}; 
\definecolor{color}{RGB}{192,140,63}
\node[box,fill=color] at (2,1){\textcolor{white}{1}}; 
\definecolor{color}{RGB}{239,185,17}
\node[box,fill=color] at (3,1){\textcolor{white}{1.2}}; 
\node[box,fill=white] at (4,1){};
\definecolor{color}{RGB}{160,108,95}
\node[box,fill=color] at (5,1){\textcolor{white}{0.9}}; 

\definecolor{color}{RGB}{235,182,20}
\node[box,fill=color] at (1,2){\textcolor{white}{1.2}}; 
\definecolor{color}{RGB}{255,202,0}
\node[box,fill=color] at (2,2){\textcolor{white}{1.3}};  
\node[box,fill=white] at (3,2){}; 
\definecolor{color}{RGB}{238,185,17}
\node[box,fill=color] at (4,2){\textcolor{white}{1.2}}; 
\definecolor{color}{RGB}{102,51,153}
\node[box,fill=color] at (5,2){\textcolor{white}{0.8}}; 

\definecolor{color}{RGB}{118.2,67,137}
\node[box,fill=color] at (1,3){\textcolor{white}{0.8}};  
\node[box,fill=white] at (2,3){}; 
\definecolor{color}{RGB}{208,155,47}
\node[box,fill=color] at (3,3){\textcolor{white}{1.1}}; 
\definecolor{color}{RGB}{192,140,63}
\node[box,fill=color] at (4,3){\textcolor{white}{1.1}};  
\definecolor{color}{RGB}{155,103,100}
\node[box,fill=color] at (5,3){\textcolor{white}{0.9}}; 

\node[box,fill=white] at (1,4){}; 
\definecolor{color}{RGB}{119,68,136}
\node[box,fill=color] at (2,4){\textcolor{white}{0.9}}; 
\definecolor{color}{RGB}{236,184,18.5}
\node[box,fill=color] at (3,4){\textcolor{white}{1.2}}; 
\definecolor{color}{RGB}{203,151,52}
\node[box,fill=color] at (4,4){\textcolor{white}{1.1}}; 
\definecolor{color}{RGB}{142,91,113}
\node[box,fill=color] at (5,4){\textcolor{white}{0.9}};

\node at (0,4) {{\tiny MNIST}};
\node at (0,3) {{\tiny EMNIST}};
\node at (0,2) {{\tiny Fashion}};
\node at (0,1) {{\tiny KMNIST}};
\node at (0,0) {{\tiny USPS}};
\node at (1,5) {{\tiny MNIST}};
\node at (2,5) {{\tiny EMNIST}};
\node at (3,5) {{\tiny Fashion}};
\node at (4,5) {{\tiny KMNIST}};
\node at (5,5) {{\tiny USPS}};
\end{tikzpicture}}
\vskip +0.1in
\caption{OTDD results ($\times10^2$) distances for $\mswgg$ (left) and Sinkhorn divergence (right) for various datasets.}
\label{fig:OTDD1}
\end{center}
\end{figure}

We check that the orders of magnitude are preserved with $\mswgg$. For example OTDD(MNIST,USPS) is smaller than OTDD(MNIST,FashionMNIST) for either Sinkhorn divergence or $\mswgg$ as distance between labels, this validate that $\mswgg$ is a meaningful distance in this case scenario. Moreover in our setup, the computation cost is more expensive for Sinkhorn than for $\mswgg$ and totally untractable for $W^2_2$. On smaller batches (see Fig \ref{fig:OTDD Appendix}), the same observation can be made: $\mswgg$ is comparable (in term of magnitude) with $W^2_2$, Sinkhorn and the Bures approximation.

We give additional results in Fig. \ref{fig:OTDD Appendix} for batches of size of $n=2000$ samples obtained with $W^2_2$, Sinkhorn divergence (setting the entropic regularization parameter to $1e^{-1}$), the Bures approximation, $\mswgg$ (random search with $L = 1000$ projections) and $\mswgg$ (optimized, with a learning rate of $5e^{-1}$, 50 iterations, $s=20$ and $\epsilon=0.5$). 
\begin{figure}[ht]
    \centering
\scalebox{.42}{\begin{tikzpicture}
    [
        box/.style={rectangle,draw=white,thick, minimum size=1cm},
    ]
\foreach \x in {0,1,...,5}{
    \foreach \y in {0,1,...,5}
        \node[box] at (\x,\y){};
}
\definecolor{color}{RGB}{157,106,98}
\node[box,fill=color] at (1,0){\textcolor{white}{0.4}}; 
\definecolor{color}{RGB}{164,112,91}
\node[box,fill=color] at (2,0){\textcolor{white}{0.5}}; 
\definecolor{color}{RGB}{109,58,146}
\node[box,fill=color] at (3,0){\textcolor{white}{0.4}}; 
\definecolor{color}{RGB}{166,144,89}
\node[box,fill=color] at (4,0){\textcolor{white}{0.4}}; 
\node[box,fill=white] at (5,0){};

\definecolor{color}{RGB}{220,167,35}
\node[box,fill=color] at (1,1){\textcolor{white}{0.5}}; 
\definecolor{color}{RGB}{215,162,40}
\node[box,fill=color] at (2,1){\textcolor{white}{0.5}}; 
\definecolor{color}{RGB}{255,202,0}
\node[box,fill=color] at (3,1){\textcolor{white}{0.6}}; 
\node[box,fill=white] at (4,1){};
\definecolor{color}{RGB}{160,108,95}
\node[box,fill=color] at (5,1){\textcolor{white}{0.4}}; 

\definecolor{color}{RGB}{237,184,19}
\node[box,fill=color] at (1,2){\textcolor{white}{0.6}}; 
\definecolor{color}{RGB}{218,165,37}
\node[box,fill=color] at (2,2){\textcolor{white}{0.5}};  
\node[box,fill=white] at (3,2){}; 
\definecolor{color}{RGB}{243,190,12}
\node[box,fill=color] at (4,2){\textcolor{white}{0.6}}; 
\definecolor{color}{RGB}{102,24,153}
\node[box,fill=color] at (5,2){\textcolor{white}{0.3}}; 

\definecolor{color}{RGB}{141,89,114}
\node[box,fill=color] at (1,3){\textcolor{white}{0.4}};  
\node[box,fill=white] at (2,3){}; 
\definecolor{color}{RGB}{225,173,29}
\node[box,fill=color] at (3,3){\textcolor{white}{0.5}}; 
\definecolor{color}{RGB}{208,156,47}
\node[box,fill=color] at (4,3){\textcolor{white}{0.5}};  
\definecolor{color}{RGB}{158,107,96}
\node[box,fill=color] at (5,3){\textcolor{white}{0.4}}; 

\node[box,fill=white] at (1,4){}; 
\definecolor{color}{RGB}{138,87,117}
\node[box,fill=color] at (2,4){\textcolor{white}{0.4}}; 
\definecolor{color}{RGB}{251,198,4}
\node[box,fill=color] at (3,4){\textcolor{white}{0.6}}; 
\definecolor{color}{RGB}{227,174,28}
\node[box,fill=color] at (4,4){\textcolor{white}{0.5}}; 
\definecolor{color}{RGB}{157,105,98}
\node[box,fill=color] at (5,4){\textcolor{white}{0.4}};

\node at (0,4) {{\tiny MNIST}};
\node at (0,3) {{\tiny EMNIST}};
\node at (0,2) {{\tiny Fashion}};
\node at (0,1) {{\tiny KMNIST}};
\node at (0,0) {{\tiny USPS}};
\node at (1,5) {{\tiny MNIST}};
\node at (2,5) {{\tiny EMNIST}};
\node at (3,5) {{\tiny Fashion}};
\node at (4,5) {{\tiny KMNIST}};
\node at (5,5) {{\tiny USPS}};
\end{tikzpicture}}
\scalebox{.42}{\begin{tikzpicture}
    [
        box/.style={rectangle,draw=white,thick, minimum size=1cm},
    ]
\foreach \x in {0,1,...,5}{
    \foreach \y in {0,1,...,5}
        \node[box] at (\x,\y){};
}
\definecolor{color}{RGB}{149,97,106}
\node[box,fill=color] at (1,0){\textcolor{white}{1}}; 
\definecolor{color}{RGB}{163,112,91}
\node[box,fill=color] at (2,0){\textcolor{white}{1}}; 
\definecolor{color}{RGB}{102,51,153}
\node[box,fill=color] at (3,0){\textcolor{white}{0.8}}; 
\definecolor{color}{RGB}{164,113,90}
\node[box,fill=color] at (4,0){\textcolor{white}{1}};  
\node[box,fill=white] at (5,0){};

\definecolor{color}{RGB}{220,168,34}
\node[box,fill=color] at (1,1){\textcolor{white}{1.2}}; 
\definecolor{color}{RGB}{217,165,38}
\node[box,fill=color] at (2,1){\textcolor{white}{1.2}};  
\definecolor{color}{RGB}{253,200,1}
\node[box,fill=color] at (3,1){\textcolor{white}{1.3}};  
\node[box,fill=white] at (4,1){};
\definecolor{color}{RGB}{165,113,90}
\node[box,fill=color] at (5,1){\textcolor{white}{1}}; 

\definecolor{color}{RGB}{242,189,13}
\node[box,fill=color] at (1,2){\textcolor{white}{1.3}};  
\definecolor{color}{RGB}{218,166,36}
\node[box,fill=color] at (2,2){\textcolor{white}{1.2}};  
\node[box,fill=white] at (3,2){}; 
\definecolor{color}{RGB}{225,202,0}
\node[box,fill=color] at (4,2){\textcolor{white}{1.3}}; 
\definecolor{color}{RGB}{103,52,152}
\node[box,fill=color] at (5,2){\textcolor{white}{0.8}}; 

\definecolor{color}{RGB}{139,87,116}
\node[box,fill=color] at (1,3){\textcolor{white}{0.9}};  
\node[box,fill=white] at (2,3){}; 
\definecolor{color}{RGB}{213,160,42}
\node[box,fill=color] at (3,3){\textcolor{white}{1.2}}; 
\definecolor{color}{RGB}{222,169,33}
\node[box,fill=color] at (4,3){\textcolor{white}{1.2}}; 
\definecolor{color}{RGB}{160,108,95}
\node[box,fill=color] at (5,3){\textcolor{white}{1}}; 

\node[box,fill=white] at (1,4){}; 
\definecolor{color}{RGB}{138,98,106}
\node[box,fill=color] at (2,4){\textcolor{white}{0.9}}; 
\definecolor{color}{RGB}{244,191,11}
\node[box,fill=color] at (3,4){\textcolor{white}{1.3}}; 
\definecolor{color}{RGB}{220,167,35}
\node[box,fill=color] at (4,4){\textcolor{white}{1.2}}; 
\definecolor{color}{RGB}{148,96,107}
\node[box,fill=color] at (5,4){\textcolor{white}{0.9}};

\node at (0,4) {{\tiny MNIST}};
\node at (0,3) {{\tiny EMNIST}};
\node at (0,2) {{\tiny Fashion}};
\node at (0,1) {{\tiny KMNIST}};
\node at (0,0) {{\tiny USPS}};
\node at (1,5) {{\tiny MNIST}};
\node at (2,5) {{\tiny EMNIST}};
\node at (3,5) {{\tiny Fashion}};
\node at (4,5) {{\tiny KMNIST}};
\node at (5,5) {{\tiny USPS}};
\end{tikzpicture}}
\scalebox{.42}{\begin{tikzpicture}
    [
        box/.style={rectangle,draw=white,thick, minimum size=1cm},
    ]
\foreach \x in {0,1,...,5}{
    \foreach \y in {0,1,...,5}
        \node[box] at (\x,\y){};
}
\definecolor{color}{RGB}{144,93,110}
\node[box,fill=color] at (1,0){\textcolor{white}{0.9}};  
\definecolor{color}{RGB}{162,110,93}
\node[box,fill=color] at (2,0){\textcolor{white}{0.9}};  
\definecolor{color}{RGB}{107,56,147}
\node[box,fill=color] at (3,0){\textcolor{white}{0.7}}; 
\definecolor{color}{RGB}{154,102,101}
\node[box,fill=color] at (4,0){\textcolor{white}{0.9}}; 
\node[box,fill=white] at (5,0){};

\definecolor{color}{RGB}{199,147,55}
\node[box,fill=color] at (1,1){\textcolor{white}{1}};  
\definecolor{color}{RGB}{186,134,69}
\node[box,fill=color] at (2,1){\textcolor{white}{1}};  
\definecolor{color}{RGB}{242,190,12}
\node[box,fill=color] at (3,1){\textcolor{white}{1.1}}; 
\node[box,fill=white] at (4,1){};
\definecolor{color}{RGB}{156,104,99}
\node[box,fill=color] at (5,1){\textcolor{white}{0.9}};  

\definecolor{color}{RGB}{255,202,0}
\node[box,fill=color] at (1,2){\textcolor{white}{1.2}}; 
\definecolor{color}{RGB}{222,169,33}
\node[box,fill=color] at (2,2){\textcolor{white}{1.1}}; 
\node[box,fill=white] at (3,2){}; 
\definecolor{color}{RGB}{225,172,30}
\node[box,fill=color] at (4,2){\textcolor{white}{1.1}};  
\definecolor{color}{RGB}{102,51,153}
\node[box,fill=color] at (5,2){\textcolor{white}{0.7}}; 

\definecolor{color}{RGB}{118,27,136}
\node[box,fill=color] at (1,3){\textcolor{white}{0.8}};  
\node[box,fill=white] at (2,3){}; 
\definecolor{color}{RGB}{224,171,31}
\node[box,fill=color] at (3,3){\textcolor{white}{1.1}}; 
\definecolor{color}{RGB}{185,133,69}
\node[box,fill=color] at (4,3){\textcolor{white}{1}}; 
\definecolor{color}{RGB}{160,108,95}
\node[box,fill=color] at (5,3){\textcolor{white}{0.9}}; 

\node[box,fill=white] at (1,4){}; 
\definecolor{color}{RGB}{117,66,137}
\node[box,fill=color] at (2,4){\textcolor{white}{0.8}}; 
\definecolor{color}{RGB}{244,191,11}
\node[box,fill=color] at (3,4){\textcolor{white}{1.1}}; 
\definecolor{color}{RGB}{199,147,56}
\node[box,fill=color] at (4,4){\textcolor{white}{1}}; 
\definecolor{color}{RGB}{147,95,108}
\node[box,fill=color] at (5,4){\textcolor{white}{0.8}};

\node at (0,4) {{\tiny MNIST}};
\node at (0,3) {{\tiny EMNIST}};
\node at (0,2) {{\tiny Fashion}};
\node at (0,1) {{\tiny KMNIST}};
\node at (0,0) {{\tiny USPS}};
\node at (1,5) {{\tiny MNIST}};
\node at (2,5) {{\tiny EMNIST}};
\node at (3,5) {{\tiny Fashion}};
\node at (4,5) {{\tiny KMNIST}};
\node at (5,5) {{\tiny USPS}};
\end{tikzpicture}}
\scalebox{.42}{\begin{tikzpicture}
    [
        box/.style={rectangle,draw=white,thick, minimum size=1cm},
    ]
\foreach \x in {0,1,...,5}{
    \foreach \y in {0,1,...,5}
        \node[box] at (\x,\y){};
}
\definecolor{color}{RGB}{150,99,105}
\node[box,fill=color] at (1,0){\textcolor{white}{1}};  
\definecolor{color}{RGB}{166,114,89}
\node[box,fill=color] at (2,0){\textcolor{white}{1.1}};  
\definecolor{color}{RGB}{102,51,153}
\node[box,fill=color] at (3,0){\textcolor{white}{0.9}}; 
\definecolor{color}{RGB}{174,122,81}
\node[box,fill=color] at (4,0){\textcolor{white}{1.1}}; 
\node[box,fill=white] at (5,0){};

\definecolor{color}{RGB}{230,178,25}
\node[box,fill=color] at (1,1){\textcolor{white}{1.3}};  
\definecolor{color}{RGB}{240,187,15}
\node[box,fill=color] at (2,1){\textcolor{white}{1.3}};  
\definecolor{color}{RGB}{249,196,5}
\node[box,fill=color] at (3,1){\textcolor{white}{1.4}}; 
\node[box,fill=white] at (4,1){};
\definecolor{color}{RGB}{173,121,82}
\node[box,fill=color] at (5,1){\textcolor{white}{1.1}}; 

\definecolor{color}{RGB}{233,180,22}
\node[box,fill=color] at (1,2){\textcolor{white}{1.3}}; 
\definecolor{color}{RGB}{217,164,38}
\node[box,fill=color] at (2,2){\textcolor{white}{1.3}}; 
\node[box,fill=white] at (3,2){}; 
\definecolor{color}{RGB}{255,202,0}
\node[box,fill=color] at (4,2){\textcolor{white}{1.4}}; 
\definecolor{color}{RGB}{106,55,146}
\node[box,fill=color] at (5,2){\textcolor{white}{0.8}}; 

\definecolor{color}{RGB}{149,98,105}
\node[box,fill=color] at (1,3){\textcolor{white}{1}}; 
\node[box,fill=white] at (2,3){}; 
\definecolor{color}{RGB}{227,175,27}
\node[box,fill=color] at (3,3){\textcolor{white}{1.3}}; 
\definecolor{color}{RGB}{237,184,18}
\node[box,fill=color] at (4,3){\textcolor{white}{1.3}}; 
\definecolor{color}{RGB}{165,113,89}
\node[box,fill=color] at (5,3){\textcolor{white}{1.1}}; 

\node[box,fill=white] at (1,4){}; 
\definecolor{color}{RGB}{152,100,103}
\node[box,fill=color] at (2,4){\textcolor{white}{1}}; 
\definecolor{color}{RGB}{234,181,21}
\node[box,fill=color] at (3,4){\textcolor{white}{1.3}}; 
\definecolor{color}{RGB}{234,181,21}
\node[box,fill=color] at (4,4){\textcolor{white}{1.3}}; 
\definecolor{color}{RGB}{154,102,101}
\node[box,fill=color] at (5,4){\textcolor{white}{1.1}};

\node at (0,4) {{\tiny MNIST}};
\node at (0,3) {{\tiny EMNIST}};
\node at (0,2) {{\tiny Fashion}};
\node at (0,1) {{\tiny KMNIST}};
\node at (0,0) {{\tiny USPS}};
\node at (1,5) {{\tiny MNIST}};
\node at (2,5) {{\tiny EMNIST}};
\node at (3,5) {{\tiny Fashion}};
\node at (4,5) {{\tiny KMNIST}};
\node at (5,5) {{\tiny USPS}};
\end{tikzpicture}}
\scalebox{.42}{\begin{tikzpicture}
    [
        box/.style={rectangle,draw=white,thick, minimum size=1cm},
    ]
\foreach \x in {0,1,...,5}{
    \foreach \y in {0,1,...,5}
        \node[box] at (\x,\y){};
}
\definecolor{color}{RGB}{150,99,105}
\node[box,fill=color] at (1,0){\textcolor{white}{1}}; 
\definecolor{color}{RGB}{166,114,89}
\node[box,fill=color] at (2,0){\textcolor{white}{1.1}}; 
\definecolor{color}{RGB}{102,51,153}
\node[box,fill=color] at (3,0){\textcolor{white}{0.8}}; 
\definecolor{color}{RGB}{174,122,81}
\node[box,fill=color] at (4,0){\textcolor{white}{1.1}}; 
\node[box,fill=white] at (5,0){};

\definecolor{color}{RGB}{230,178,25}
\node[box,fill=color] at (1,1){\textcolor{white}{1.3}};  
\definecolor{color}{RGB}{240,188,14}
\node[box,fill=color] at (2,1){\textcolor{white}{1.3}};  
\definecolor{color}{RGB}{250,196,5}
\node[box,fill=color] at (3,1){\textcolor{white}{1.3}};  
\node[box,fill=white] at (4,1){};
\definecolor{color}{RGB}{173,121,82}
\node[box,fill=color] at (5,1){\textcolor{white}{1.1}}; 

\definecolor{color}{RGB}{233,180,22}
\node[box,fill=color] at (1,2){\textcolor{white}{1.3}}; 
\definecolor{color}{RGB}{217,164,38}
\node[box,fill=color] at (2,2){\textcolor{white}{1.2}};  
\node[box,fill=white] at (3,2){}; 
\definecolor{color}{RGB}{255,202,0}
\node[box,fill=color] at (4,2){\textcolor{white}{1.4}}; 
\definecolor{color}{RGB}{106,55,150}
\node[box,fill=color] at (5,2){\textcolor{white}{0.8}}; 

\definecolor{color}{RGB}{149,98,105}
\node[box,fill=color] at (1,3){\textcolor{white}{1}}; 
\node[box,fill=white] at (2,3){}; 
\definecolor{color}{RGB}{227,175,27}
\node[box,fill=color] at (3,3){\textcolor{white}{1.3}}; 
\definecolor{color}{RGB}{237,184,18}
\node[box,fill=color] at (4,3){\textcolor{white}{1.3}};  
\definecolor{color}{RGB}{165,114,90}
\node[box,fill=color] at (5,3){\textcolor{white}{1.1}}; 

\node[box,fill=white] at (1,4){}; 
\definecolor{color}{RGB}{152,100,103}
\node[box,fill=color] at (2,4){\textcolor{white}{1}}; 
\definecolor{color}{RGB}{233,181,21}
\node[box,fill=color] at (3,4){\textcolor{white}{1.3}}; 
\definecolor{color}{RGB}{233,181,21}
\node[box,fill=color] at (4,4){\textcolor{white}{1.3}}; 
\definecolor{color}{RGB}{157,102,101}
\node[box,fill=color] at (5,4){\textcolor{white}{1}};

\node at (0,4) {{\tiny MNIST}};
\node at (0,3) {{\tiny EMNIST}};
\node at (0,2) {{\tiny Fashion}};
\node at (0,1) {{\tiny KMNIST}};
\node at (0,0) {{\tiny USPS}};
\node at (1,5) {{\tiny MNIST}};
\node at (2,5) {{\tiny EMNIST}};
\node at (3,5) {{\tiny Fashion}};
\node at (4,5) {{\tiny KMNIST}};
\node at (5,5) {{\tiny USPS}};
\end{tikzpicture}}
\vskip +0.1in
\caption{OTDD distance with $W^2_2$ (left), Sinkhorn (left-mid), Bures (middle) and random search $\mswgg$ (right-mid) and $\mswgg$-optimization (right) distances between labels distribution $\times 10^2$}
    \label{fig:OTDD Appendix}
\end{figure}

Note that the Figs. \ref{fig:OTDD1} and \ref{fig:OTDD Appendix} are not symmetric (OTDD(KMNIST,FashionMNIST)$\neq$ OTDD(FashinMNIST,KMNIST)) because of the random aspect of batches.

\newpage

\end{document}